\documentclass{article} %
\usepackage{iclr2024_conference,times}

\usepackage{amsmath,amsfonts,bm}

\def\figref#1{Fig.~\ref{#1}}

\def\eqref#1{Eq.~(\ref{#1})}

\def\1{\bm{1}}

\def\eps{{\epsilon}}

\def\vmu{{\bm{\mu}}}
\def\vtheta{{\bm{\theta}}}

\def\va{{\bm{a}}}

\def\vs{{\bm{s}}}

\def\vw{{\bm{w}}}

\def\vy{{\bm{y}}}

\def\mSigma{{\bm{\Sigma}}}

\DeclareMathAlphabet{\mathsfit}{\encodingdefault}{\sfdefault}{m}{sl}
\SetMathAlphabet{\mathsfit}{bold}{\encodingdefault}{\sfdefault}{bx}{n}

\newcommand{\E}{\mathbb{E}}

\DeclareMathOperator*{\argmin}{arg\,min}

\usepackage{xcolor}

\usepackage{tikz}
\usetikzlibrary{positioning, arrows, automata, matrix}
\usepackage{pgfplots}
\pgfplotsset{compat=1.8}
\usepackage{pgfplotstable}
\usepgfplotslibrary{groupplots}
\usepgfplotslibrary{colorbrewer}
\usetikzlibrary{external}
\usetikzlibrary{backgrounds, calc}
\usetikzlibrary{spy}
\usetikzlibrary{fit}
\usetikzlibrary{arrows.meta}
\usepackage{adjustbox}
\usepackage{graphicx}

\definecolor{C0}{HTML}{1f77b4}  %
\definecolor{C1}{HTML}{ff7f0e}  %
\definecolor{C2}{HTML}{2ca02c}  %
\definecolor{C3}{HTML}{d62728}  %
\definecolor{C4}{HTML}{9467bd}  %
\definecolor{C5}{HTML}{8c564b}  %
\definecolor{C6}{HTML}{e377c2}  %
\definecolor{C7}{HTML}{7f7f7f}  %
\definecolor{C8}{HTML}{bcbd22}  %
\definecolor{C9}{HTML}{17becf}  %
\definecolor{C10}{HTML}{7fbc41} %

\definecolor{darkgray176}{RGB}{176,176,176}
\definecolor{lightgray204}{RGB}{204,204,204}

\usepackage{hyperref}
\usepackage{url}
\usepackage{xspace}
\usepackage{amssymb}
\usepackage{multirow}

\usepackage[acronym]{glossaries}
\usepackage[inline]{enumitem}
\usepackage{graphicx}
\usepackage{subcaption}
\usepackage{booktabs}
\usepackage{setspace}
\usepackage{wrapfig}
\usepackage{afterpage} %

\usepackage[utf8]{inputenc}
\usepackage{algorithm}
\usepackage{algorithmicx}
\usepackage{algpseudocode}

\newcommand{\todo}[1]{{\color{red}#1}}

\renewcommand{\vec}[1]{{\boldsymbol{#1}}}
\newif\ifshowqa
\showqatrue
\showqafalse
\definecolor{papercolor}{HTML}{C7EDCC}

\title{\centering Open the Black Box: Step-based Policy \\Updates for Temporally-Correlated Episodic \\ Reinforcement Learning}

\author{
~~~ Ge~Li$^{1,}$\thanks{Corresponding author. Email to $<$geli.bruce.ai@gmail.com, ge.li@kit.edu$>$} ~~~~~ Hongyi Zhou$^1$ ~~~~~ Dominik Roth$^1$ ~~~~~ Serge Thilges$^1$ ~~~~~ Fabian Otto$^{2,3}$
\\[0.2em]
~~~~~~~~~~~~~~~~~~~~~~~~~~~~~~~~~~~~~~~
\textbf{Rudolf ~Lioutikov}$^{1}$ ~~~~~ \textbf{Gerhard ~Neumann}$^{1}$
\\[0.5em]
~~ $^{1}$Karlsruhe Institute of Technology, Germany ~~~~~~ $^{2}$University of Tübingen, Germany 
\\[0.2em]
~~~~~~~~~~~~~~~~~~~~~~~~~~~~~~~~~~ $^{3}$Bosch Center for Artificial Intelligence, Germany
}

\iclrfinalcopy %
\begin{document}

\maketitle

\glsdisablehyper

\newacronym{acnmp}{ACNMP}{Adaptive Conditional Neural Movement Primitive}
\newacronym{ba}{BA}{Bayesian Aggregation}
\newacronym{bbrl}{BBRL}{Black Box Reinforcement Learning}
\newacronym{bc}{BC}{boundary condition}
\newacronym{clv}{CLV}{conditional latent variable}
\newacronym{cnmp}{CNMP}{Conditional Neural Movement Primitive}
\newacronym{cnn}{CNNs}{convolutional neural networks}
\newacronym{cnp}{CNP}{Conditional Neural Processes}
\newacronym{dmp}{DMP}{Dynamic Movement Primitive}
\newacronym{dof}{DoF}{Degree of Freedom}
\newacronym{es}{ES}{evolution strategies}
\newacronym{gp}{GPs}{Gaussian Processes}
\newacronym{gsde}{gSDE}{Generalized State Dependent Exploration}
\newacronym{idmp}{IDMP}{Integral form of Dynamic Movement Primitive}
\newacronym{ik}{IK}{inverse kinematics}
\newacronym{il}{IL}{Imitation Learning}
\newacronym{ma}{MA}{Mean Aggregation}
\newacronym{ml}{ML}{Machine Learning}
\newacronym{mp}{MP}{Movement Primitive}
\newacronym{mse}{MSE}{mean squared error}
\newacronym{ndp}{NDP}{Neural Dynamic Policies}
\newacronym{nmp}{ProDMP}{Probabilistic Dynamic Movement Primitive}
\newacronym{nn}{NN}{neural network}
\newacronym{np}{NP}{Neural Processes}
\newacronym{ode}{ODE}{ordinary differential equation}
\newacronym{pdmp}{ProDMP}{Probabilistic Dynamic Movement Primitive}
\newacronym{pink}{PINK}{Pink Noise}
\newacronym{ppo}{PPO}{Proximal Policy Optimization}
\newacronym{promp}{ProMP}{Probabilistic Movement Primitive}
\newacronym{rl}{RL}{Reinforcement Learning}
\newacronym{sse}{SSE}{Summed Squared Error}
\newacronym{tcp}{TCE}{Temporally-Correlated Episodic RL}
\newacronym{tce}{TCE}{Temporally-Correlated Episodic RL}
\newacronym{td3}{TD3}{twin-delayed deep deterministic policy gradient}
\newacronym{trpl}{TRPL}{trust region projection layer}
\newacronym{trpo}{TRPO}{trust region policy optimization}
\newacronym{sac}{SAC}{soft actor critic}

\newcommand{\dof}{\acrshort{dof}\xspace}
\newcommand{\fig}{Fig.\xspace}
\newcommand{\dmp}{\acrshortpl{dmp}\xspace}
\newcommand{\Dmp}{\acrlongpl{dmp}\xspace}
\newcommand{\DMP}{\acrfullpl{dmp}\xspace}
\renewcommand{\mp}{\acrshortpl{mp}\xspace}
\newcommand{\Mp}{\acrlongpl{mp}\xspace}
\newcommand{\MP}{\acrfullpl{mp}\xspace}
\newcommand{\nn}{\acrshortpl{nn}\xspace}
\newcommand{\Nn}{\acrlongpl{nn}\xspace}
\newcommand{\NN}{\acrfullpl{nn}\xspace}
\newcommand{\ode}{\acrshort{ode}\xspace}
\newcommand{\Ode}{\acrlong{ode}\xspace}
\newcommand{\ODE}{\acrfull{ode}\xspace}
\newcommand{\pdmp}{\acrshortpl{pdmp}\xspace}
\newcommand{\Pdmp}{\acrlongpl{pdmp}\xspace}
\newcommand{\PDMP}{\acrfullpl{pdmp}\xspace}
\newcommand{\promp}{\acrshortpl{promp}\xspace}
\newcommand{\Promp}{\acrlongpl{promp}\xspace}
\newcommand{\PROMP}{\acrfullpl{promp}\xspace}
\newcommand{\rl}{\acrshort{rl}\xspace}
\newcommand{\Rl}{\acrlong{rl}\xspace}
\newcommand{\RL}{\acrfull{rl}\xspace}
\newcommand{\tcp}{\acrshort{tcp}\xspace}
\newcommand{\Tcp}{\acrlong{tcp}\xspace}
\newcommand{\TCP}{\acrfull{tcp}\xspace}
\newcommand{\tce}{\acrshort{tcp}\xspace}
\newcommand{\Tce}{\acrlong{tcp}\xspace}
\newcommand{\TCE}{\acrfull{tcp}\xspace}
\newcommand{\trpl}{\acrshort{trpl}\xspace}
\newcommand{\Trpl}{\acrlong{trpl}\xspace}
\newcommand{\TRPL}{\acrfull{trpl}\xspace}

\newcommand{\ipb}{\bm{\Phi}}
\newcommand{\ivb}{\dot{\bm{\Phi}}}
\newcommand{\bipb}{\bm{\Psi}}
\newcommand{\bivb}{\dot{\bm{\Psi}}}

\newcommand{\ipbt}{\bm{\Phi}^\intercal}
\newcommand{\ivbt}{\dot{\bm{\Phi}}^\intercal}
\newcommand{\bipbt}{\bm{\Psi}^\intercal}
\newcommand{\bivbt}{\dot{\bm{\Psi}}^\intercal}

\newcommand{\ipbb}{\bm{\Phi}_b}
\newcommand{\ivbb}{\dot{\bm{\Phi}}_b}
\newcommand{\bipbb}{\bm{\Psi}_b}
\newcommand{\bivbb}{\dot{\bm{\Psi}}_b}

\newcommand{\ipbbt}{\bm{\Phi}^\intercal_b}
\newcommand{\ivbbt}{\dot{\bm{\Phi}}^\intercal_b}
\newcommand{\bipbbt}{\bm{\Psi}^\intercal_b}
\newcommand{\bivbbt}{\dot{\bm{\Psi}}^\intercal_b}

\newcommand{\inv}[1]{{#1}^{-1}}
\newcommand{\invT}[1]{{#1}^{-T}}
\newcommand{\norm}[1]{\left\lVert#1\right\rVert}
\newcommand{\old}[1]{{#1}_{\textrm{old}}}
\newcommand{\oldInv}[1]{{#1}_{\textrm{old}}^{-1}}

\newcommand{\ie}{i.\,e.\xspace}

\newcommand{\eg}{e.\,g.\xspace}

\newcommand{\cf}{cf.\xspace}

\renewcommand{\fig}{Fig.\xspace}

\newcommand{\ps}{per-step \xspace}

\newcommand{\st}{\text{s.\,t.\xspace}}

\newcommand{\ts}{time-step \xspace}

\begin{abstract}

Current advancements in reinforcement learning (RL) have predominantly focused on learning step-based policies that generate actions for each perceived state. While these methods efficiently leverage step information from environmental interaction, they often ignore the temporal correlation between actions, resulting in inefficient exploration and unsmooth trajectories that are challenging to implement on real hardware. Episodic RL (ERL) seeks to overcome these challenges by exploring in parameters space that capture the correlation of actions. However, these approaches typically compromise data efficiency, as they treat trajectories as opaque \emph{black boxes}. In this work, we introduce a novel ERL algorithm, Temporally-Correlated Episodic RL (TCE), which effectively utilizes step information in episodic policy updates, opening the 'black box' in existing ERL methods while retaining the smooth and consistent exploration in parameter space. TCE synergistically combines the advantages of step-based and episodic RL, achieving comparable performance to recent ERL methods while maintaining data efficiency akin to state-of-the-art (SoTA) step-based RL. 
Our code is available at \url{https://github.com/BruceGeLi/TCE_RL}.
\end{abstract}

\section{Introduction}
\begin{figure}[t!]
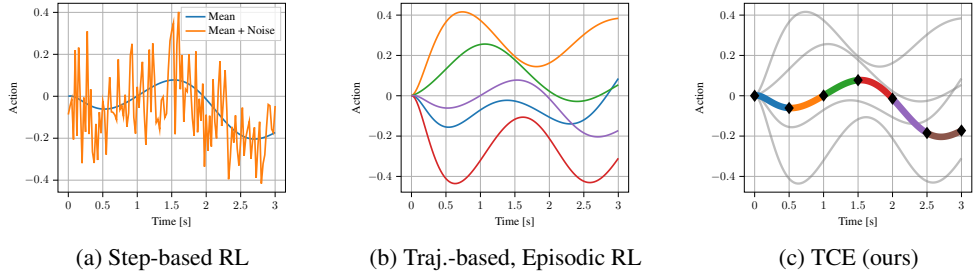
    
    \centering
    \hspace*{\fill}%
    \begin{subfigure}{0.3\textwidth}        
        \resizebox{0.9\textwidth}{!}{\begin{tikzpicture}

\definecolor{darkgray176}{RGB}{176,176,176}
\definecolor{darkorange25512714}{RGB}{255,127,14}
\definecolor{lightgray204}{RGB}{204,204,204}
\definecolor{steelblue31119180}{RGB}{31,119,180}
\def\linewidth{0.5mm}
\begin{axis}[
legend cell align={left},
legend style={fill opacity=0.8, draw opacity=1, text opacity=1, draw=lightgray204},
tick align=outside,
tick pos=left,
x grid style={darkgray176},
xlabel={Time [s]},
xmajorgrids,
xmin=-0.15, xmax=3.15,
xtick style={color=black},
y grid style={darkgray176},
ylabel={Action},
ymajorgrids,
ymin=-0.457129457592964, ymax=0.443328180909157,
ytick style={color=black}
]
\addplot [line width=\linewidth, steelblue31119180]
table {%
0 0
0.0250000022351742 0.00132888555526733
0.0500000044703484 0.000835120677947998
0.0750000029802322 -0.00119742751121521
0.100000008940697 -0.00430575013160706
0.125 -0.00826326012611389
0.150000005960464 -0.0127565264701843
0.174999997019768 -0.0176048576831818
0.200000017881393 -0.0226060748100281
0.225000008940697 -0.027616024017334
0.25 -0.0325145721435547
0.275000005960464 -0.0371871888637543
0.300000011920929 -0.0415735244750977
0.324999988079071 -0.0455831587314606
0.349999994039536 -0.0491979122161865
0.375 -0.0523473620414734
0.400000035762787 -0.0550429224967957
0.425000011920929 -0.0572300851345062
0.450000017881393 -0.058940201997757
0.474999994039536 -0.0601321756839752
0.5 -0.060849666595459
0.524999976158142 -0.0610618591308594
0.550000011920929 -0.0608198642730713
0.575000047683716 -0.0601015090942383
0.600000023841858 -0.0589609146118164
0.625 -0.0573831498622894
0.649999976158142 -0.0554222166538239
0.675000011920929 -0.0530692040920258
0.699999988079071 -0.0503761768341064
0.725000023841858 -0.0473393797874451
0.75 -0.0440076291561127
0.774999976158142 -0.0403813719749451
0.800000071525574 -0.0365063548088074
0.825000047683716 -0.0323863923549652
0.850000023841858 -0.0280632078647614
0.875000059604645 -0.0235447883605957
0.900000035762787 -0.0188689231872559
0.925000011920929 -0.0140470564365387
0.949999988079071 -0.00911316275596619
0.975000023841858 -0.0040835440158844
1 0.00101196765899658
1.02499997615814 0.00615376234054565
1.04999995231628 0.0113139152526855
1.07500004768372 0.0164682269096375
1.10000002384186 0.0215928554534912
1.125 0.0266581773757935
1.15000009536743 0.0316434502601624
1.17500007152557 0.0365153551101685
1.20000004768372 0.0412558913230896
1.22500002384186 0.0458242297172546
1.25 0.0502094626426697
1.27500009536743 0.0543646812438965
1.29999995231628 0.0582819581031799
1.32500004768372 0.0619113445281982
1.35000002384186 0.0652486085891724
1.375 0.0682399272918701
1.39999997615814 0.0708852410316467
1.42500007152557 0.073128879070282
1.45000004768372 0.0749759674072266
1.47500002384186 0.0763683915138245
1.5 0.0773192048072815
1.52499997615814 0.0777676701545715
1.54999995231628 0.0777335166931152
1.57499992847443 0.0771585702896118
1.59999990463257 0.076069176197052
1.62499988079071 0.0744092464447021
1.65000009536743 0.0722140669822693
1.67499995231628 0.0694298148155212
1.70000004768372 0.0661012530326843
1.72499990463257 0.0621816515922546
1.75 0.0577211976051331
1.77499985694885 0.0526820421218872
1.79999995231628 0.0471206307411194
1.82500004768372 0.0410100817680359
1.85000002384186 0.0344118475914001
1.875 0.0273085832595825
1.89999997615814 0.0197672247886658
1.92499995231628 0.0117847323417664
1.94999992847443 0.00342720746994019
1.97499990463257 -0.00529712438583374
2 -0.0143193602561951
2.02500009536743 -0.0236175060272217
2.04999995231628 -0.0331268608570099
2.07500004768372 -0.0428137183189392
2.09999990463257 -0.0526146292686462
2.125 -0.0624864995479584
2.14999985694885 -0.0723739564418793
2.17499995231628 -0.0822190642356873
2.20000004768372 -0.0919783413410187
2.22499990463257 -0.101586788892746
2.25 -0.111006081104279
2.27500009536743 -0.120169281959534
2.29999995231628 -0.129049569368362
2.32499980926514 -0.137572884559631
2.34999990463257 -0.145726323127747
2.375 -0.153436571359634
2.40000009536743 -0.160702258348465
2.42499995231628 -0.16745188832283
2.45000004768372 -0.173695653676987
2.47499990463257 -0.179366737604141
2.5 -0.184488445520401
2.52499985694885 -0.189000338315964
2.55000019073486 -0.192935973405838
2.57500004768372 -0.196242570877075
2.59999990463257 -0.198965460062027
2.625 -0.2010597884655
2.65000009536743 -0.202580541372299
2.67499995231628 -0.20349046587944
2.69999980926514 -0.20384955406189
2.72499990463257 -0.203634172677994
2.75 -0.202909767627716
2.77500009536743 -0.201656997203827
2.79999995231628 -0.199943691492081
2.82500004768372 -0.19776263833046
2.84999990463257 -0.195178627967834
2.875 -0.192194640636444
2.89999985694885 -0.188872009515762
2.92500019073486 -0.185217648744583
2.95000004768372 -0.181291848421097
2.97499990463257 -0.177107751369476
3 -0.172720789909363
};
\addlegendentry{Mean}
\addplot [line width=\linewidth, darkorange25512714]
table {%
0 -0.0871067941188812
0.0250000022351742 -0.0466643609106541
0.0500000044703484 -0.0138547271490097
0.0750000029802322 -0.206003680825233
0.100000008940697 0.220340833067894
0.125 -0.189371466636658
0.150000005960464 0.232070788741112
0.174999997019768 -0.0314996428787708
0.200000017881393 -0.318943947553635
0.225000008940697 -0.00609868764877319
0.25 -0.169629544019699
0.275000005960464 0.309652626514435
0.300000011920929 -0.180265724658966
0.324999988079071 0.0332649126648903
0.349999994039536 -0.264639049768448
0.375 -0.318326205015182
0.400000035762787 -0.131667047739029
0.425000011920929 -0.304049879312515
0.450000017881393 -0.0757898241281509
0.474999994039536 -0.175909876823425
0.5 -0.0242453664541245
0.524999976158142 -0.0731267109513283
0.550000011920929 -0.278208315372467
0.575000047683716 -0.0516375079751015
0.600000023841858 0.0908294171094894
0.625 -0.0743347480893135
0.649999976158142 -0.0772670730948448
0.675000011920929 -0.0893141776323318
0.699999988079071 -0.113172799348831
0.725000023841858 0.175964429974556
0.75 -0.0628472417593002
0.774999976158142 -0.044495165348053
0.800000071525574 -0.0954918563365936
0.825000047683716 0.00723355263471603
0.850000023841858 -0.145052462816238
0.875000059604645 -0.0800334960222244
0.900000035762787 0.19461415708065
0.925000011920929 0.250030338764191
0.949999988079071 -0.0998282805085182
0.975000023841858 -0.177886828780174
1 0.0450027100741863
1.02499997615814 -0.0547232925891876
1.04999995231628 0.0615614913403988
1.07500004768372 0.164167672395706
1.10000002384186 -0.0708993598818779
1.125 -0.230964571237564
1.15000009536743 -0.0141178034245968
1.17500007152557 0.222333490848541
1.20000004768372 -0.0756463035941124
1.22500002384186 0.252258837223053
1.25 0.0923311486840248
1.27500009536743 -0.125776216387749
1.29999995231628 -0.10867315530777
1.32500004768372 -0.15740954875946
1.35000002384186 -0.0187034532427788
1.375 -0.00278232991695404
1.39999997615814 -0.101434767246246
1.42500007152557 -0.0828027129173279
1.45000004768372 0.122021056711674
1.47500002384186 0.264373332262039
1.5 0.312158703804016
1.52499997615814 0.348185002803802
1.54999995231628 -0.180797308683395
1.57499992847443 0.247303754091263
1.59999990463257 0.402398288249969
1.62499988079071 0.033596333116293
1.65000009536743 0.196172416210175
1.67499995231628 -0.122513487935066
1.70000004768372 0.0342631414532661
1.72499990463257 0.022512074559927
1.75 0.200467243790627
1.77499985694885 0.0437218770384789
1.79999995231628 -0.105816155672073
1.82500004768372 -0.188050910830498
1.85000002384186 0.214035972952843
1.875 -0.171650514006615
1.89999997615814 -0.0119154341518879
1.92499995231628 -0.11694923043251
1.94999992847443 0.0600371770560741
1.97499990463257 -0.335512608289719
2 -0.0879189968109131
2.02500009536743 -0.260032534599304
2.04999995231628 -0.0926248282194138
2.07500004768372 -0.239701211452484
2.09999990463257 0.080620214343071
2.125 -0.121564075350761
2.14999985694885 -0.200891211628914
2.17499995231628 0.030279628932476
2.20000004768372 0.16723096370697
2.22499990463257 -0.141702994704247
2.25 -0.0130326375365257
2.27500009536743 0.123909443616867
2.29999995231628 -0.0709145218133926
2.32499980926514 -0.395591557025909
2.34999990463257 -0.291295826435089
2.375 -0.145825833082199
2.40000009536743 -0.221105992794037
2.42499995231628 -0.0912398844957352
2.45000004768372 -0.249355316162109
2.47499990463257 -0.330182701349258
2.5 -0.211284175515175
2.52499985694885 -0.0541482418775558
2.55000019073486 -0.284439712762833
2.57500004768372 -0.195837095379829
2.59999990463257 -0.157006978988647
2.625 -0.120034337043762
2.65000009536743 -0.298808604478836
2.67499995231628 -0.204703688621521
2.69999980926514 -0.148429602384567
2.72499990463257 -0.388559401035309
2.75 -0.238040640950203
2.77500009536743 -0.121494948863983
2.79999995231628 -0.416199564933777
2.82500004768372 -0.321691453456879
2.84999990463257 -0.279848992824554
2.875 -0.148517996072769
2.89999985694885 -0.03905189037323
2.92500019073486 -0.0450767576694489
2.95000004768372 -0.157088220119476
2.97499990463257 -0.173583820462227
3 -0.0461886376142502
};
\addlegendentry{Mean + Noise}
\end{axis}

\end{tikzpicture}}%
        \caption{Step-based RL}
        \label{subfig:step_based_method}
    \end{subfigure}
    \hfill%
    \begin{subfigure}{0.3\textwidth}        
        \resizebox{0.9\textwidth}{!}{\input{figure/related_works/bbrl}}%
        \caption{Traj.-based, Episodic RL}
        \label{subfig:traj_based_method}
    \end{subfigure}
    \hfill
    \begin{subfigure}{0.3\textwidth}        
        \resizebox{0.9\textwidth}{!}{\input{figure/related_works/tcp}}%
        \caption{\tcp (ours)}
        \label{subfig:tcp_rl}
    \end{subfigure}
    \hspace*{\fill}%
    \caption{Illustration of exploration strategies: (a) SRL samples actions by adding noise to the predicted mean, resulting in inconsistent exploration and jerky actions. However, their leverage of step-based information leads to efficient policy updates. (b) ERL samples complete trajectories in a parameter space and generate consistent control signals. Yet, they often treat trajectories as single data points and overlook the step-based information during the interaction, causing inefficient policy update. (c) \tcp combines the benefits of both, using per-step information for policy update while sampling complete trajectories with broader exploration and high smoothness.
    }
    \label{fig:exploration_strategy}
\end{figure}

By considering how policies interact with the environment, reinforcement learning (RL) methodologies can be classified into two distinct categories: step-based RL (SRL) and episodic RL (ERL). 
SRL predicts actions for each perceived state, while ERL selects an entire behavioral sequence at the start of an episode. Most predominant deep RL methods, such as PPO \citep{schulman2017proximal} and SAC \citep{haarnoja2018soft}, fall into the category of SRL. In these methods, the step information — comprising state, action, reward, subsequent state, and done signal received by the RL agent at each discrete time step — is pivotal for policy updates. This granular data aids in estimating the policy gradient \citep{williams1992simple, sutton1999policy}, approximating state or state-action value functions \citep{haarnoja2018soft}, and assessing advantages \citep{schulman2015high}. 
Although SRL methods have achieved great success in various domains, they often face significant exploration challenges. Exploration in SRL, often based on a stochastic policy like a factorized Gaussian, typically lacks temporal and cross-DoF (degrees of freedom) correlations. This deficiency leads to inconsistent and inefficient exploration across state and action spaces \citep{raffin2022smooth, schumacher2023:deprl}, as shown in \fig\ref{subfig:step_based_method}. Furthermore, the high variance in trajectories generated through such exploration can cause suboptimal convergence and training instability, a phenomenon highlighted by considerable performance differences across various random seeds \citep{agarwal2021deep}.

\textbf{Episodic RL}, in contrast to SRL, represents a distinct branch of RL that emphasizes the maximization of returns over entire episodes \citep{whitley1993genetic,igel2003neuroevolution,peters2008reinforcement}, rather than focusing on the internal evolution of the environment interaction.
This approach shifts the solution search from per-step actions to a parameterized trajectory space, employing techniques like \MP \citep{schaal2006dynamic, paraschos2013probabilistic}.
Such exploration strategy allows for broader exploration horizons and ensures consistent trajectory smoothness across task episodes, as illustrated in \fig\ref{subfig:traj_based_method}. 
Additionally, it is theoretically capable of capturing temporal correlations and interdependencies among \dof. 
ERL typically treats entire trajectories as single data points, often overlooking the internal changes in the environment and state transitions. This approach leads to training predominantly using black-box optimization methods \citep{salimans2017evolution, tangkaratt2017policy, celik2022specializing, otto2023deep}. The term \emph{black box} in our title reflects this reliance on black-box optimization, which tends to overlook detailed step-based information acquired during environmental interactions. However, this often results in a lack of attention to the individual contributions of each segment of the trajectory to the overall task success. Consequently, while ERL excels in expansive exploration and maintaining trajectory smoothness, it typically requires a larger volume of samples for effective policy training. In contrast, step-based RL methods have demonstrated notable advancements in learning efficiency by utilizing this detailed step-based information.

\\[-0.7em]\textbf{Open the Black Box.}
In this paper, our goal is to integrate step-based information into the policy update process of ERL. Our proposed method, \TCE, moves beyond the traditional approach of treating an entire trajectory as a single data point. Instead, we transform trajectory-wide elements, such as reproducing likelihood and advantage, into their segment-wise counterparts. This enables us to leverage the step-based information to recognize and accentuate the unique contributions of each trajectory segment to overall task success. Through this innovative approach, we have opened the black box of ERL, making it more effective while retaining its strength.
As a further step, we explore the benefits of fully-correlated trajectory exploration in deep ERL. 
We demonstrate that leveraging full covariance matrices for trajectory distributions significantly improves policy quality in existing black-box ERL methods like \citet{otto2023deep}.
\\[0.3em]
\textbf{Our contributions} are summarized as:
\begin{enumerate*}[label=(\alph*)]
\item We propose \tcp, a novel RL framework that integrates step-based information into the policy updates of ERL, while preserving the broad exploration scope and trajectory smoothness characteristic of ERL.
\item We provide an in-depth analysis of exploration strategies that effectively capture both temporal and degrees of freedom (DoF) correlations, demonstrating their beneficial impact on policy quality and trajectory smoothness.
\item We conduct a comprehensive evaluation of our approach on multiple simulated robotic manipulation tasks, comparing its performance against other baseline methods.
\end{enumerate*}

\section{Preliminaries}

\subsection{Episodic \Rl}
\paragraph{Markov Decision Process (MDP).} We consider a MDP problem of a policy search defined by a tuple ($\mathcal{S, A, T, R, P}_0, \mathcal{\gamma}$). We assume the state space $\mathcal{S}$ and action space $\mathcal{A}$ are continuous and the transition probabilities $\mathcal{T}: \mathcal{S}\times\mathcal{S}\times\mathcal{A}\rightarrow[0,1]$ describe the state transition probability to $\vs_{t+1}$, given the current state $\vs_t\in\mathcal{S}$ and action $\va_t\in\mathcal{A}$.
The initial state distribution is denoted as $\mathcal{P}_0:\mathcal{S}\rightarrow[0,1]$. The reward $r_t(\vs_t, \va_t)$ returned by the environment is given by a function $\mathcal{R}: \mathcal{S}\times\mathcal{A} \rightarrow \mathbb{R}$ and $\gamma \in [0, 1]$ describes the discount factor. The goal of RL in general is to find a policy $\pi$ that maximizes the expected accumulated reward, namely return, as $R = \mathbb{E}_{\mathcal{T}, \mathcal{P}_0, \pi}[\sum_{t=0}^{\infty}\gamma^t r_t]$.

\textbf{Episodic RL} \citep{whitley1993genetic} focuses on maximizing the return $R=\sum_{t=0}^{T}[\gamma^t r_t]$ over a task episode of length $T$, irrespective of the state transitions within the episode. 
This approach typically employs a parameterized trajectory generator, like \mp \citep{schaal2006dynamic}, to predict a trajectory parameter vector $\vw$. This vector is then used to generate a complete reference trajectory $\vy(\vw) = [y_t]_{t=0:T}$. The resulting trajectory is executed using a trajectory tracking controller to accomplish the task. 
In this context, $y_t \in \mathbb{R}^D$ denotes the trajectory value at time $t$ for a system with $D$ \dof, differentiating it from the per-step action $\va$ used in SRL.
It is important to note that, although ERL predicts an entire action trajectory, it still maintains the \emph{Markov Property}, where the state transition probability depends only on the given current state and action \citep{sutton2018reinforcement}. In this respect, the action selection process in ERL is fundamentally similar to techniques like action repeat \citep{braylan2015frame} and temporally correlated action selection \citep{raffin2022smooth,eberhard2022pink}.
In contrast to SRL, ERL predicts the trajectory parameters as $\pi(\vw|\vs)$, which shifts the solution search from the per-step action space $\mathcal{A}$ to the parameter space $\mathcal{W}$.
Therefore, a trajectory parameterized by a vector $\vw$ is typically treated as a single data point in $\mathcal{W}$. Consequently, ERL commonly employs black-box optimization methods for problem-solving \citep{salimans2017evolution, otto2023deep}.
The general learning objective of ERL is formally expressed as
\begin{equation}
    J = \int \pi_\vtheta(\vw|\vs) [R(\vs, \vw) - V^\pi(\vs)]d \vw= \E_{\vw\sim\pi_\vtheta(\vw|\vs)} [A(\vs, \vw)], \label{eq:traj_rl}
\end{equation}
where $\pi_\vtheta$ represents the policy, parameterized by $\vtheta$, \eg using \nn. 
The initial state $\vs\in \mathcal{S}$ characterizes the starting configuration of the environment and the task goal, serving as the input to the policy. 
The $\pi_\vtheta(\vw|\vs)$ indicates the likelihood of selecting the trajectory parameter $\vw$.
The term $R(\vs, \vw)=\sum_{t=0}^{T}[\gamma^t r_t]$ represents the return obtained from executing the trajectory, while $ V^\pi(\vs) = \E_{\vw\sim\pi_\vtheta(\vw|\vs)} [R(\vs, \vw)]$ denotes the expected return across all possible trajectories under policy $\pi_\vtheta$. 
Their subtraction is defined as the advantage function $A(\vs, \vw)$, which 
quantifies the benefit of selecting a specific trajectory. 
By using parameterized trajectory generators like MPs, ERL benefits from consistent exploration, smooth trajectories, and robustness against local optima, as noted by \citet{otto2023deep}.
However, its policy update strategy incurs a trade-off in terms of learning efficiency, as valuable step-based information is overlooked during policy updates. 
Furthermore, existing method like \citet{bahl2020neural, otto2023deep} commonly use factorized Gaussian policies, which inherently limits their capacity to capture all relevant movement correlations.

\subsection{Using Movement Primitives for Trajectory Representation}
\label{subsec:background_mp}
The Movement Primitives (MP), as a parameterized trajectory generator, play an important role in ERL and robot learning. This section highlights key MP methodologies and their mathematical foundations, deferring a more detailed discussion to Appendix \ref{app:pdmp}.
\citet{schaal2006dynamic} introduced the \DMP method, incorporating a force signal into a dynamical system to produce smooth trajectories from given initial robot states. Following this, \citet{paraschos2013probabilistic} developed \PROMP, which leverages a linear basis function representation to map parameter vectors to trajectories and their corresponding distributions. The probability of observing a trajectory $[y_t]_{t=0:T}$ given a specific weight vector distribution $p(\bm{w}) \sim \mathcal{N}(\bm{w}|\bm{\mu_w}, \bm{\Sigma_w})$ is represented as a linear basis function model:
\begin{align}
    [y_t]_{t=0:T} &= \bm{\Phi}_{0:T}^\intercal \bm{w} + \epsilon_{y}, \label{eq:promp}\\
    p([y_t]_{t=0:T};~\bm{\mu}_{\vy}, \bm{\Sigma}_{\vy}) &= \mathcal{N}( \bm{\Phi}_{0:T}^\intercal\bm{\mu_w}, ~\bm{\Phi}_{0:T}^\intercal \bm{\Sigma_w} \bm{\Phi}_{0:T}~+ \sigma_y^2 \bm{I}).   
    \label{eq:promp_p_of_y}
\end{align}
Here, $\epsilon_{y}$ is zero-mean white noise with variance $\sigma_y^2$. The matrix $\bm{\Phi}_{0:T}$ houses the basis functions for each time step $t$. Additionally, $p([y_t]_{t=0:T};~\bm{\mu}_{\vy}, \bm{\Sigma}_{\vy})$ defines the trajectory distribution coupling the \dof and time steps, mapped from $p(\bm{w})$. 
For a $D$-DoF system with $N$ parameters per DoF and $T$ time steps, the dimensions of the variables in \eqref{eq:promp} and \ref{eq:promp_p_of_y} are as follows: 
$\vw, \bm{\mu}_{\vw}: D\cdot N$;
$\bm{\Sigma}_w: D\cdot N\times D\cdot N$;
$\bm{\Phi}_{0:T}: D\cdot N\times D\cdot T$;
$\vy, \bm{\mu}_{\vy}: D\cdot T$;
$\bm{\Sigma}_{\vy}: D\cdot T\times D\cdot T$.

Recently, \citet{li2023prodmp} introduced \PDMP, a hybrid approach that blends the pros of both methods. Similar to \acrshort{promp}, \pdmp defines a trajectory as $y(t) = \ipb(t)^\intercal \vw + c_1y_1(t) + c_2y_2(t)$. The added terms $c_1y_1(t) + c_2y_2(t)$ are included to ensure accurate trajectory initialization. This formulation combines the distributional modeling benefits of \acrshort{promp} with the precision in trajectory initiation offered by \acrshort{dmp}.

\subsection{Calculation of Trajectory Distribution and Reconstruction Likelihood}
\label{subsec:background_traj_ll}
\def\arrowblue{black!40!blue}
\def\arrowgreen{black!40!green}
\usetikzlibrary{arrows,matrix,positioning,patterns,decorations.pathreplacing,calc}
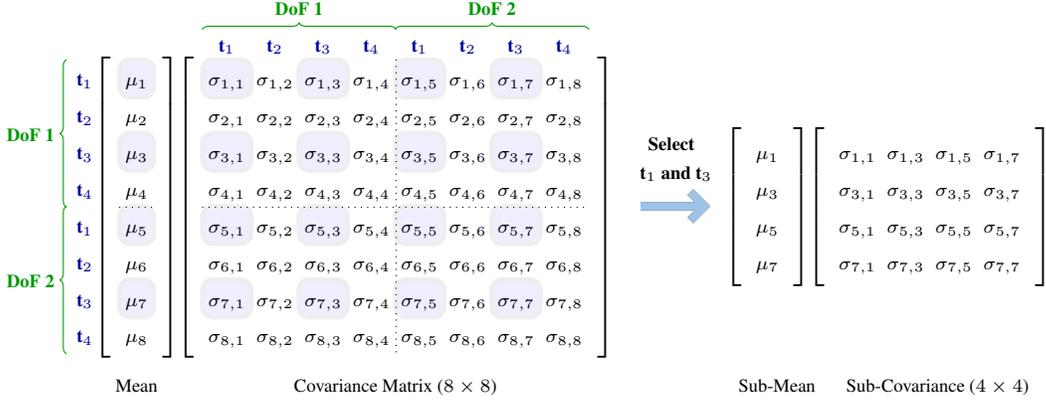
\begin{figure*}[t!]
    \vspace{-0.2cm}
    \centering
    \tikzstyle{text_style}=[font=\scriptsize\bfseries, text height=1.5ex, text depth=.25ex, anchor=center]
    \tikzstyle{text_style2}=[font=\scriptsize, text height=1.5ex, text depth=.25ex, anchor=center]
    \tikzstyle{matrix_style}=[inner xsep=2pt, inner ysep=3pt] %
    \begin{tikzpicture}[baseline,decoration=brace, rrnode_nn/.style={rectangle, draw, minimum size=5mm, rounded corners=1mm, scale=\sca},]
        \node[ ]  (start) at (0,0) {$ $};
        \matrix [matrix of math nodes, left delimiter=[, right delimiter={]},text_style, matrix_style] (mean) [right=1cm of start]
        {
            \mu_{1}\\ 
            \mu_{2}\\ 
            \mu_{3}\\ 
            \mu_{4}\\ 
            \mu_{5}\\ 
            \mu_{6}\\ 
            \mu_{7}\\ 
            \mu_{8}\\         
        };
        \draw [dotted] (mean-4-1.south west) -- (mean-4-1.south east);        

        \node [below=0.3cm of mean, text_style2] {Mean};
        
        \draw[decorate,transform canvas={xshift=-7mm}, color=\arrowgreen, text_style] (mean-4-1.south west) -- node[left=1pt] {DoF 1} (mean-1-1.north west); %
        \draw[decorate,transform canvas={xshift=-7mm}, color=\arrowgreen, text_style] (mean-8-1.south west) -- node[left=1pt] {DoF 2} (mean-5-1.north west); %
        
        \node [left=1.3em of mean-1-1.west, color=\arrowblue, text_style] {t$_1$};
        \node [left=1.3em of mean-2-1.west, color=\arrowblue, text_style] {t$_2$};
        \node [left=1.3em of mean-3-1.west, color=\arrowblue, text_style] {t$_3$};
        \node [left=1.3em of mean-4-1.west, color=\arrowblue, text_style] {t$_4$};
        \node [left=1.3em of mean-5-1.west, color=\arrowblue, text_style] {t$_1$};
        \node [left=1.3em of mean-6-1.west, color=\arrowblue, text_style] {t$_2$};
        \node [left=1.3em of mean-7-1.west, color=\arrowblue, text_style] {t$_3$};
        \node [left=1.3em of mean-8-1.west, color=\arrowblue, text_style] {t$_4$};        

        \draw[rounded corners,ultra thick, draw=black, fill=blue, opacity=0.07] (mean-1-1.south west) rectangle (mean-1-1.north east);             
        \draw[rounded corners,ultra thick, draw=black, fill=blue, opacity=0.07] (mean-3-1.south west) rectangle (mean-3-1.north east);
        \draw[rounded corners,ultra thick, draw=black, fill=blue, opacity=0.07] (mean-5-1.south west) rectangle (mean-5-1.north east);             
        \draw[rounded corners,ultra thick, draw=black, fill=blue, opacity=0.07] (mean-7-1.south west) rectangle (mean-7-1.north east);

        \matrix [matrix of math nodes, left delimiter=[, right delimiter={]}, text_style, matrix_style] (m) [right=0.5cm of mean]
        {
            \sigma_{1,1} & \sigma_{1,2} & \sigma_{1,3} & \sigma_{1,4} & \sigma_{1,5} & \sigma_{1,6}& \sigma_{1,7}& \sigma_{1,8} \\ 
            \sigma_{2,1} & \sigma_{2,2} & \sigma_{2,3} & \sigma_{2,4} & \sigma_{2,5} & \sigma_{2,6}& \sigma_{2,7}& \sigma_{2,8} \\ 
            \sigma_{3,1} & \sigma_{3,2} & \sigma_{3,3} & \sigma_{3,4} & \sigma_{3,5} & \sigma_{3,6}& \sigma_{3,7}& \sigma_{3,8} \\ 
            \sigma_{4,1} & \sigma_{4,2} & \sigma_{4,3} & \sigma_{4,4} & \sigma_{4,5} & \sigma_{4,6}& \sigma_{4,7}& \sigma_{4,8} \\ 
            \sigma_{5,1} & \sigma_{5,2} & \sigma_{5,3} & \sigma_{5,4} & \sigma_{5,5} & \sigma_{5,6}& \sigma_{5,7}& \sigma_{5,8} \\ 
            \sigma_{6,1} & \sigma_{6,2} & \sigma_{6,3} & \sigma_{6,4} & \sigma_{6,5} & \sigma_{6,6}& \sigma_{6,7}& \sigma_{6,8} \\ 
            \sigma_{7,1} & \sigma_{7,2} & \sigma_{7,3} & \sigma_{7,4} & \sigma_{7,5} & \sigma_{7,6}& \sigma_{7,7}& \sigma_{7,8} \\ 
            \sigma_{8,1} & \sigma_{8,2} & \sigma_{8,3} & \sigma_{8,4} & \sigma_{8,5} & \sigma_{8,6}& \sigma_{8,7}& \sigma_{8,8} \\             
        };        

        \node (cov_txt)[below=0.3cm of m, text_style2] {Covariance Matrix ($8\times 8$)};
        
        \draw[rounded corners,ultra thick, draw=black, fill=blue, opacity=0.07] (m-1-1.south west) rectangle (m-1-1.north east);
        \draw[rounded corners,ultra thick, draw=black, fill=blue, opacity=0.07] (m-3-1.south west) rectangle (m-3-1.north east);             
        \draw[rounded corners,ultra thick, draw=black, fill=blue, opacity=0.07] (m-5-1.south west) rectangle (m-5-1.north east);       
        \draw[rounded corners,ultra thick, draw=black, fill=blue, opacity=0.07] (m-7-1.south west) rectangle (m-7-1.north east);              
        \draw[rounded corners,ultra thick, draw=black, fill=blue, opacity=0.07] (m-1-3.south west) rectangle (m-1-3.north east);             
        \draw[rounded corners,ultra thick, draw=black, fill=blue, opacity=0.07] (m-3-3.south west) rectangle (m-3-3.north east);             
        \draw[rounded corners,ultra thick, draw=black, fill=blue, opacity=0.07] (m-5-3.south west) rectangle (m-5-3.north east);             
        \draw[rounded corners,ultra thick, draw=black, fill=blue, opacity=0.07] (m-7-3.south west) rectangle (m-7-3.north east);     
        \draw[rounded corners,ultra thick, draw=black, fill=blue, opacity=0.07] (m-1-5.south west) rectangle (m-1-5.north east);
        \draw[rounded corners,ultra thick, draw=black, fill=blue, opacity=0.07] (m-3-5.south west) rectangle (m-3-5.north east);             
        \draw[rounded corners,ultra thick, draw=black, fill=blue, opacity=0.07] (m-5-5.south west) rectangle (m-5-5.north east);       
        \draw[rounded corners,ultra thick, draw=black, fill=blue, opacity=0.07] (m-7-5.south west) rectangle (m-7-5.north east);              
        \draw[rounded corners,ultra thick, draw=black, fill=blue, opacity=0.07] (m-1-7.south west) rectangle (m-1-7.north east);             
        \draw[rounded corners,ultra thick, draw=black, fill=blue, opacity=0.07] (m-3-7.south west) rectangle (m-3-7.north east);             
        \draw[rounded corners,ultra thick, draw=black, fill=blue, opacity=0.07] (m-5-7.south west) rectangle (m-5-7.north east);             
        \draw[rounded corners,ultra thick, draw=black, fill=blue, opacity=0.07] (m-7-7.south west) rectangle (m-7-7.north east);

        \draw [dotted] (m-4-1.south west) -- (m-4-8.south east);        
        \draw [dotted] (m-1-4.north east) -- (m-8-4.south east);

        \draw[decorate,transform canvas={yshift=4mm}, color=\arrowgreen, text_style] (m-1-1.north west) -- node[above=1pt] {DoF 1} (m-1-4.north east); %
        \draw[decorate,transform canvas={yshift=4mm}, color=\arrowgreen, text_style] (m-1-5.north west) -- node[above=1pt] {DoF 2} (m-1-8.north east); %
        
        \node [above=0.6em of m-1-1.north, color=\arrowblue, text_style] {t$_1$};
        \node [above=0.6em of m-1-2.north, color=\arrowblue, text_style] {t$_2$};
        \node [above=0.6em of m-1-3.north, color=\arrowblue, text_style] {t$_3$};
        \node [above=0.6em of m-1-4.north, color=\arrowblue, text_style] {t$_4$};
        \node [above=0.6em of m-1-5.north, color=\arrowblue, text_style] {t$_1$};
        \node [above=0.6em of m-1-6.north, color=\arrowblue, text_style] {t$_2$};
        \node [above=0.6em of m-1-7.north, color=\arrowblue, text_style] {t$_3$};
        \node [above=0.6em of m-1-8.north, color=\arrowblue, text_style] {t$_4$};        
        
        \matrix [matrix of math nodes, left delimiter=[, right delimiter={]},text_style, matrix_style] (mean_small) [right=2cm of m.east]
        {
            \mu_{1}\\ 
            \mu_{3}\\ 
            \mu_{5}\\ 
            \mu_{7}\\ 
        };
        \node (sub_mean)[right=3.6cm of cov_txt, text_style2] {Sub-Mean};
        
        \matrix [matrix of math nodes, left delimiter=[, right delimiter={]}, text_style, matrix_style] (m_small) [right=0.5cm of mean_small.east ]
        {
            \sigma_{1,1} & \sigma_{1,3} & \sigma_{1,5} & \sigma_{1,7} \\ 
            \sigma_{3,1} & \sigma_{3,3} & \sigma_{3,5} & \sigma_{3,7} \\ 
            \sigma_{5,1} & \sigma_{5,3} & \sigma_{5,5} & \sigma_{5,7} \\ 
            \sigma_{7,1} & \sigma_{7,3} & \sigma_{7,5} & \sigma_{7,7} \\ 
        };

        \node (sub_cov)[right=1.5cm of sub_mean, text_style2] {Sub-Covariance ($4\times4$)};
        \node (arrow) [right=0.5cm of m] {\includegraphics[scale=0.06]{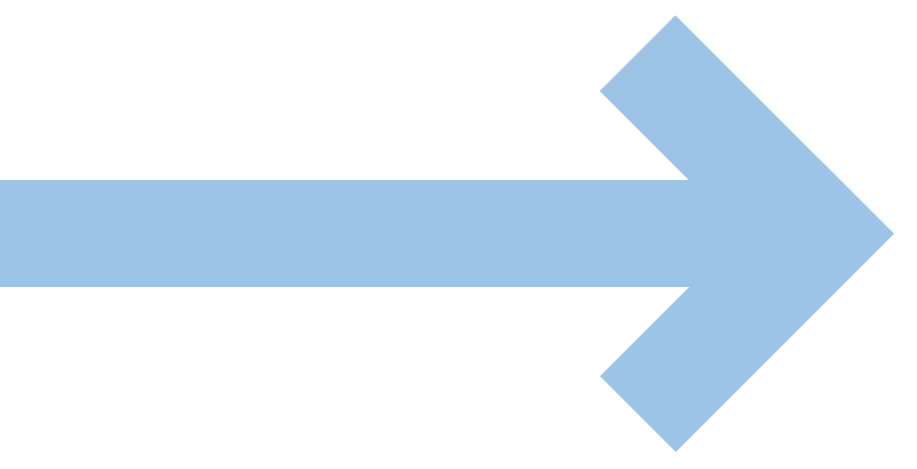}};        
        \matrix (t1t3) [above=0.3cm of arrow, matrix of nodes, nodes in empty cells, column sep=-\pgflinewidth, row sep=1.5\pgflinewidth, text_style, inner ysep=1.2pt] {Select~~~\\t$_1$ and t$_3$\\};  
        
    \end{tikzpicture}    
    \vspace{-0.2cm}
    \caption{Reduce the trajectory distribution dimensions using two time steps \citep{li2023prodmp}, shown in an element-wise format. Here, the trajectory has two \dof and four time steps, with $D\cdot T=8$. \textbf{Left}: The $8$-dim mean vector and the $8\times8$-dim covariance matrix of the original trajectory distribution, capture correlations across both \dof and time steps. 
    \textbf{Right}: Randomly selecting two time points, \eg t$_1$ and t$_3$, yields a reduced distribution while still capturing the movement correlations. 
    }
    \label{fig:ll_est}
\end{figure*}

Computing the trajectory distribution and reconstruction likelihood is crucial for policy updates in ERL.
Previous methods like \citet{bahl2020neural,otto2023deep} represented the trajectory distribution using the parameter distribution $p(\vw)$ and the likelihood of a sampled trajectory $\vy^*$ with its parameter vector as $p(\vw^*|\vmu_{\vw}, \sigma_{\vw}^2)$. 
However, this approach treats an entire trajectory as a singular data point and fails to efficiently utilize step-based information.
In contrast, research in imitation learning, including works by \citet{paraschos2013probabilistic,gomez2016using}, maps parameter distributions to trajectory space and allows the exploitation of trajectory-specific information. 
Yet, the likelihood computation in this space is computationally intensive, primarily due to the need to invert a high-dimensional covariance matrix, a process with an {\footnotesize $O((D\cdot T)^3)$} time complexity.
Recent studies, like those by \citep{seker2019conditional, akbulut2021acnmp, przystupa2023deep}, advocates for directly modeling the trajectory distribution using neural networks.
These methods typically employ a factorized Gaussian distribution $\mathcal{N}(y| \mu_{y}, \sigma^2_{y})$, instead of a full Gaussian distribution $\mathcal{N}(\vy| \vmu_{\vy}, \mSigma_{\vy})$ that accounts for both the \dof and time steps. 
This choice mitigates the computational burden of likelihood calculations, but comes at the cost of neglecting key temporal correlations and interactions between different \dof.
To address these challenges, \citet{li2023prodmp} introduced a novel approach for estimating the trajectory likelihood with a set of paired time points ${(t_k, t_k')}, k=1,...,K$, as
\begin{equation}    
    \log p([\vy_t]_{t=0:T}) \approx \frac{1}{K}\sum_{k=1}^K\log \mathcal{N}(\bm{y}_{(t_k, t_k')} | \bm{\mu}_{(t_k, t_k')}, \bm{\Sigma}_{(t_k, t_k')}),
    \label{eq:rec_ll}
\end{equation}
As shown in \fig\ref{fig:ll_est}, this method scales down the dimensions of a trajectory distribution from $D\cdot T$ to a more manageable $D\cdot 2$. Through the use of batched, randomly selected time pairs during training, the method is proved to efficiently capture correlations while reducing computational cost. 

\subsection{Using Trust Regions for stable policy update}
\label{subsec:TRPL}
In ERL, the parameter space $\mathcal{W}$ typically exhibits higher dimensionality compared to the action space $\mathcal{A}$. This complexity presents unique challenges in maintaining stable policy updates. 
Trust regions methods \citep{schulman2015trust, schulman2017proximal} has long been recognized as an effective technique for ensuring the stability and convergence of policy gradient methods.
While popular methods such as PPO approximate trust regions using surrogate cost functions, they lack the capacity for exact enforcement.
To tackle this issue, \citet{otto2021differentiable} introduced \TRPL, a mathematically rigorous and scalable technique that precisely enforces trust regions in deep \rl algorithms.
By incorporating differentiable convex optimization layers \citep{agrawal2019differentiable}, this method not only allows for trust region enforcement for each input state, but also demonstrates significant effectiveness and stability in high-dim parameter space, as validated in method like BBRL \cite{otto2023deep}.
The \trpl takes standard outputs of a Gaussian policy—namely, the mean vector $\bm{\mu}$ and covariance matrix $\bm{\Sigma}$ —and applies a state-specific projection operation to maintain trust regions. 
The adjusted Gaussian policy, parameterized by $\tilde{\bm{\mu}}$ and $\tilde{\bm{\Sigma}}$, forms the basis for subsequent computations. Let $d_\textrm{mean}$ and $d_\textrm{cov}$ be the dissimilarity measures, \eg KL-divergence, for mean and covariance, bounded by $\epsilon_\mu$ and $\epsilon_\Sigma$ respectively. The optimization for each state $\vs$ is formulated as:
\begin{equation}       
    \vspace{-0.03cm}
    \begin{aligned}
    \argmin_{\tilde{\bm{\mu}}_s} d_\textrm{mean} \left(\tilde{\bm{\mu}}_s, \bm{\mu}(\vs) \right),  \quad &\st \quad d_\textrm{mean} \left(\tilde{\bm{\mu}}_s,  \old{\bm{\mu}}(\vs) \right) \leq \epsilon_{\mu}, \textrm{ and}\\
    \argmin_{\tilde{\bm{\Sigma}}_s} d_\textrm{cov} \left(\tilde{\bm{\Sigma}}_s, \bm{\Sigma}(\vs) \right), \quad &\st \quad d_\textrm{cov} \left(\tilde{\bm{\Sigma}}_s, \old{\bm{\Sigma}}(\vs) \right) \leq \epsilon_\Sigma.   
    \end{aligned}
    \label{eq:trust_region}
    \vspace{-0.03cm}
\end{equation}

\section{Use Step-based Information for ERL Policy Updates}

\begin{figure}[t!]
    \newcommand{\arrowscale}{0.05}
    \newcommand{\obsscale}{0.16}     
    \newcommand{\nnscale}{0.05}
    \newcommand{\mathscale}{0.1}
    \newcommand{\fsize}{\footnotesize}
    \newcommand{\vbelow}{0.2cm}
    \definecolor{lightgreen}{RGB}{144,238,144}
    \definecolor{darkgreen}{RGB}{0,128,0}
    \definecolor{lightblue}{RGB}{173,216,230}
    \definecolor{gray}{RGB}{128,128,128}
    \tikzstyle{aligned text}=[font=\scriptsize\bfseries, text height=1.5ex, text depth=.25ex, anchor=center]
    \tikzstyle{aligned text2}=[font=\scriptsize, text height=1.5ex, text depth=.25ex, anchor=center]
    
    \centering
    \begin{tikzpicture}

        \node (obs) at (0,0) {\includegraphics[scale=\obsscale]{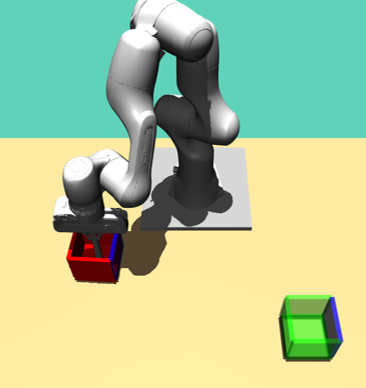}};
        \node (obs_arrow) [above right=-0.9cm and -0.01cm of obs] {\includegraphics[scale=\arrowscale]        
        {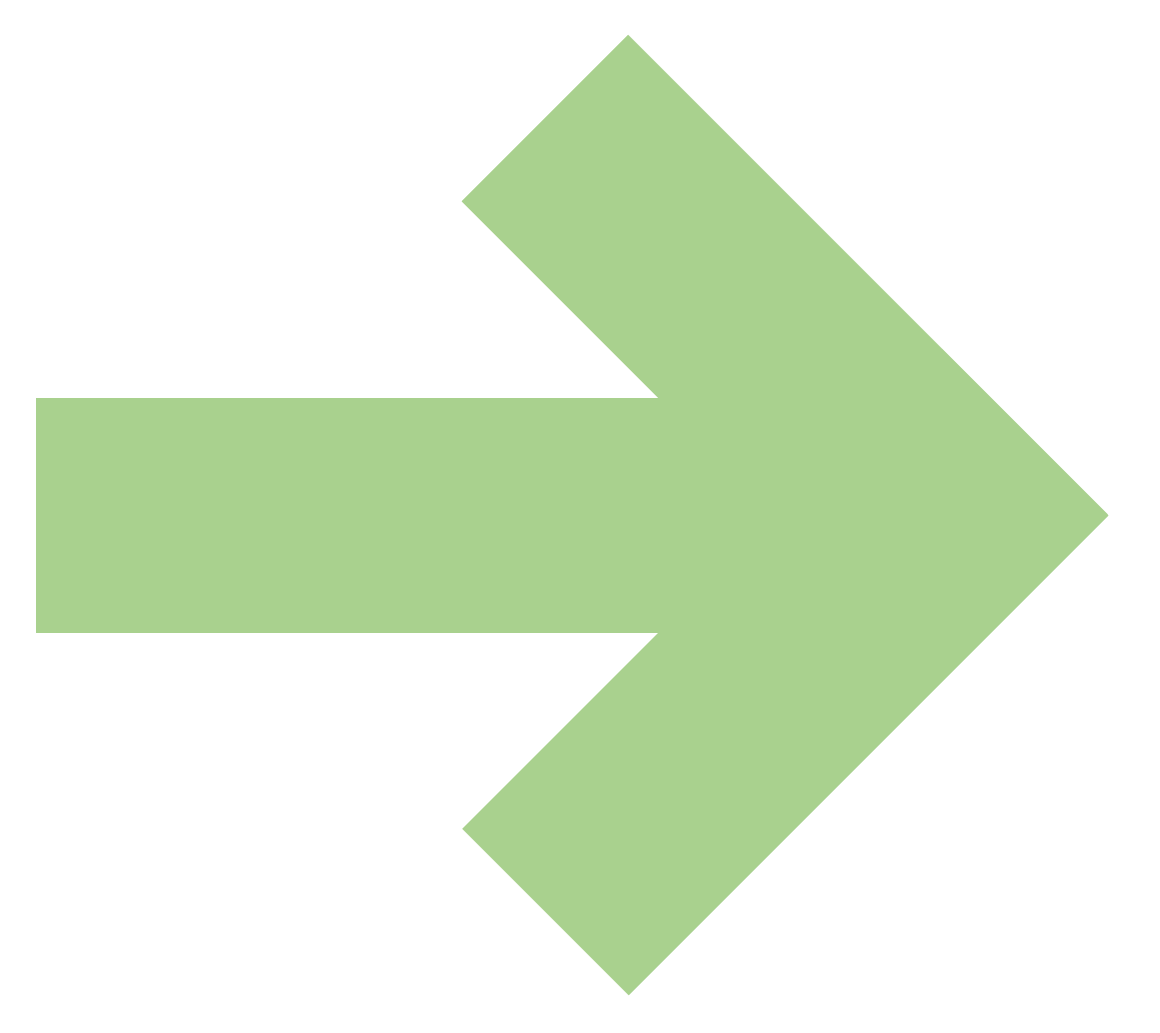}};
        \node (obs_txt)[above=\vbelow of obs, aligned text] {Observation};
        \node (obs_txt2)[above=0.17cm of obs_txt, aligned text] {Task};
        
        \node (policy_box)[rectangle, anchor=north west, draw=gray, fill=lightblue, fill opacity=0.1, minimum width=6.4cm, minimum height=2cm, rounded corners=5pt, line width=0.5mm] at (1.5,1.8) {};
        \node (policy_box_txt)[above=-0.25cm of policy_box, aligned text] {RL Policy {\footnotesize$\pi_{\vtheta}$} for Trajectory Parameter Selection};
        
        \node (nn) [right=0.0cm of obs_arrow] {\includegraphics[scale=\nnscale]{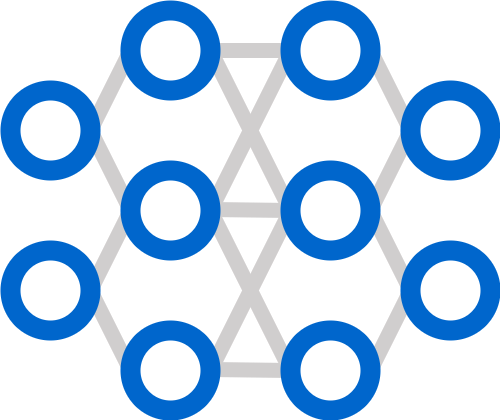}};
        \node (nn_arrow) [right=0.0cm of nn] {\includegraphics[scale=\arrowscale]{figure/elements/arrow/green.png}};
        \node (nn_txt)[right=1.45cm of obs_txt, aligned text2] {Policy Nets};
        
        \matrix (pred) [right=-0.2cm of nn_arrow, matrix of nodes, nodes in empty cells, column sep=-\pgflinewidth, row sep=1.5\pgflinewidth] {
                    $\vmu_{\vw}$\\
                    $\mSigma_{\vw}$\\
                };        
        \node (pred_arrow) [right=-0.12cm of pred] {\includegraphics[scale=\arrowscale]{figure/elements/arrow/green.png}};
        \node (pred_txt)[right=3.cm of obs_txt, aligned text2] {Predict};
        
        \node (sample)[right=-0.1cm of pred_arrow, font=\fsize] {$~~\vw^* \sim \mathcal{N}(\vmu_{\vw}, \mSigma_{\vw})$};
        \node (sample_arrow) [below right=-1.6cm and 7.1cm of obs] {\includegraphics[scale=\arrowscale]{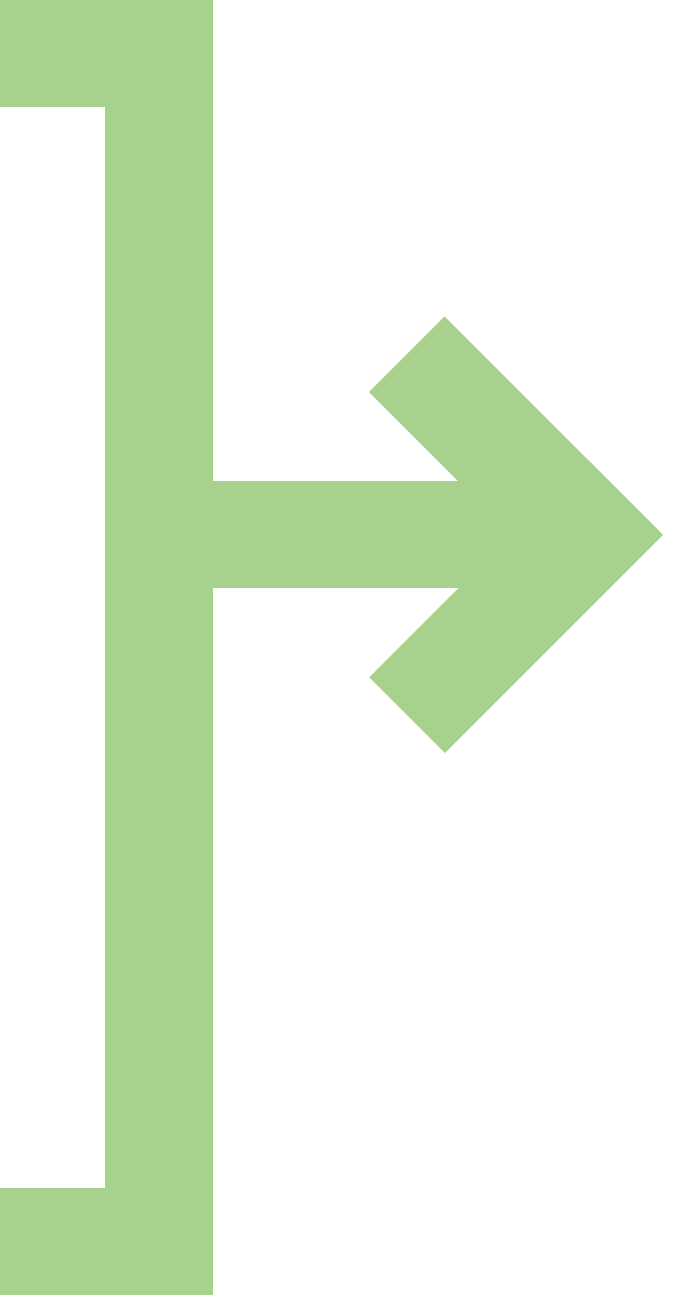}};
        \node (sample_txt)[right=5.45cm of obs_txt, aligned text2] {Sample Traj. Parameter};

        \node (obs_arrow2_new) [below right=-0.7cm and -0.03cm of obs] {\includegraphics[scale=\arrowscale]    {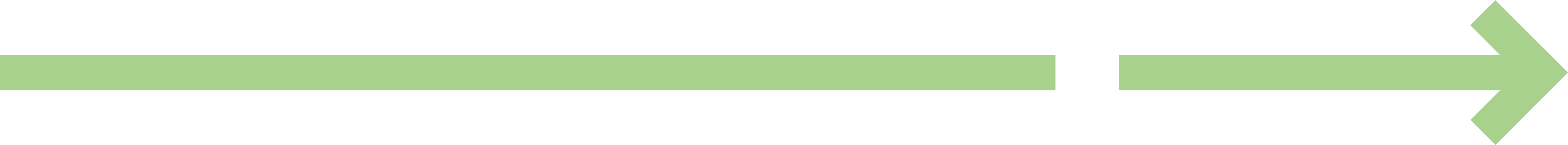}};
        
        \node (init_cond)[right=-0.1cm of obs_arrow2_new, font=\fsize] {{\scriptsize Robot Initial State} $\vy_b, \dot{\vy}_b$};        

        \node (prodmp_box)[rectangle, anchor=north west, draw=gray, fill=lightblue, fill opacity=0.1, minimum width=1.9cm, minimum height=2cm, rounded corners=5pt, line width=0.5mm] at (8.8,1.8) {};
        \node (prodmp_box_txt)[right=2.32cm of policy_box_txt, aligned text] {\pdmp};
        \node (traj_txt)[right=9cm of obs_txt, aligned text2] {Compute Traj.};
        \matrix (traj) [below=-0.2cm of traj_txt, matrix of nodes, nodes in empty cells, column sep=-\pgflinewidth, row sep=1.5\pgflinewidth, font=\fsize] {
                    $[\vy_t]_{t=0:T}$\\
                    $[\dot{\vy}_t]_{t=0:T}$\\
                };        
        \node (traj_arrow) [below right=-0.35cm and 5.8cm of pred_arrow] {\includegraphics[scale=\arrowscale]{figure/elements/arrow/green.png}};

        \node (exe) [right=10.4cm of obs] {\includegraphics[scale=\obsscale]{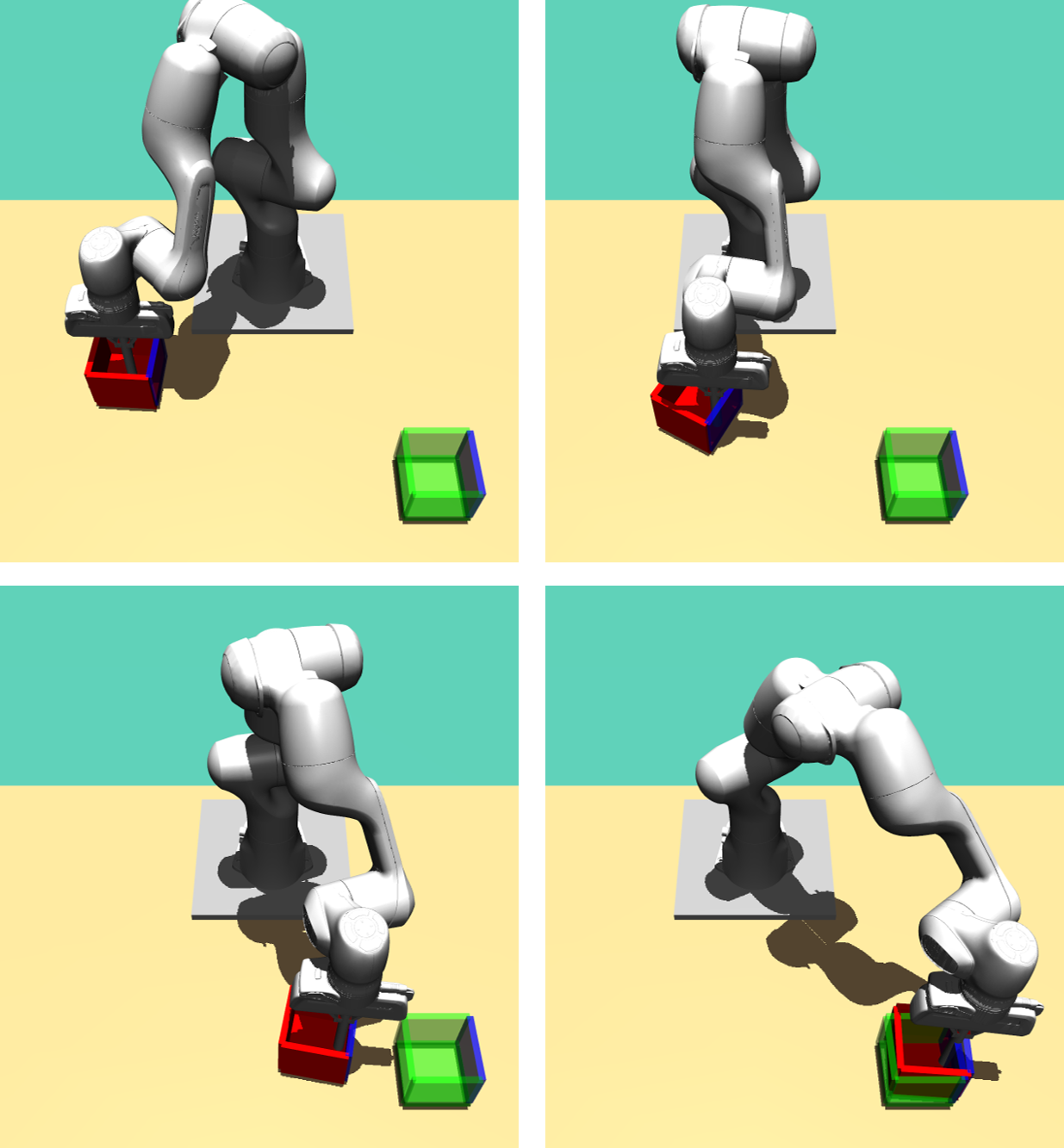}};
        \node (exe_txt1)[right=1.8cm of prodmp_box_txt, aligned text] {Environment};
        \node (exe_txt)[right=11.45cm of obs_txt, aligned text] {Interaction};

        \node (s_and_r_arrow) [below=0cm of exe] {\includegraphics[scale=\arrowscale]{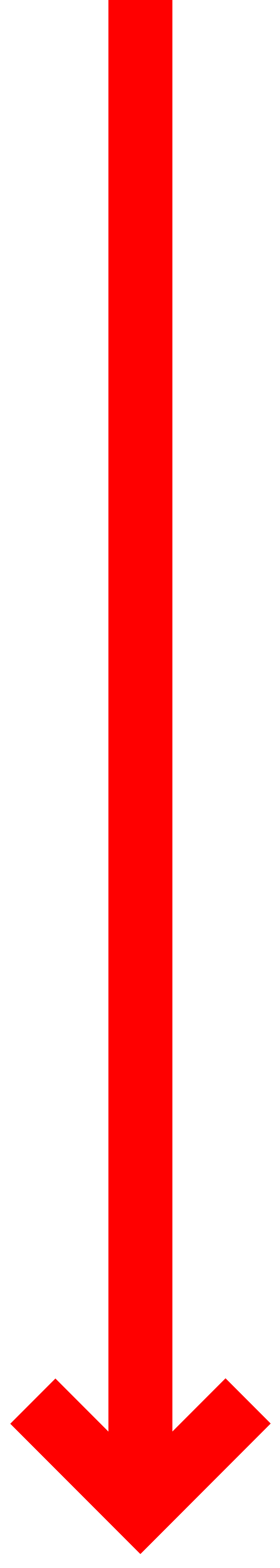}};
        \node (s_and_r_txt)[below=0.1cm of s_and_r_arrow, aligned text] {Segment states};
        \node (s_and_r_txt2)[below=0.17cm of s_and_r_txt, aligned text] {and rewards};
        \matrix (s_and_r) [below=0cm of s_and_r_txt2, matrix of nodes, nodes in empty cells, column sep=-\pgflinewidth, row sep=1.5\pgflinewidth, font=\fsize] {
                    $\vs_{t_k}, \vs_{t_k'}$\\ \\
                    $[r_t]_{t=t_k:t_k'}$\\
                };

        \node (adv_box)[rectangle, anchor=north east, draw=gray, fill=lightblue, fill opacity=0.1, minimum width=4.95cm, minimum height=2.55cm, rounded corners=5pt, line width=0.5mm] at (10.55, -3.25) {};
        \node (adv_box_txt)[left=3.2cm of s_and_r_txt, aligned text] {Segment Evaluation};

        \node (critic_in_arrow) [above left=-0.65cm and -0.1cm of s_and_r] {\includegraphics[scale=\arrowscale]{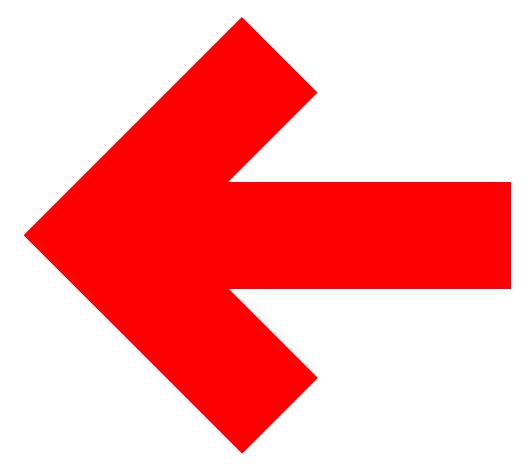}};
        \node (nn_critic) [left=0.5cm of critic_in_arrow] {\includegraphics[scale=\nnscale]{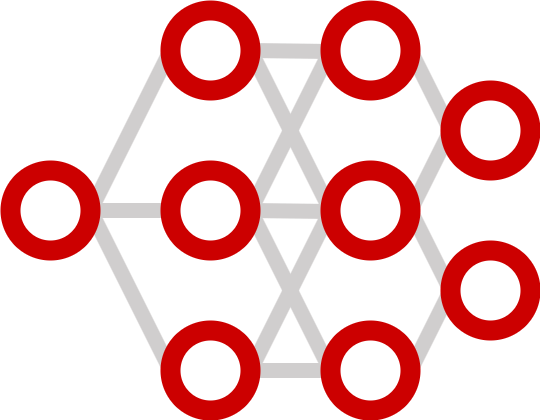}};
        \node (nn_critic_txt)[left=1.8cm of s_and_r_txt2, aligned text2] {V-func. Net};
        \node (critic_out_arrow) [left=0.2cm of nn_critic] {\includegraphics[scale=\arrowscale]{figure/elements/arrow/red_left.png}};
                    
        \node (ac_reward_arrow) [below left=-0.83 and -0.1cm of s_and_r] {\includegraphics[scale=\arrowscale]{figure/elements/arrow/red_left.png}};
        \node (ac_reward_txt)[left=0.4cm of ac_reward_arrow, font=\fsize] {{\scriptsize Segment return} $R_k = \sum_{t=t_k}^{t_k'-1}\gamma^{t-t_k}r_t$};

        \node (Value_txt)[below left=-0.6cm and 0.1cm of critic_out_arrow, font=\fsize] {$V(\vs_{t_k}), V(\vs_{t_k'})$};
        \node (seg_v_txt)[left=1.95cm of nn_critic_txt, aligned text2] {Segment Values};

        \node (prodmp_ll_box)[rectangle, anchor=north west, draw=gray, fill=lightblue, fill opacity=0.1, minimum width=2.37cm, minimum height=1.7cm, rounded corners=5pt, line width=0.5mm] at (5.4,-0.9) {};
        \node (prodmp_ll_box_txt)[above=-0.25cm of prodmp_ll_box, aligned text] {\pdmp};
        \node (prodmp_ll_box_txt2)[below=0.1cm of prodmp_ll_box_txt, aligned text2] {Compute Segment};
        \node (prodmp_ll_box_txt3)[below=0.cm of prodmp_ll_box_txt2, aligned text2] {Mean and Cov.};
        \node (ll_input)[below=-0.1cm of prodmp_ll_box_txt3, font=\fsize] {$\bm{\mu}_{(t_k, t_k')}, \bm{\Sigma}_{(t_k, t_k')}$};
 
        \node (w_in_arrow) [below right=-0.15cm and -0.85cm of pred] {\includegraphics[scale=\arrowscale]{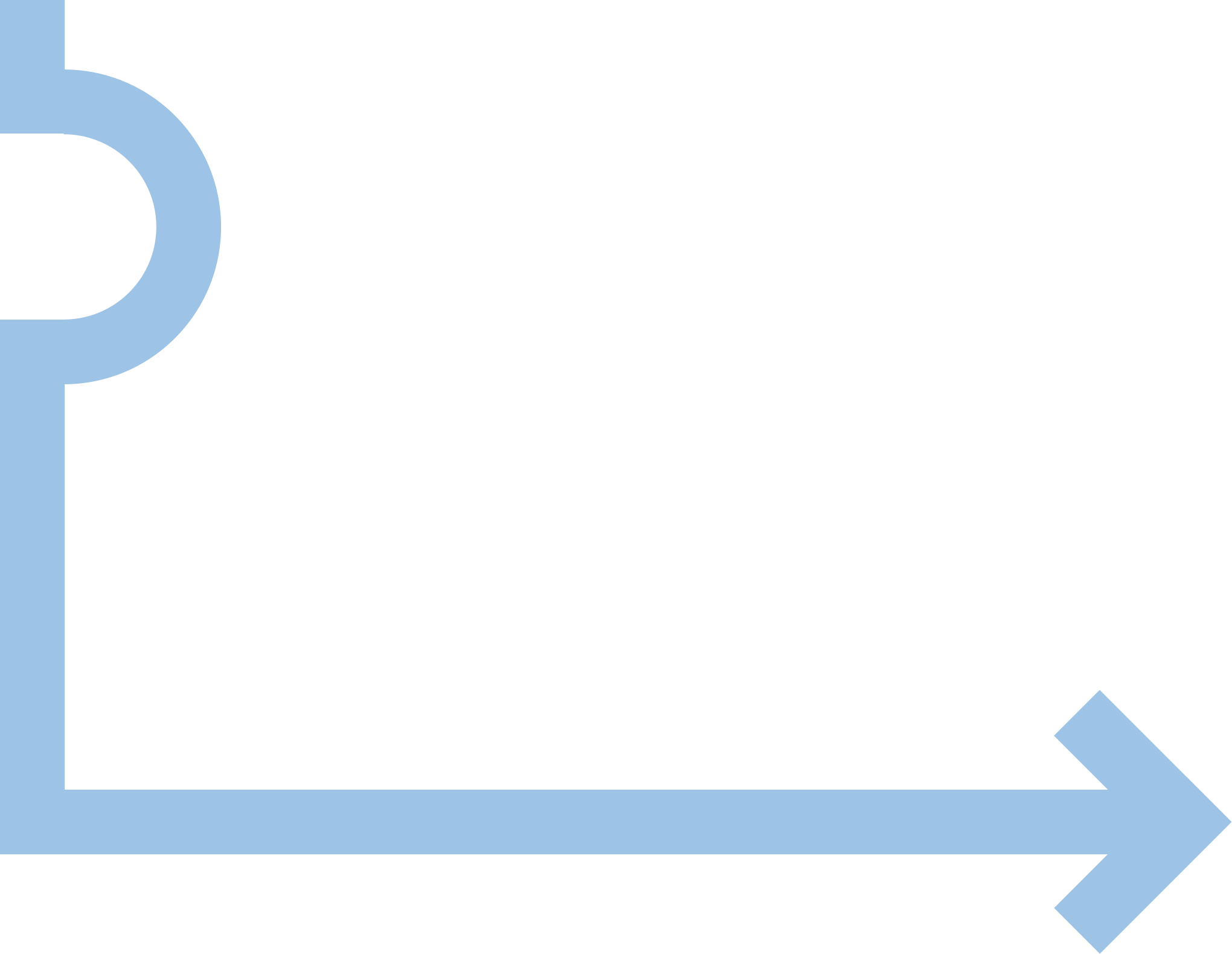}};
        \node (init_cond_in_arrow) [below right=-0.3cm and -0.23cm of init_cond] {\includegraphics[scale=\arrowscale]{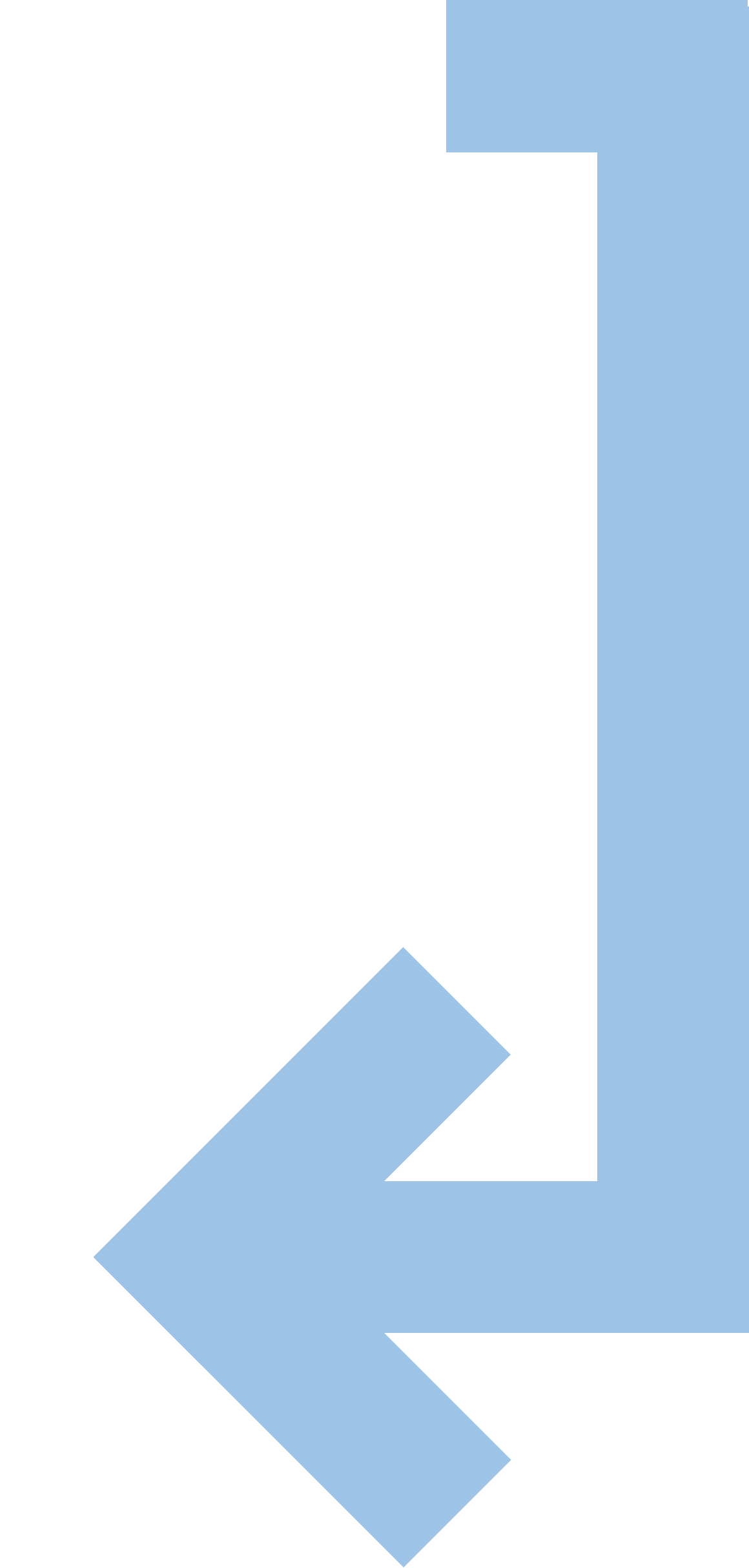}};

        \node (segment_in_arrow) [below=-0.11cm of prodmp_box] {\includegraphics[scale=\arrowscale]{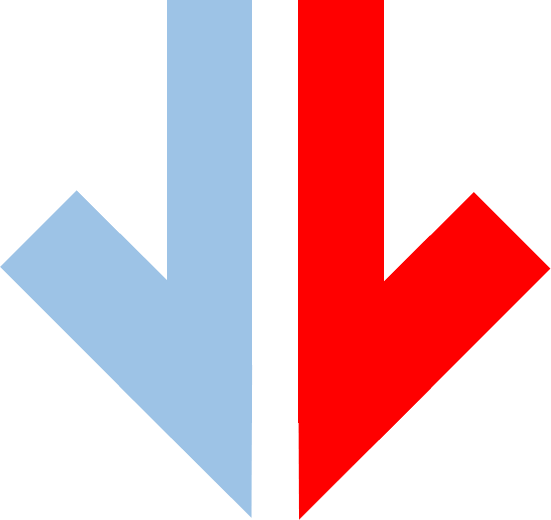}};

        \node (segment_txt) [below=-0.0cm of segment_in_arrow, aligned text] {Traj. to Segments};
        
        \node (segment_plot) [below=-0.15cm of segment_txt] {\includegraphics[scale=0.09]{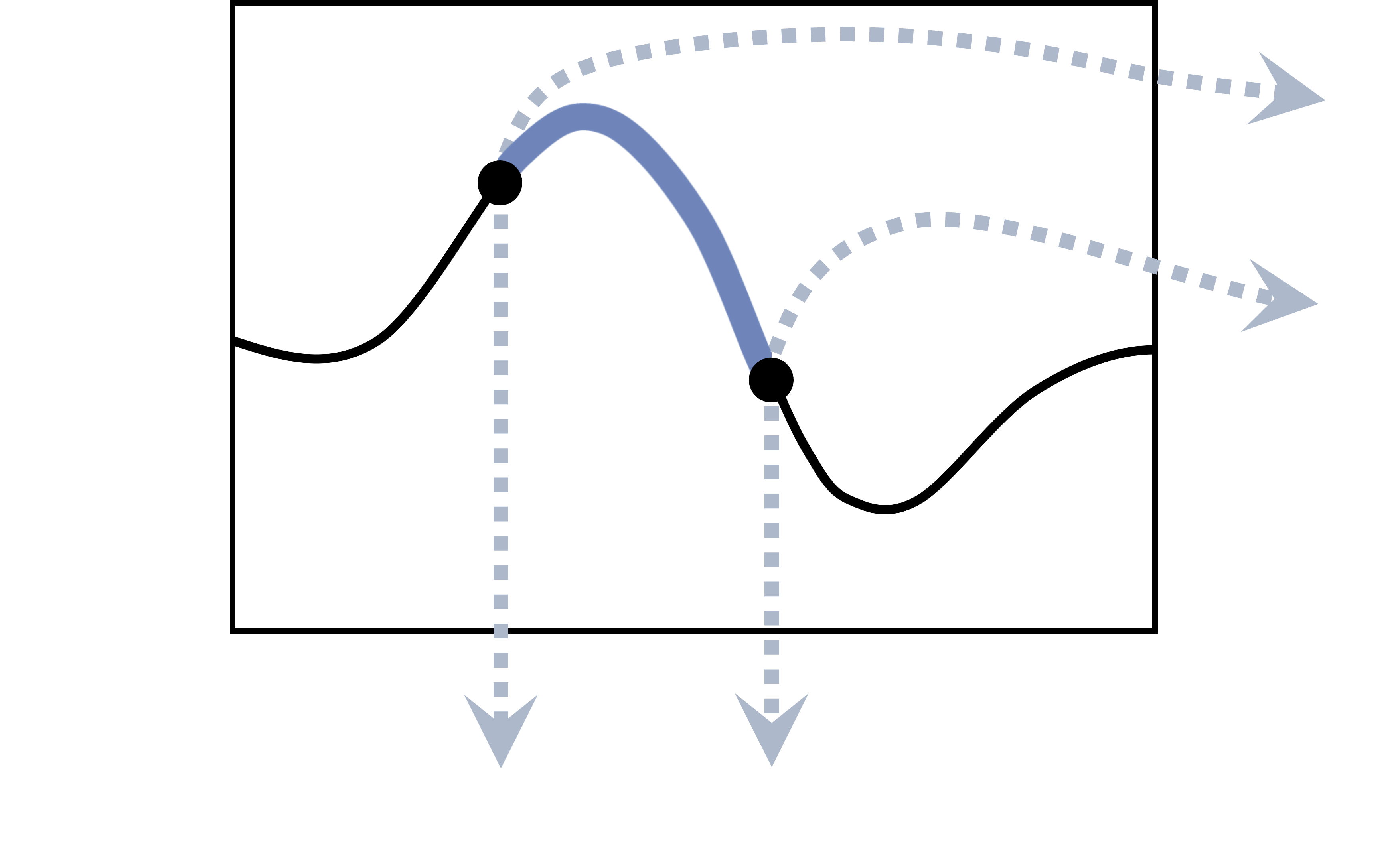}};        

        \matrix (time_pair_v) [above right=-0.52cm and -0.05cm of segment_plot, matrix of nodes, nodes in empty cells, column sep=-\pgflinewidth, row sep=1.5\pgflinewidth, aligned text] {
                    $t_k$\\$t_k'$\\};  
        
        \node (time_pair_h) [below left=-0.07cm and -1.3cm of segment_plot, aligned text] {$t_k, ~~t_k'$};
        \node (pos_pair) [below=0.05cm of time_pair_h, aligned text] {$\vy_{t_k}, \vy_{t_k'}$};

        \node (time_pair_v_arrow) [right=-0.15cm of time_pair_v] {\scalebox{2.5}[1]{\includegraphics[scale=\arrowscale]{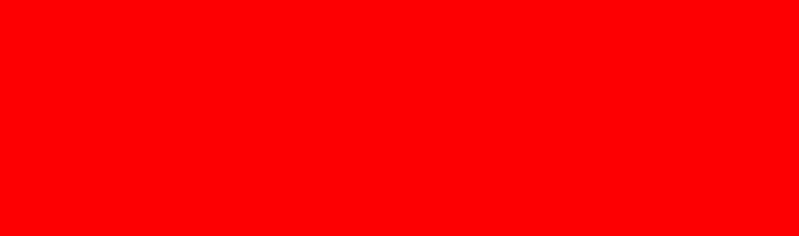}}};

        \node (segment_out_arrow) [anchor=north west, below left=-0.7cm and -3.73cm of prodmp_ll_box] {\includegraphics[scale=\arrowscale]{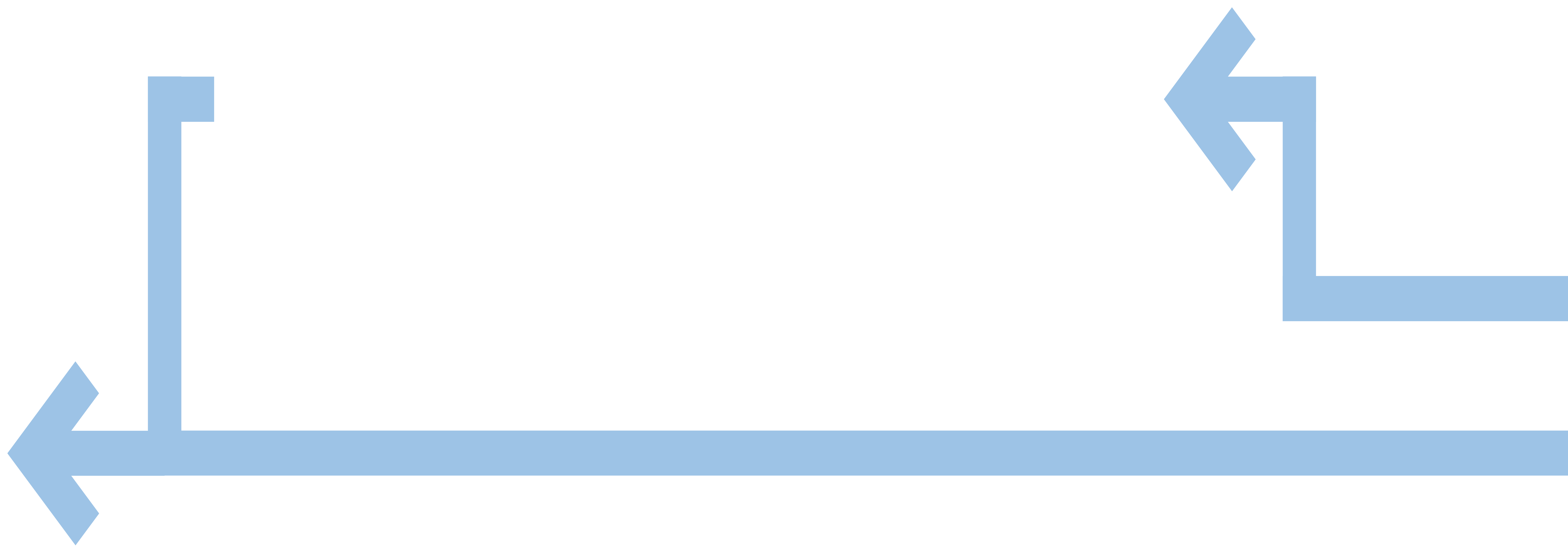}};
        
        \node (update_box)[rectangle, anchor=north east, draw=gray, fill=lightblue, fill opacity=0.1, minimum width=5.5cm, minimum height=4cm, rounded corners=5pt, line width=0.5mm] at (4.65, -1.8) {};
        \node (update_box_txt)[above=-0.35cm of update_box, aligned text] {Policy Update};

        \node (segment_txt) [below left=-0.3cm and 1.1cm of segment_out_arrow, aligned text2] {Likelihood, \eqref{eq:local_likelihood}};
        \node (segment_txt2) [above=0cm of segment_txt, aligned text2] {Compute Segment};
        \node (segment_ll_math) [below=-0.cm of segment_txt, font=\fsize] {$p([\vy_t]_{t=t_k:t_k'})$};

        \node (adv_arrow_in) [above left=-1.3cm and 5.7cm of s_and_r] {\includegraphics[scale=\arrowscale]{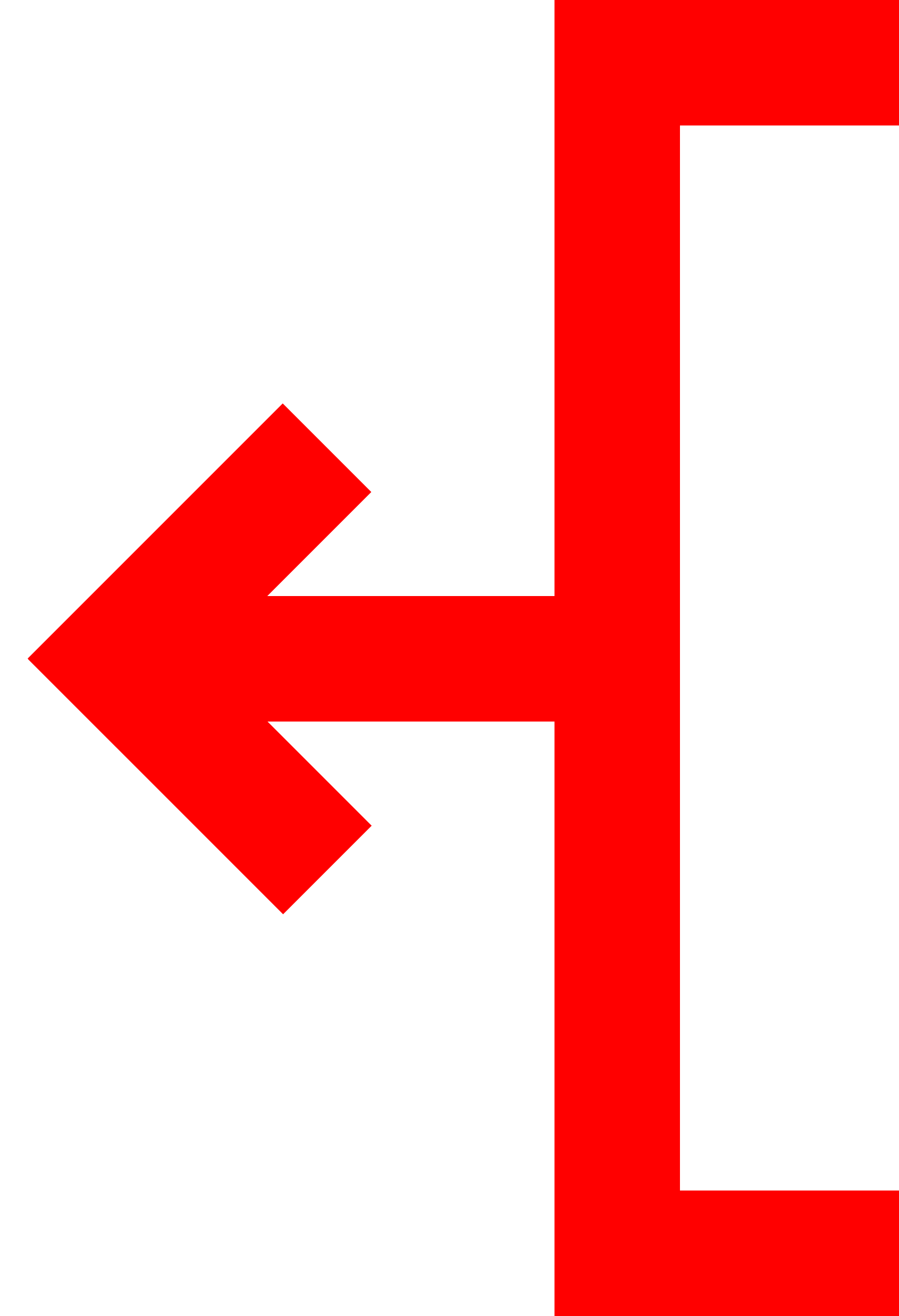}};
        \node (adv_txt)[below=1cm of segment_ll_math, aligned text2] {Advantage, \eqref{eq:adv_to_v}};
        \node (adv_txt2)[above=0.1cm of adv_txt, aligned text2] {Compute Segment};        
        \node (adv_math)[below=-0.cm of adv_txt, font=\fsize] {$A([\vy_t]_{t=t_k:t_k'})$};

        \node (objective_arrow) [anchor=north west, below left=-0.6cm and -0.1cm of segment_ll_math] {\includegraphics[scale=\arrowscale]{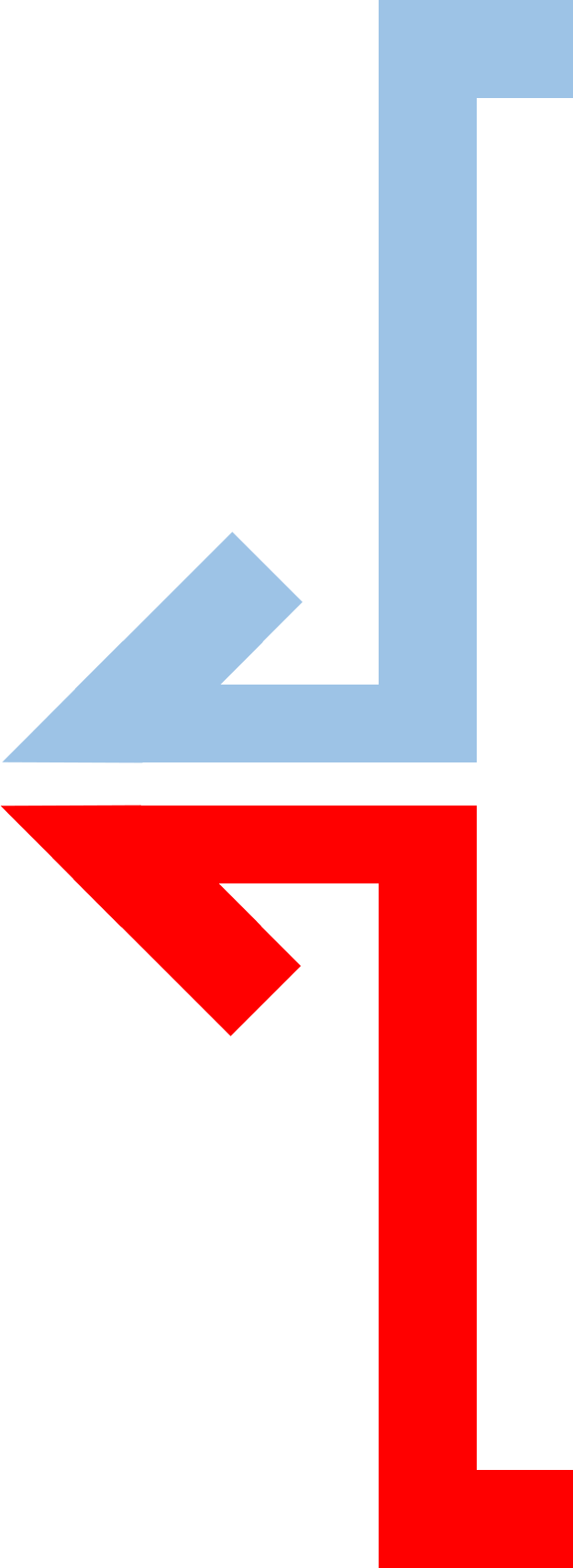}};

        \node (obj_txt) [left=1.9cm of segment_txt2, aligned text2] {Maximize Segment};
        \node (obj_txt2) [below=0.1cm of obj_txt, aligned text2] {Advantage-weighted};
        \node (obj_txt3) [below=0.1cm of obj_txt2, aligned text2] {Likelihood $J$, \eqref{eq:learning_obj}.};
        \node (obj_txt4) [below=0.4cm of obj_txt3, aligned text2] {Policy Gradient Update};
        \node (obj_math) [below=-0.1cm of obj_txt4, font=\fsize]{${\vtheta} \leftarrow {\vtheta}+ \alpha \nabla_{\vtheta}J$,};
        \node (obj_math2) [below=0.3cm of obj_math, aligned text2]{Constraint to Trust Region};
        \node (obj_math3) [below=0.1cm of obj_math2, aligned text2]{of {\footnotesize$\vmu_{\vw}$} and {\footnotesize$\mSigma_{\vw}$}};

    \end{tikzpicture}
    \definecolor{myblue}{RGB}{157,195,230}
    \definecolor{mygreen}{RGB}{169,209,142}
    \definecolor{myred}{RGB}{255,0,0}
    \vspace{-0.3cm}
    \caption{The \tcp framework. The entire learning framework can be divided into three main parts. The first part, shown in {\color{mygreen} \textbf{green}} arrows, involves trajectory sampling, generation, and execution, detailing how the robot is controlled to complete a given task. The second part, indicated in {\color{myblue} \textbf{blue}} arrows, focuses on estimating the likelihood of selecting a particular segment of the sampled trajectory. The third part, marked by {\color{myred} \textbf{red}} arrows, deals with segment evaluation and advantage computation, assessing how much each segment contributes to the successful task completion.}    
    \label{fig:tcp_framework}
\end{figure}

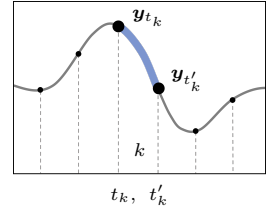
\begin{wrapfigure}{r}{0.25\textwidth}
  \centering
    \newcommand{\fsize}{\footnotesize}
    
    \tikzstyle{aligned text}=[font=\scriptsize\bfseries, text height=1.5ex, text depth=.25ex, anchor=center]
    \tikzstyle{aligned text2}=[font=\scriptsize, text height=1.5ex, text depth=.25ex, anchor=center]
    
    \centering
    \vspace{-0.6cm}
    \begin{tikzpicture}
    
        \node (figure) at (0,0) {\includegraphics[width=0.24\textwidth]{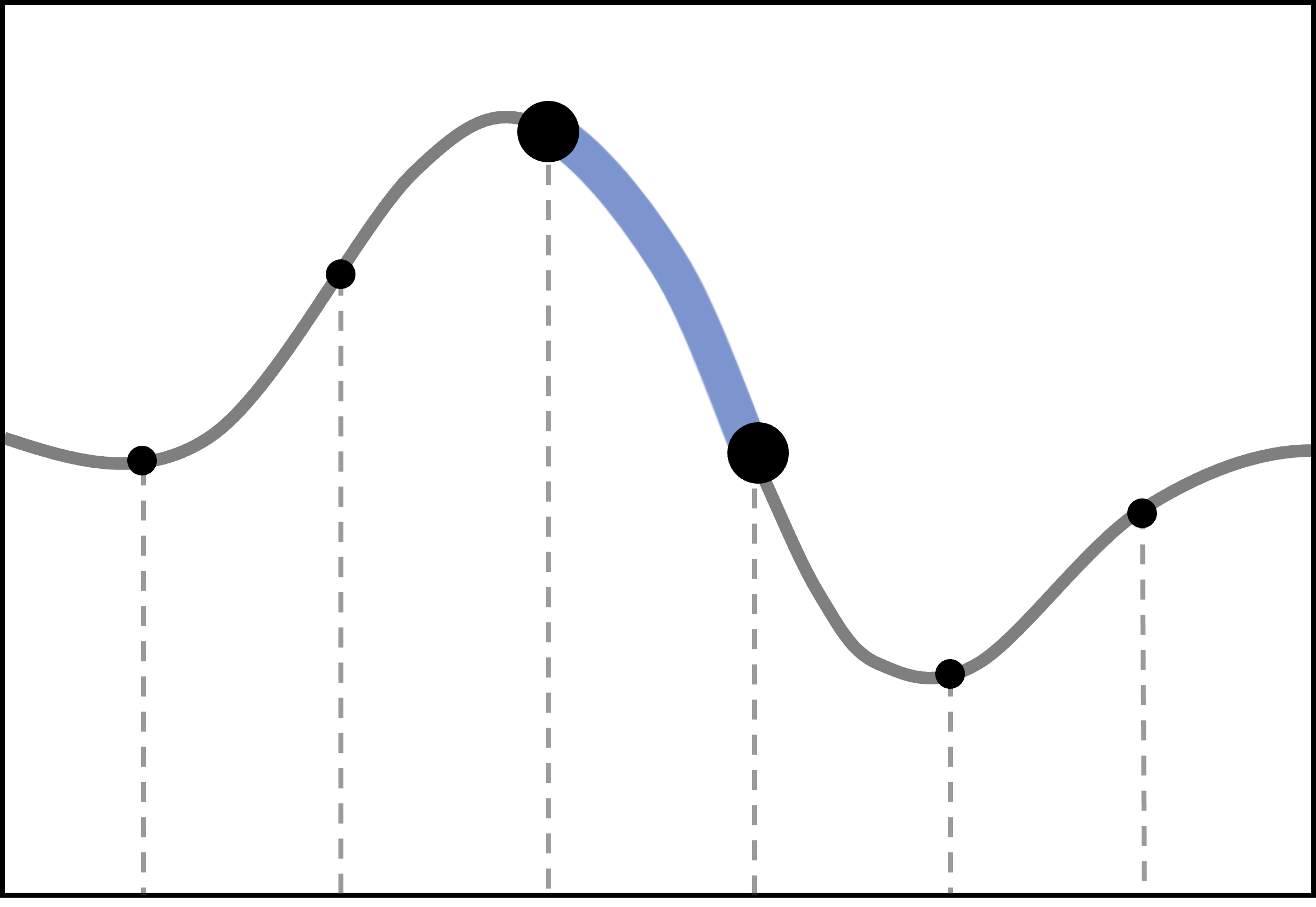}};
        \node (t) [below=0.1cm of figure, aligned text] {$t_k,~~t_k'$};
        \node (k) [above=0.32cm of t, aligned text] {${k}$};

        \node (y_k) [above right=2.15cm and -0.27cm of t, aligned text] {$\vy_{t_k}\quad$};
        \node (y_k') [above right=1.35cm and 0.12cm of t, aligned text] {$\vy_{t_k'}$};
                                    
    \end{tikzpicture}
    \vspace{-0.6cm}
  \caption{Divide a trajectory into $K$ segments}
  \label{fig:traj2seg}
  \vspace{-0.3cm}
\end{wrapfigure}

We introduce an innovative framework of ERL that builds on traditional ERL foundations, aiming to facilitate an efficient policy update mechanism while preserving the intrinsic benefits of ERL. The key innovation lies in redefining the role of trajectories in the policy update process. In contrast to previous methods which consider an entire trajectory as a single data point, our approach breaks down the trajectory into individual segments. Each segment is evaluated and weighted based on its distinct contribution to the task success. This method allows for a more effective use of step-based information in ERL. The comprehensive structure of this framework is depicted in Figure \ref{fig:tcp_framework}.

\textbf{Trajectory Prediction and Generation.}
As highlighted by green arrows in \figref{fig:tcp_framework}, we adopt a structure similar to previous ERL works, such as the one described by \citet{otto2023deep}. 
However, this part distinguish itself by using the most recent \pdmp for trajectory generation and distribution modeling, due to the improved support for trajectory initialization. Additionally, we enhance the previous framework by using a full covariance matrix policy $\pi(\vw|\vs) = \mathcal{N}(\vw| \vmu_{\vw}, \mSigma_{\vw})$ as opposed to a factorized Gaussian policy, to capture a broader range of movement correlations. 

\textbf{Trajectory Likelihood Estimation.}
In \rl, the likelihood of previously sampled actions, along with their associated returns, is often used to adjust the chance of these actions being selected in future policies.
In previous ERL methods, this process typically involves the probability of choosing an entire trajectory.
However, our framework adopts a different strategy, as shown in blue arrows in \figref{fig:tcp_framework}.
Using the techniques in Sections \ref{subsec:background_mp} and \ref{subsec:background_traj_ll}, our approach begins by selecting K paired time steps.
We then transform the parameter likelihood into a trajectory likelihood, which is calculated using these K pairwise likelihoods. 
This approach, depicted in Figure \ref{fig:traj2seg}, effectively divides the whole trajectory into K distinct segments, with each segment defined by a pair of time steps.
In essence, this method allows us to break down the overall trajectory likelihood into individual segment likelihoods, offering a more detailed view of the trajectory's contribution to task success.
\begin{align}    
    &\text{Trajectory to Segments:}\quad\quad p([\vy_t]_{t=0:T}|\vs) \triangleq \{p([\vy_t]_{t=t_k:t_k'}|\vs) \}_{k=1...K}, \label{eq:traj2seg}\\
    &\text{Local Representation:}\quad\quad~~~ p([\vy_t]_{t=t_k:t_k'}|\vs) \triangleq p([\vy_{t_k}, \vy_{t_k'}] | \bm{\mu}_{(t_k, t_k')}(\vs), \bm{\Sigma}_{(t_k, t_k')}(\vs)).
    \label{eq:local_likelihood}
\end{align}

\textbf{Definition of Segment Advantages.}
Similar to standard SRL methods, we leverage the advantage function to evaluate the benefits of executing individual segments within a sampled trajectory. 
When being at state $\vs_{t_k}$ and following the trajectory segment $[\vy_t]_{t=t_k:t_k'}$, the segment-wise advantage function $A(\vs_{t_k},[\vy_t]_{t=t_k:t_k'})$ quantifies the difference between the actual return obtained by executing this sampled trajectory segment and the expected return from a randomly chosen segment, as
\begin{align}        
    \vspace{-0.03cm}
    A(\vs_{t_k}, [\vy_t]_{t=t_k:t_k'}) 
    &= \sum_{t=t_k}^{t=t_k'-1} \gamma^{t-t_k} r_t +  \gamma^{t_k'-t_k}V^{\pi_\textrm{old}}(\vs_{t_k'}) -  V^{\pi_\textrm{old}}(\vs_{t_k}), \label{eq:adv_to_v}
    \vspace{-0.03cm}
\end{align}
where $V^{\pi_\textrm{old}}(\vec s_{t_k})$ denotes the value function of the current policy. In our method, the estimation of $V^{\pi_\textrm{old}}(\vec s_{t_k})$ is consistent with the approaches commonly used in SRL and is independent of the design choice of segment selections. We use NNs to estimate $V^{\pi}(\vs)\approx V^{\pi}_{\phi}(\vs)$ which is fitted on targets of true return or obtained by general advantage estimation (GAE) \citep{schulman2015high}.

\textbf{Learning Objective.}
By replacing the trajectory likelihood and advantage with their segment-based counterparts in the original ERL learning objective as stated in \eqref{eq:traj_rl}, we propose the learning objective of our method as follows
\begin{equation}
    J(\vtheta) = \E_{\pi_{\text{old}}} \left[
    \frac{1}{K} \sum_{k=1}^K %
    \frac{p_{{\pi_{\text{new}}}}([\vy_t]_{t=t_k:t_k'}|\vs) } %
    {p_{{\pi_{\text{old}}}}([\vy_t]_{t=t_k:t_k'}|\vs)} %
    A^{\pi_{\text{old}}}(\vs_{t_k},[\vy_t]_{t=t_k:t_k'}) %
    \right].
    \label{eq:learning_obj}
\end{equation}
Here, $\vs$ denotes the initial state of the episode, used for selecting the parameter vector $\vec w$, and $\vs_{t_k}$ is the state of the system at time $t_k$.
The learning objective takes the \emph{Importance Sampling} to update policies using data from previous policies \citep{schulman2015trust, schulman2017proximal, otto2021differentiable}. 
Our method retains the advantages of exploration in parameter space and generating smooth trajectories.
This enables us to enhance the likelihood of segments with high advantage and reduce the likelihood of less rewarding ones during policy updates.
To ensure a stable update for the full covariance Gaussian policy $\pi_{\vtheta}(\vw|\vs) = \mathcal{N}(\vmu_{\vw}, \mSigma_{\vw})$, we deploy a differentiable Trust Region Projection step \citep{otto2021differentiable} after each policy update iteration as previously discussed in Section \ref{subsec:TRPL}.

\section{Related Works}
\paragraph{Improve Exploration and Smoothness in Step-based RL.}
SRL methods, such as PPO and SAC, interact with the environment by performing a sampled action at each time-step.
This strategy often results in a control signal with high-frequency noise, making it unsuitable for direct use in robotic systems \citep{raffin2022smooth}.
A prevalent solution is to reduce the sampling frequency, a technique commonly known as \emph{frame skip} \citep{braylan2015frame}.
Here, the agent only samples actions every k-th time step and replicates this action for the skipped steps.
Similar approaches decide whether to repeat the last action or to sample a new action in every time step \citep{biedenkapp2021temporl, yu2021taac}.
This concept is also echoed in works such as general State Dependent Exploration (gSDE) \citep{raffin2022smooth, ruckstiess2008state, chiappa2023latent}, where the applied noise is sampled in a state-dependent fashion; leading to smooth changes of the disturbance between consecutive steps.
However, while these methods improved the smoothness in small segments, they struggled to model long-horizon correlations.
Another area of concern is the utilization of white noise during sampling, which fails to consider the temporal correlations between time steps and results in a random walk with suboptimal exploration. 
To mitigate this, previous research, such as \cite{lillicrap2015continuous} and \cite{eberhard2022pink}, have integrated colored noise into the \rl policy, aiming to foster exploration that is correlated across time steps. 
While these methods have shown advantages over white noise approaches, they neither improve the trajectory's smoothness, nor adequately capture the cross-DoF correlations.

\textbf{Episodic RL.}
The early ERL approaches used black-box optimization to evolve parameterized policies, e.g., small NN \citep{whitley1993genetic, igel2003neuroevolution, gomez2008accelerated}. However, these early works were limited to tasks with low-dimensional action space, for instance, the Cart Pole. Although recent works \citep{salimans2017evolution, mania2018simple} have shown that, with more computing, these methods can achieve comparable asymptotic performance with step-based algorithms in some locomotion tasks, none of these methods can deal with tasks with context variations (e.g., changing goals). Another research line in ERL works with more compact policy representation.
\cite{peters2008reinforcement, kober2008policy} first combined ERL with MPs, reducing the dimension of searching space from NN parameter space to MPs parameter space with the extra benefits of smooth trajectories generation. \cite{abdolmaleki2015model} proposed a model-based method to improve the sample efficiency. Notably, although those methods can already handle some complex manipulation tasks such as robot baseball \citep{peters2008reinforcement}, none of them can deal with contextual tasks. To deal with that problem, \citep{abdolmaleki2017contextual} further extends that utilizes linear policy conditioned on the context. Another recent work from this research line \citep{celik2022specializing} proposed using a Mixture of Experts (MoE) to learn versatile skills under the ERL framework.

\textbf{BBRL.} As the first method that utilizes non-linear adaptation to contextual ERL, Deep Black Box Reinforcement Learning (BBRL) \citep{otto2023deep} is the most related work to our method. BBRL applies trust-region-constrained policy optimization to learn a weight adaptation policy for MPs. Despite demonstrating great success in learning tasks with sparse and non-Markovian rewards, it requires significantly more samples to converge compared to SoTA SRL methods. This could be attributed to its black-box nature, where the trajectory from the entire episode is treated as a single data point, and the trajectory return is calculated by summing up all step rewards within the episode.

\section{Experiments}
\label{sec:experiment}
We evaluate the effectiveness of our model through experiments on a variety of simulated robot manipulation tasks. The performance of \tcp is compared against well-known deep RL algorithms as well as methods specifically designed for correlated actions and consistent trajectories. The evaluation is designed to answer the following questions:
\begin{enumerate*}[label=(\alph*)]
\item Can our model effectively train the policy across diverse tasks, incorporating various robot types, control paradigms (task and joint space), and reward configurations?
\item Does the incorporation of movement correlations lead to higher task success rates?
\item Are there limitations or trade-offs when adopting our proposed learning strategy?
\end{enumerate*}

For the comparative evaluation, we select the following methods: PPO, SAC, TRPL, gSDE, PINK \citep{eberhard2022pink} and BBRL. Descriptions, hyper-parameters, and references to the used code bases of these methods can be found in Appendix \ref{appsub:baseline_details}.%
We use step-based methods (PPO, SAC, TRPL, gSDE, and PINK) to predict the lower-level actions for each task. On the other hand, for episodic methods like BBRL and \tcp, we predict position and velocity trajectories and then employ a PD controller to compute the lower-level control commands.
Across all experiments, we measure task success in terms of the number of environment interactions required. Each algorithm is evaluated using 20 distinct random seeds. Results are quantified using the Interquartile Mean (IQM) and are accompanied by a 95\% stratified bootstrap confidence interval \citep{agarwal2021deep}.

\subsection{Large Scale Robot Manipulation Benchmark using Metaworld}
\begin{wrapfigure}{r}{0.65\textwidth}
    \vspace{-0.5cm}
    \hspace*{\fill}%
    \resizebox{0.65\textwidth}{!}{    
        \begin{tikzpicture} 
\def\linewidthtcp{1mm}
\def\linewidthothers{0.5mm}

    \begin{axis}[%
    hide axis,
    xmin=10,
    xmax=50,
    ymin=0,
    ymax=0.4,
    legend style={
        draw=white!15!black,
        legend cell align=left,
        legend columns=-1, 
        legend style={
            draw=none,
            column sep=1ex,
            line width=1pt
        }
    },
    ]
    \addlegendimage{C0, line width=\linewidthtcp}
    \addlegendentry{\acrshort{tcp} (ours)};    
    \addlegendimage{C9, line width=\linewidthothers}
    \addlegendentry{\acrshort{bbrl}};
    \addlegendimage{C2, line width=\linewidthothers}
    \addlegendentry{\acrshort{ppo}};
    \addlegendimage{C3, line width=\linewidthothers}
    \addlegendentry{\acrshort{trpl}};
    \addlegendimage{C4, line width=\linewidthothers}
    \addlegendentry{\acrshort{sac}};    
    \addlegendimage{C5, line width=\linewidthothers}
    \addlegendentry{\acrshort{gsde}};
    \addlegendimage{C6, line width=\linewidthothers}
    \addlegendentry{\acrshort{pink}};

    \end{axis}
\end{tikzpicture}
    }%
    \hspace*{\fill}%
    \newline
    \centering
    \hspace*{\fill}%
    \begin{subfigure}{0.315\textwidth}        
        \resizebox{\textwidth}{!}{\input{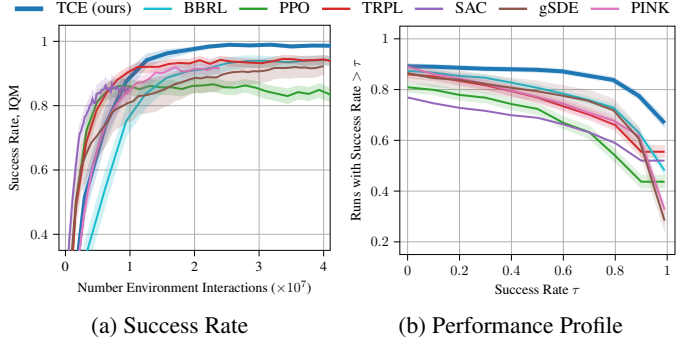}}%
        \caption{Success Rate}
        \label{subfig:meta_sample}
    \end{subfigure}
    \hfill%
    \begin{subfigure}{0.32\textwidth}        
        \resizebox{\textwidth}{!}{        
        \begin{tikzpicture}
\def\tcp_color{C0}
\def\bbrlcolor{C9}
\def\bbrlcov_color{C1}
\def\ppo_color{C2}
\def\trpl_color{C3}
\def\sac_color{C4}
\def\gsde_color{C5}
\def\pink_color{C6}
\def\linewidthtcp{1mm}
\def\linewidthothers{0.5mm}

\begin{axis}[
legend cell align={left},
legend style={
  fill opacity=0.8,
  draw opacity=1,
  text opacity=1,
  at={(0.03,0.03)},
  anchor=south west,
  draw=lightgray204
},
tick align=outside,
tick pos=left,
scaled x ticks=false,
x grid style={darkgray176},
xlabel={Success Rate \(\displaystyle \tau\)},
xmajorgrids,
xmin=-0.03, xmax=1.03,
xtick style={color=black},
y grid style={darkgray176},
ylabel={
Runs with Success Rate \(\displaystyle > \tau\)},
ymajorgrids,
ymin=0.15, ymax=1.05,
ytick style={color=black}
]

\path [draw=\ppo_color, fill=\ppo_color, opacity=0.15]
(axis cs:0,0.827)
--(axis cs:0,0.791)
--(axis cs:0.1,0.78)
--(axis cs:0.2,0.758)
--(axis cs:0.3,0.748)
--(axis cs:0.4,0.721)
--(axis cs:0.5,0.701975)
--(axis cs:0.6,0.645975)
--(axis cs:0.7,0.607)
--(axis cs:0.8,0.513)
--(axis cs:0.9,0.412)
--(axis cs:0.99,0.412)
--(axis cs:0.99,0.462)
--(axis cs:0.99,0.462)
--(axis cs:0.9,0.462)
--(axis cs:0.8,0.563)
--(axis cs:0.7,0.656)
--(axis cs:0.6,0.692)
--(axis cs:0.5,0.746)
--(axis cs:0.4,0.764)
--(axis cs:0.3,0.788)
--(axis cs:0.2,0.799)
--(axis cs:0.1,0.818)
--(axis cs:0,0.827)
--cycle;

\addplot [line width=\linewidthothers, \ppo_color]
table {%
0 0.809
0.1 0.799
0.2 0.779
0.3 0.768
0.4 0.743
0.5 0.724
0.6 0.669
0.7 0.632
0.8 0.539
0.9 0.437
0.99 0.437
};

\path [draw=\trpl_color, fill=\trpl_color, opacity=0.15]
(axis cs:0,0.876)
--(axis cs:0,0.845)
--(axis cs:0.1,0.836)
--(axis cs:0.2,0.816)
--(axis cs:0.3,0.798975)
--(axis cs:0.4,0.775)
--(axis cs:0.5,0.749)
--(axis cs:0.6,0.714975)
--(axis cs:0.7,0.681)
--(axis cs:0.8,0.639)
--(axis cs:0.9,0.533)
--(axis cs:0.99,0.533)
--(axis cs:0.99,0.579)
--(axis cs:0.99,0.579)
--(axis cs:0.9,0.579)
--(axis cs:0.8,0.681)
--(axis cs:0.7,0.721)
--(axis cs:0.6,0.754)
--(axis cs:0.5,0.789)
--(axis cs:0.4,0.814)
--(axis cs:0.3,0.836)
--(axis cs:0.2,0.852)
--(axis cs:0.1,0.869)
--(axis cs:0,0.876)
--cycle;

\addplot [line width=\linewidthothers, \trpl_color]
table {%
0 0.86
0.1 0.852
0.2 0.834
0.3 0.817
0.4 0.794
0.5 0.769
0.6 0.734
0.7 0.701
0.8 0.66
0.9 0.555
0.99 0.555
};

\path [draw=\bbrlcolor, fill=\bbrlcolor, opacity=0.15]
(axis cs:0,0.884)
--(axis cs:0,0.864)
--(axis cs:0.099,0.857)
--(axis cs:0.198,0.843)
--(axis cs:0.297,0.835)
--(axis cs:0.396,0.816)
--(axis cs:0.495,0.795)
--(axis cs:0.594,0.772)
--(axis cs:0.693,0.746)
--(axis cs:0.792,0.714)
--(axis cs:0.891,0.611)
--(axis cs:0.99,0.463)
--(axis cs:0.99,0.498)
--(axis cs:0.99,0.498)
--(axis cs:0.891,0.648)
--(axis cs:0.792,0.744)
--(axis cs:0.693,0.773)
--(axis cs:0.594,0.797)
--(axis cs:0.495,0.82)
--(axis cs:0.396,0.84)
--(axis cs:0.297,0.858)
--(axis cs:0.198,0.865)
--(axis cs:0.099,0.877)
--(axis cs:0,0.884)
--cycle;

\addplot [line width=\linewidthothers, \bbrlcolor]
table {%
0 0.874
0.099 0.867
0.198 0.854
0.297 0.847
0.396 0.828
0.495 0.808
0.594 0.785
0.693 0.76
0.792 0.729
0.891 0.63
0.99 0.481
};

\path [draw=\sac_color, fill=\sac_color, opacity=0.15]
--(axis cs:0,0.751)
--(axis cs:0.1,0.728)
--(axis cs:0.2,0.709)
--(axis cs:0.3,0.696)
--(axis cs:0.4,0.679)
--(axis cs:0.5,0.669)
--(axis cs:0.6,0.643)
--(axis cs:0.7,0.612)
--(axis cs:0.8,0.57)
--(axis cs:0.9,0.501975)
--(axis cs:0.99,0.501975)
--(axis cs:0.99,0.54)
--(axis cs:0.99,0.54)
--(axis cs:0.9,0.54)
--(axis cs:0.8,0.608)
--(axis cs:0.7,0.649)
--(axis cs:0.6,0.682)
--(axis cs:0.5,0.709)
--(axis cs:0.4,0.718)
--(axis cs:0.3,0.735)
--(axis cs:0.2,0.747)
--(axis cs:0.1,0.764)
--(axis cs:0,0.788)
--cycle;

\addplot [line width=\linewidthothers, \sac_color]
table {%
0 0.769
0.1 0.746
0.2 0.728
0.3 0.716
0.4 0.699
0.5 0.689
0.6 0.663
0.7 0.631
0.8 0.589
0.9 0.52
0.99 0.52
};

\path [draw=\tcp_color, fill=\tcp_color, opacity=0.15]
(axis cs:0,0.901)
--(axis cs:0,0.883)
--(axis cs:0.099,0.881)
--(axis cs:0.198,0.877)
--(axis cs:0.297,0.873)
--(axis cs:0.396,0.871)
--(axis cs:0.495,0.868)
--(axis cs:0.594,0.862)
--(axis cs:0.693,0.845)
--(axis cs:0.792,0.827)
--(axis cs:0.891,0.76)
--(axis cs:0.99,0.649)
--(axis cs:0.99,0.688)
--(axis cs:0.99,0.688)
--(axis cs:0.891,0.791)
--(axis cs:0.792,0.849)
--(axis cs:0.693,0.866)
--(axis cs:0.594,0.882)
--(axis cs:0.495,0.887)
--(axis cs:0.396,0.889)
--(axis cs:0.297,0.891)
--(axis cs:0.198,0.895)
--(axis cs:0.099,0.899)
--(axis cs:0,0.901)
--cycle;

\addplot [line width=\linewidthtcp, \tcp_color]
table {%
0 0.892
0.099 0.89
0.198 0.886
0.297 0.882
0.396 0.88
0.495 0.878
0.594 0.872
0.693 0.855
0.792 0.838
0.891 0.775
0.99 0.668
};

\path [draw=\pink_color, fill=\pink_color, opacity=0.15]
(axis cs:0,0.904705882352941)
--(axis cs:0,0.874117647058824)
--(axis cs:0.099,0.843529411764706)
--(axis cs:0.198,0.823529411764706)
--(axis cs:0.297,0.803529411764706)
--(axis cs:0.396,0.775294117647059)
--(axis cs:0.495,0.750588235294118)
--(axis cs:0.594,0.724705882352941)
--(axis cs:0.693,0.688235294117647)
--(axis cs:0.792,0.655294117647059)
--(axis cs:0.891,0.592941176470588)
--(axis cs:0.99,0.304705882352941)
--(axis cs:0.99,0.350588235294118)
--(axis cs:0.99,0.350588235294118)
--(axis cs:0.891,0.641176470588235)
--(axis cs:0.792,0.698823529411765)
--(axis cs:0.693,0.731764705882353)
--(axis cs:0.594,0.767058823529412)
--(axis cs:0.495,0.794117647058823)
--(axis cs:0.396,0.818823529411765)
--(axis cs:0.297,0.843529411764706)
--(axis cs:0.198,0.862352941176471)
--(axis cs:0.099,0.878823529411765)
--(axis cs:0,0.904705882352941)
--cycle;

\addplot [line width=\linewidthothers, \pink_color]
table {%
0 0.889411764705882
0.099 0.861176470588235
0.198 0.843529411764706
0.297 0.823529411764706
0.396 0.796470588235294
0.495 0.772941176470588
0.594 0.745882352941176
0.693 0.710588235294118
0.792 0.677647058823529
0.891 0.617647058823529
0.99 0.327058823529412
};

\path [draw=\gsde_color, fill=\gsde_color, opacity=0.15]
(axis cs:0,0.898230088495575)
--(axis cs:0,0.831858407079646)
--(axis cs:0.099,0.811946902654867)
--(axis cs:0.198,0.803097345132743)
--(axis cs:0.297,0.787610619469027)
--(axis cs:0.396,0.769911504424779)
--(axis cs:0.495,0.756637168141593)
--(axis cs:0.594,0.736725663716814)
--(axis cs:0.693,0.716814159292035)
--(axis cs:0.792,0.676991150442478)
--(axis cs:0.891,0.561946902654867)
--(axis cs:0.99,0.243362831858407)
--(axis cs:0.99,0.325221238938053)
--(axis cs:0.99,0.325221238938053)
--(axis cs:0.891,0.652654867256637)
--(axis cs:0.792,0.763274336283186)
--(axis cs:0.693,0.798672566371681)
--(axis cs:0.594,0.816371681415929)
--(axis cs:0.495,0.834070796460177)
--(axis cs:0.396,0.845132743362832)
--(axis cs:0.297,0.858407079646018)
--(axis cs:0.198,0.873893805309734)
--(axis cs:0.099,0.878373893805309)
--(axis cs:0,0.898230088495575)
--cycle;

\addplot [line width=\linewidthothers, \gsde_color]
table {%
0 0.865044247787611
0.099 0.845132743362832
0.198 0.838495575221239
0.297 0.823008849557522
0.396 0.807522123893805
0.495 0.794247787610619
0.594 0.776548672566372
0.693 0.756637168141593
0.792 0.719026548672566
0.891 0.606194690265487
0.99 0.283185840707965
};

\end{axis}
\end{tikzpicture}}%
        \caption{Performance Profile}
        \label{subfig:meta_perform}
    \end{subfigure}
    \hspace*{\fill}%
    \caption{Metaworld Evaluation. (a) Overall Success Rate across all 50 tasks, reported using Interquartile Mean (IQM) \citep{agarwal2021deep}. (b) Performance profile, illustrating the fraction of runs that exceed the threshold specified on the x-axis.}
    \label{fig:meta}
  \vspace{-0.3cm}
\end{wrapfigure}

We begin our evaluation using the Metaworld benchmark \citep{yu2020meta}, a comprehensive testbed that includes 50 manipulation tasks of varying complexity. Control is executed in a 3-DOF task space along with the finger closeness, and a dense reward signal is employed. In contrast to the original evaluation protocol, we introduce a more stringent framework. Specifically, each episode is initialized with a randomly generated context, rather than a fixed one. Additionally, we tighten the success criteria to only consider a task successfully completed if the objective is maintained until the final time step. This adjustment mitigates scenarios where transient successes are followed by erratic agent behavior.
While we train separate policies for each task, the hyperparameters remain constant across all 50 tasks.
For each method, we report the overall success rate as measured by the IQM across the 50 sub-tasks in \fig\ref{subfig:meta_sample}. The performance profiles are presented in \fig\ref{subfig:meta_perform}.

In both metrics, our method significantly outperforms the baselines in achieving task success. BBRL exhibits the second-best performance in terms of overall consistency across tasks but lags in training speed compared to step-based methods. We attribute this difference to the use of per-step dense rewards, which enables faster policy updates in step-based approaches. \tcp leverages the advantages of both paradigms, surpassing other algorithms after approximately $10^7$ environment interactions.
Notably, the off-policy methods SAC and PINK were trained with fewer samples than used for on-policy methods due to their limitations in parallel environment utilization. 
PINK achieved superior final performance but at the expense of sample efficiency compared to SAC.
In Appendices \ref{appsub:metaperfprof} and \ref{appsub:meta50}, we provide the results for 50 tasks and a performance profile analysis of \tcp.
\subsection{Joint Space Control with Multi Task Objectives}
Next, we investigate the advantages of modeling complete movement correlations and the utility of intermediate feedback for policy optimization. To this end, we enhance the BBRL algorithm by expanding its factorized Gaussian policy to accommodate full covariance (BBRL Cov.), thereby capturing movement correlations. 
Both the original and augmented versions of BBRL are included in the subsequent task evaluations.
We evaluate the methods on a customized Hopper Jump task, sourced from OpenAI Gym \citep{brockman2016openai}. 
This 3-DoF task primarily focuses on maximizing jump height while also accounting for precise landing at a designated location. Control is executed in joint space. We report the max jump height as the main metric of success in \fig\ref{subfig:hopper_height}. 
Our method exhibits quick learning and excels in maximizing jump height. 
Both BBRL versions exhibit similar performance, while BBRL Cov. demonstrates marginal improvements over the original. However, they both fall behind \tcp in training speed, highlighting the efficiency gains we achieve through intermediate state-based policy updates.
Step-based methods like PPO and TRPL tend to converge to sub-optimal policies. The only exception is gSDE. As an augmented step-based method, it enables smoother and more consistent exploration but exhibits significant sensitivity to model initialization (random seeds), evident from the wide confidence intervals.

\subsection{Contact-rich Manipulation with Dense and Sparse Reward Settings}

\begin{figure}[tbh]
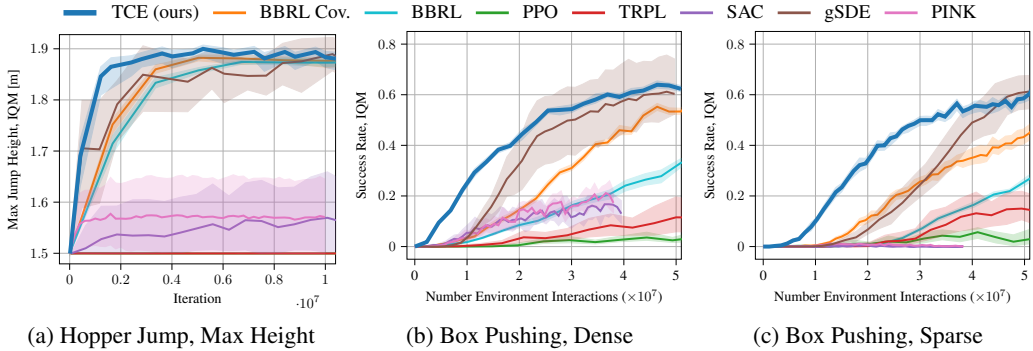

    \centering
    \hspace*{\fill}%
    \resizebox{0.9\textwidth}{!}{
        \begin{tikzpicture} 
\def\linewidthtcp{1mm}
\def\linewidthothers{0.5mm}

    \begin{axis}[%
    hide axis,
    xmin=10,
    xmax=50,
    ymin=0,
    ymax=0.4,
    legend style={
        draw=white!15!black,
        legend cell align=left,
        legend columns=-1, 
        legend style={
            draw=none,
            column sep=1ex,
            line width=1pt
        }
    },
    ]
    \addlegendimage{C0, line width=\linewidthtcp}
    \addlegendentry{\acrshort{tcp} (ours)};    
    \addlegendimage{C1, line width=\linewidthothers}
    \addlegendentry{\acrshort{bbrl} Cov.};
    \addlegendimage{C9, line width=\linewidthothers}
    \addlegendentry{\acrshort{bbrl}};
    \addlegendimage{C2, line width=\linewidthothers}
    \addlegendentry{\acrshort{ppo}};
    \addlegendimage{C3, line width=\linewidthothers}
    \addlegendentry{\acrshort{trpl}};
    \addlegendimage{C4, line width=\linewidthothers}
    \addlegendentry{\acrshort{sac}};    
    \addlegendimage{C5, line width=\linewidthothers}
    \addlegendentry{\acrshort{gsde}};
    \addlegendimage{C6, line width=\linewidthothers}
    \addlegendentry{\acrshort{pink}};

    \end{axis}
\end{tikzpicture}
    }%
    \hspace*{\fill}%
    \newline    
    \hspace*{\fill}%
    \begin{subfigure}{0.32\textwidth}        
        \resizebox{\textwidth}{!}{\input{figure/hopper_jump/plot/height}}%
        \caption{Hopper Jump, Max Height}
        \label{subfig:hopper_height}
    \end{subfigure}
    \hfill%
    \begin{subfigure}{0.32\textwidth}
        \resizebox{\textwidth}{!}{\input{figure/box_pushing/plot/bp_random_dense_iqm}}%
        \caption{Box Pushing, Dense}
        \label{subfig:bp_dense_success}
    \end{subfigure}
    \hfill    
    \begin{subfigure}{0.32\textwidth}
        \resizebox{\textwidth}{!}{\input{figure/box_pushing/plot/bp_random_t_sparse_iqm}}%
        \caption{Box Pushing, Sparse}
        \label{subfig:bp_sparse_success}
    \end{subfigure}
    \hspace*{\fill}%
    \caption{Task Evaluation of (a) Hopper Jump Max Height. (b) Box Pushing success rate in dense reward, and (c) Box Pushing success rate in sparse reward setting.}
    \label{fig:bp}
\end{figure}

We further turn to a 7-\dof robot box-pushing task adapted from \citep{otto2023deep}. The task requires the robot's end-effector, equipped with a rod, to maneuver a box to a specified target position and orientation. 
The difficulty lies in the need for continuous, correlated movements to both position and orient the box accurately. 
To amplify the complexity, the initial pose of the box is randomized. We test two reward settings: dense and sparse. The dense setting offers intermediate rewards inversely proportional to the current distance between the box and its target pose, while the sparse setting only allocates rewards at the episode's end based on the final task state.
Performance metrics for both settings are shown in \fig\ref{subfig:bp_dense_success} and \ref{subfig:bp_sparse_success}. In either case, \tcp and gSDE exhibit superior performance but with \tcp demonstrating greater consistency across different random seeds. The augmented BBRL version outperforms its original counterpart, emphasizing the need for fully correlated movements in tasks that demand consistent object manipulation.
The other step-based methods struggle to learn the task effectively, even when dense rewards are provided. This further highlights the advantages of modeling the movement trajectory as a unified action, as opposed to a step-by-step approach.
\subsection{Hitting Task with High Sparsity Reward Setting}

\begin{wrapfigure}{r}{0.33\textwidth}
    \vspace{-0.5cm}
    \hspace*{\fill}%
    \resizebox{0.33\textwidth}{!}{    
        \begin{tikzpicture} 
\def\linewidthtcp{1mm}
\def\linewidthothers{0.5mm}

    \begin{axis}[%
    hide axis,
    xmin=10,
    xmax=50,
    ymin=0,
    ymax=0.4,
    legend style={
        draw=white!15!black,
        legend cell align=left,
        legend columns=3, 
        legend style={
            draw=none,
            column sep=1ex,
            line width=1pt
        }
    },
    ]
    \addlegendimage{C0, line width=\linewidthtcp}
    \addlegendentry{\acrshort{tcp} (ours)};
    \addlegendimage{C9, line width=\linewidthothers}
    \addlegendentry{\acrshort{bbrl}};
    \addlegendimage{C2, line width=\linewidthothers}
    \addlegendentry{\acrshort{ppo}};
    \addlegendimage{C3, line width=\linewidthothers}
    \addlegendentry{\acrshort{trpl}};
    \addlegendimage{C1, line width=\linewidthothers}
    \addlegendentry{\acrshort{bbrl} Cov.};
    \addlegendimage{C4, line width=\linewidthothers}
    \addlegendentry{\acrshort{sac}};    
    \addlegendimage{C5, line width=\linewidthothers}
    \addlegendentry{\acrshort{gsde}};
    \addlegendimage{C6, line width=\linewidthothers}
    \addlegendentry{\acrshort{pink}};

    \end{axis}
\end{tikzpicture}
    }%
    \newline
    \hspace*{\fill}%
    \begin{subfigure}{0.33\textwidth}
        \resizebox{\textwidth}{!}{\input{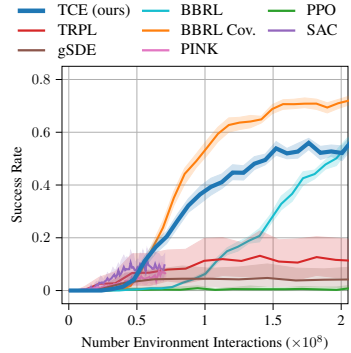}}%
    \end{subfigure}
    \hspace*{\fill}%
    \vspace{-0.5cm}
    \caption{Table Tennis with High reward sparsity.}
    \label{fig:tt_success_alone}
  \vspace{-0.3cm}
\end{wrapfigure}

In our last experiment, we assess the limitations of our method using a 7-\dof robot table tennis task, originally from \citep{celik2022specializing}. The robot aims to return a randomly incoming ball to a desired target on the opponent's court. 
To enhance the task's realism, we randomize the robot's initial state instead of using a fixed starting pose.
This task is distinct due to its one-shot nature: the robot has only one chance to hit the ball and loses control over the ball's trajectory thereafter. The need for diverse hitting strategies like forehand and backhand adds complexity and increases the number of samples required for training.
Performance metrics are presented in \fig\ref{fig:tt_success_alone}. The BBRL Cov. emerges as the leader, achieving a 20\% higher success rate than other methods. It is followed by \tcp and the original BBRL, with \tcp displaying intermediate learning speeds between the two BBRL versions. Step-based methods, led by TRPL at a mere 15\% task success, struggle notably in this setting.
We attribute the underperformance of \tcp and step-based methods to the task's reward sparsity, which complicates the value function learning of SRL and \tcp. Despite these challenges, \tcp maintains its edge over other baselines, further attesting to its robustness, even under stringent conditions.

\section{Conclusion}

We introduced \tcp that synergizes the exploration advantages of ERL with the sample efficiency of SRL. Empirical evaluation showcases that \tcp matches the sample efficiency of SRL and consistently delivers competitive asymptotic performance across various tasks. Furthermore, we demonstrated both the sample efficiency and policy performance of episodic policies can be further improved by incorporating proper correlation modeling.
Despite the promise, several opening questions remain for future work. Firstly, \tcp yields moderate results for tasks characterized by particularly sparse reward settings, as observed in scenarios like table tennis. Secondly, ERL approaches often need a low-level tracking controller, which might not be feasible for certain task types, such as locomotion. Additionally, the current open-loop control setup lacks the adaptability needed for complex control problems in dynamic environments where immediate feedback and swift adaptation are crucial. These issues will be at the forefront of our future work.

\newpage
\section{Acknowledgement}
We thank our colleagues Onur Celik, Maximilian Xiling Li, Vaisakh Kumar Shaj, and Balázs Gyenes at KIT for the valuable discussion, technical support and proofreading.
We thank the anonymous reviewers for their insightful feedback which greatly improved the quality of this paper. 

The research presented in this paper was funded by the Deutsche Forschungsgemeinschaft (DFG, German Research Foundation) – 448648559, and was supported in part by the Helmholtz Association of German Research Centers.
Gerhard Neumann was supported in part by Carl Zeiss Foundation through
the Project JuBot (Jung Bleiben mit Robotern). 
The authors acknowledge support by the state of BadenWürttemberg through bwHPC, as well as the HoreKa supercomputer funded by the Ministry of Science, Research and the Arts Baden-Württemberg and by the Federal Ministry of Education and Research.

\bibliography{iclr2024_conference}
\bibliographystyle{iclr2024_conference}

\newpage
\appendix
{
\Large
\centering
List of content in Appendix
}

\begin{enumerate}
    \renewcommand{\labelenumi}{\Alph{enumi}.}
    \item Algorithm box as a complementary to \figref{fig:tcp_framework}.
    \item Mathematical formulations of MP methods used for trajectory generation.
    \item Experiment settings as a complementary to Section \ref{sec:experiment}.
    \item Additional evaluation and metrics to prove the effectiveness of TCE.
    \item Hyper-parameters selection and sweeping.
    \vspace{1cm}    
\end{enumerate}

\section{Algorithm Box}
\begin{algorithm}
\caption{\TCE}
\begin{algorithmic}[1]
\State Initialize policy parameters $\theta$ and value function parameters $\phi$
\For{iteration = 1, 2, ...}    
    \State Get the initial state $\vs_0$
    \State Predict the mean $\bm{\mu}_{\vw}$, covariance $\bm{\Sigma}_{\vw}$, and sample $\vw*$
    \State Generate the trajectory $\vy*$ using \eqref{eq:promp} and execute it in the environment
    \State Collect step-based information through the execution
    \State Update $\phi$, use true return or GAE style return \citep{schulman2015high}
    \\
    \State Select $K$ time-pairs, e.g. choose every 10 steps along the trajectory
    \State Compute the segment-wise likelihood $\{p_k^{\text{old}}\}_{k=1:K}$ using \eqref{eq:traj2seg} and \ref{eq:local_likelihood} under $\pi^{\text{old}}$
    \For{update epoch = 1, 2, ...}
        \State Make prediction of the mean $\bm{\mu}_{\vw}^{\text{new}}$, covariance $\bm{\Sigma}_{\vw}^{\text{new}}$ under the latest policy $\pi^{\text{new}}$
        \State Enforce Trust Region by projecting $\bm{\mu}_{\vw}^{\text{new}}$ and $\bm{\Sigma}_{\vw}^{\text{new}}$ through TRPL using \eqref{eq:trust_region}
        \State Get projected policy $\tilde{\pi}^{\text{new}}$, represented by $\tilde{\bm{\mu}}_{\vw}^{\text{new}}$ and $\tilde{\bm{\Sigma}}_{\vw}^{\text{new}}$
        \State Compute the segment-wise likelihood $\{p_k^{\text{new}}\}_{k=1:K}$ using \eqref{eq:traj2seg} and \ref{eq:local_likelihood} under $\tilde{\pi}^{\text{new}}$
        \State Update $\theta$ by taking a gradient step w.r.t. $J(\theta)$ in \eqref{eq:learning_obj}
    \EndFor
\EndFor
\end{algorithmic}
\label{algo:tce}
\end{algorithm}

\newpage
\section{Mathematical formulations of \Mp.}
\label{app:pdmp}
In this section, we provide a concise overview of the mathematical formulations of movement primitives utilized in this paper. We begin with the fundamentals of \dmp and \promp, followed by a detailed presentation of \acrshortpl{pdmp}. This includes a focus on trajectory computation and the mapping between parameter distributions and trajectory distributions. For clarity, we begin with a single \acrshort{dof} system and then present the full trajectory distribution using a multi-\acrshort{dof} systems.

\subsection{\DMP} 
\label{subapp:dmp}
\citet{schaal2006dynamic, ijspeert2013dynamical} describe a single movement as a trajectory $[y_t]_{t=0:T}$, which is governed by a second-order linear dynamical system with a non-linear forcing function $f$. The mathematical representation is given by
\begin{equation}
    \tau^2\ddot{y} = \alpha(\beta(g-y)-\tau\dot{y})+ f(x), \quad f(x) = x\frac{\sum\varphi_i(x)w_i}{\sum\varphi_i(x)} = x\bm{\varphi}_x^\intercal\bm{w},
    \label{eq:dmp_original}
\end{equation}
where $y = y(t),~\dot{y}=\mathrm{d}y/\mathrm{d}t,~\ddot{y} =\mathrm{d}^2y/\mathrm{d}t^2$ denote the position, velocity, and acceleration of the system at a specific time $t$, respectively. 
Constants $\alpha$ and $\beta$ are spring-damper parameters, $g$ signifies a goal attractor, and $\tau$ is a time constant that modulates the speed of trajectory execution. To ensure convergence towards the goal, \acrshortpl{dmp} employ a forcing function governed by an exponentially decaying phase variable $x(t)=\exp(-\alpha_x/\tau ; t)$. 
Here, $\varphi_i(x)$ represents the basis functions for the forcing term. The trajectory's shape as it approaches the goal is determined by the weight parameters $w_i \in \bm{w}$, for $i=1,...,N$.
The trajectory $[y_t]_{t=0:T}$ is typically computed by numerically integrating the dynamical system from the start to the end point \citep{ridge2020training, bahl2020neural}. However, this numerical process is computationally intensive, and complicates a directly translation between a parameter distribution $p(w)$ to its corresponding trajectory distribution $p(y)$ \citep{amor2014interaction, meier2016probabilistic}. Previous method, such as GMM/GMR \citep{calinon2012statistical, calinon2016tutorial, yang2018robot} used Gaussian Mixture Models to cover the trajectories' domain. However, this neither captures temporal correlation nor provide a generative model for the trajectories.

\subsection{\PROMP} 
\label{subapp:promp}
\citet{paraschos2013probabilistic} introduced a framework for modeling \mp using trajectory distributions, capturing both temporal and inter-dimensional correlations. Unlike \dmp that use a forcing term, \promp directly model the intended trajectory. The probability of observing a 1-DoF trajectory $[y_t]_{t=0:T}$ given a specific weight vector distribution $p(\bm{w}) \sim \mathcal{N}(\bm{w}|\bm{\mu_w}, \bm{\Sigma_w})$ is represented as a linear basis function model:
\begin{align}
    &\text{Linear basis function:}\quad~~[y_t]_{t=0:T} = \bm{\Phi}_{0:T}^\intercal \bm{w} + \epsilon_{y}, \\
    &\text{Mapping distribution:}\quad~~~ p([y_t]_{t=0:T};~\bm{\mu}_{\vy}, \bm{\Sigma}_{\vy}) = \mathcal{N}( \bm{\Phi}_{0:T}^\intercal\bm{\mu_w}, ~\bm{\Phi}_{0:T}^\intercal \bm{\Sigma_w} \bm{\Phi}_{0:T}~+ \sigma_y^2 \bm{I}).   
\end{align}
Here, $\epsilon_{y}$ is zero-mean white noise with variance $\sigma_y^2$. The matrix $\bm{\Phi}_{0:T}$ houses the basis functions for each time step $t$. Similar to \acrshortpl{dmp}, these basis functions can be defined in terms of a phase variable instead of time.
\acrshortpl{promp} allows for flexible manipulation of \acrshort{mp} trajectories through probabilistic operators applied to $p(\bm{w})$, such as conditioning, combination, and blending \citep{maeda2014learning, gomez2016using, shyam2019improving, rozo2022orientation, zhou2019learning}.
However, \acrshortpl{promp} lack an intrinsic dynamic system, which means they cannot guarantee a smooth transition from the robot's initial state or between different generated trajectories.

\subsection{\PDMP}
\label{subapp:pdmp}
\paragraph{Solving the \ode underlying \dmp}
\citet{li2023prodmp} noted that the governing equation of \acrshortpl{dmp}, as specified in Eq.~(\ref{eq:dmp_original}), admits an analytical solution. This is because it is a second-order linear non-homogeneous \acrshort{ode} with constant coefficients. The original \acrshort{ode} and its homogeneous counterpart can be expressed in standard form as follows:
{
\begin{align}
    \text{Non-homo. ODE:} ~~~\ddot{y} + \frac{\alpha}{\tau}\dot{y} + \frac{\alpha\beta}{\tau^2} y &=\frac{f(x)}{\tau^2}+\frac{\alpha \beta}{\tau^2} g \equiv F(x, g), \label{eq:dmp_non_homo}\\
    \text{Homo. ODE:}~~~\ddot{y} + \frac{\alpha}{\tau}\dot{y} + \frac{\alpha\beta}{\tau^2} y &= 0.
    \label{eq:dmp_homo}
\end{align}
}%
The solution to this \acrshort{ode} is essentially the position trajectory, and its time derivative yields the velocity trajectory. These are formulated as:
\begin{align}
    y &= \begin{bmatrix} y_2\bm{p_2}-y_1\bm{p_1} & y_2q_2 - y_1q_1\end{bmatrix}
         \begin{bmatrix}\bm{w}\\g \end{bmatrix} + c_1y_1 + c_2y_2
         \label{eq:prodmp_pos_detail}\\
    \dot{y} &= \begin{bmatrix}\dot{y}_2\bm{p_2}-\dot{y}_1\bm{p_1} & \dot{y}_2q_2 - \dot{y}_1q_1\end{bmatrix}
               \begin{bmatrix}\vspace{-0.0cm}\bm{w}\\g \end{bmatrix} + c_1\dot{y}_1 + c_2\dot{y}_2
        \label{eq:prodmp_vel_detail}.
\end{align}
Here, the learnable parameters $\bm{w}$ and $g$ which control the shape of the trajectory, are separable from the remaining terms. Time-dependent functions $y_1, y_2, \bm{p}_1$, $\bm{p}_2$, $q_1$, $q_2$ in the remaining terms offer the basic support to generate the trajectory. The functions $y_1, y_2$ are the complementary solutions to the homogeneous \ode presented in \eqref{eq:dmp_homo}, and $\dot{y}_1, \dot{y}_2$ their time derivatives respectively. These time-dependent functions take the form as:
{
\small
\begin{align}
    y_1(t) &= \exp\left(-\frac{\alpha}{2\tau}t\right), \quad\quad\quad\quad\quad\quad\quad\quad\quad~~
    y_2(t) = t\exp\left(-\frac{\alpha}{2\tau}t\right),
    \label{eq:dmp_y1_y2} \\
    \bm{p}_1(t) &= \frac{1}{\tau^2}\int_0^t t'\exp\Big(\frac{\alpha}{2\tau}t'\Big)x(t')\bm{\varphi}_x^\intercal \mathrm{d}t', \quad~~~ 
    \bm{p}_2(t) = \frac{1}{\tau^2}\int_0^t \exp\Big(\frac{\alpha}{2\tau}t'\Big)x(t')\bm{\varphi}_x^\intercal \mathrm{d}t',\label{eq:dmp_p1_p2}\\
    q_1(t) &= \Big(\frac{\alpha}{2\tau}t  - 1\Big)\exp\Big(\frac{\alpha}{2\tau}t\Big) +1,
    \quad\quad\quad\quad  q_2(t) = \frac{\alpha}{2\tau} \bigg[\exp\Big(\frac{\alpha}{2\tau}t\Big)-1\bigg]. \label{eq:dmp_q1_q2}   
\end{align}}
It's worth noting that the $\bm{p}_1$ and $\bm{p}_2$ cannot be analytically derived due to the complex nature of the forcing basis terms $\bm{\varphi}_x$. As a result, they need to be computed numerically. Despite this, isolating the learnable parameters, namely $\bm{w}$ and $g$, allows for the reuse of the remaining terms across all generated trajectories.
These residual terms can be more specifically identified as the position and velocity basis functions, denoted as $\ipb(t)$ and $\ivb(t)$, respectively. When both $\bm{w}$ and $g$ are included in a concatenated vector, represented as $\bm{w}_g$, the expressions for position and velocity trajectories can be formulated in a manner akin to that employed by \acrshortpl{promp}:
\begin{align}
    \textbf{Position:}\quad y(t) &= \ipb(t)^\intercal \vw_g + c_1y_1(t) + c_2y_2(t), \label{eq:prodmp_pos}\\ 
    \textbf{Velocity:}\quad \dot{y}(t)&= \ivb(t)^\intercal\vw_g + c_1\dot{y}_1(t) + c_2\dot{y}_2(t). \label{eq:prodmp_vel}
\end{align}
\textbf{In the main paper, for simplicity and notation convenience, we use $\vw$ instead of $\vw_g$ to describe the parameters and goal of \pdmp.}

\paragraph{Smooth trajectory transition}
The coefficients $c_1$ and $c_2$ serve as solutions to the initial value problem delineated by the Eq.(\ref{eq:prodmp_pos})(\ref{eq:prodmp_vel}). 
\citeauthor{li2023prodmp} propose utilizing the robot's initial state or the replanning state, characterized by the robot's position and velocity ($y_b, \dot{y}_b$) to ensure a smooth commencement or transition from a previously generated trajectory.
Denote the values of the complementary functions and their derivatives at the condition time $t_b$ as $y_{1_b}, y_{2_b}, \dot{y}_{1_b}$ and $\dot{y}_{2_b}$. Similarly, denote the values of the position and velocity basis functions at this time as $\ipbb$ and $\ivbb$ respectively. Using these notations, $c_1$ and $c_2$ can be calculated as follows:
\begin{equation}
    \begin{bmatrix}c_1\\c_2\end{bmatrix}
    = \begin{bmatrix}\frac{\dot{y}_{2_b}y_b-y_{2_b}\dot{y}_b}{y_{1_b}\dot{y}_{2_b}-y_{2_b}\dot{y}_{1_b}} +\frac{y_{2_b}\ivbbt-\dot{y}_{2_b}\ipbbt}{y_{1_b}\dot{y}_{2_b}-y_{2_b}\dot{y}_{1_b}}\bm{w}_g\\[0.7em]
    \frac{y_{1_b}\dot{y}_b-\dot{y}_{1_b}y_b}{y_{1_b}\dot{y}_{2_b}-y_{2_b}\dot{y}_{1_b}}+\frac{\dot{y}_{1_b}\ipbbt- y_{1_b}\ivbbt}{y_{1_b}\dot{y}_{2_b}-y_{2_b}\dot{y}_{1_b}}\bm{w}_g\end{bmatrix}.
    \label{eq:dmp_c}
\end{equation}
Substituting Eq.~(\ref{eq:dmp_c}) into Eq.~(\ref{eq:prodmp_pos}) and Eq.~(\ref{eq:prodmp_vel}), the position and velocity trajectories take the form as
\begin{align}
    y &= \xi_1 y_b  + \xi_2 \dot{y}_b + [\xi_3 \ipbb + \xi_4 \ivbb+ \ipb]^\intercal\bm{w}_g, \label{eq:prodmp_pos_full}\\
    \dot{y}&= \dot{\xi}_1 y_b +\dot{\xi}_2 \dot{y}_b + [\dot{\xi}_3\ipbb + \dot{\xi}_4 \ivbb + \ivb]^\intercal\bm{w}_g \label{eq:prodmp_vel_full}
\end{align}
Here, $\xi_k$ for $k\in\{1,2,3,4\}$ serve as intermediate terms that are derived from the complementary functions and the initial conditions. The formations of these terms are elaborated below. To find their derivatives {\small$\dot{\xi}_k$}, one can simply replace $y_1, y_2$ with their time derivatives $\dot{y}_1, \dot{y}_2$ in the equations.
{
\footnotesize
\begin{align*}
    \xi_1(t)&=\frac{\dot{y}_{2_b}y_1-\dot{y}_{1_b}y_2}{y_{1_b}\dot{y}_{2_b}-y_{2_b}\dot{y}_{1_b}},\quad\quad\xi_2(t)=\frac{y_{1_b}y_2-y_{2_b}y_1}{y_{1_b}\dot{y}_{2_b}-y_{2_b}\dot{y}_{1_b}},\\
    \xi_3(t)&=\frac{\dot{y}_{1_b}y_2-\dot{y}_{2_b}y_1}{y_{1_b}\dot{y}_{2_b}-y_{2_b}\dot{y}_{1_b}},\quad\quad\xi_4(t)=\frac{y_{2_b}y_1- y_{1_b}y_2}{y_{1_b}\dot{y}_{2_b}-y_{2_b}\dot{y}_{1_b}}.
\end{align*}
}
\paragraph{Extend to a High \dof system}
Both \promp and \pdmp can be generalized to accommodate high-\dof systems. This allows for the capture of both temporal correlations and interactions among various \dof. Such generalization is implemented through modifications to matrix structures and vector concatenations, as illustrated in \citet{paraschos2013probabilistic,li2023prodmp}.
To be specific, the basis functions $\ipb, \ivb$, along with their values at the condition time $\ipbb, \ivbb$, are extended to block-diagonal matrices $\bipb, \bivb$, $\bipbb$ and $\bivbb$ respectively. This extension is executed by tiling the existing basis function matrices $D$ times along their diagonal, where $D$ is the number of \dof. Additionally, the robot initial conditions for each \dof are concatenated into one  vectors. For instance, the initial positions $y^1_b,~...,~y^D_b$ are unified into a single vector $\bm{y}_b = [y^1_b,~...,~y^D_b]^\intercal$. In this way, the position and velocity trajectories are extended as 
{
\begin{align}
    \bm{y} &= \xi_1  \bm{y}_b + \xi_2  \dot{\bm{y}}_b +[\xi_3  \bipbb + \xi_4  \bivbb+ \bipb]^\intercal\bm{w}_g,\label{eq:idmp_pos_multi_full}\\
    \dot{\bm{y}} &= \dot{\xi}_1  \bm{y}_b +\dot{\xi}_2  \dot{\bm{y}}_b +[\dot{\xi}_3  \bipbb + \dot{\xi}_4  \bivbb + \bivb]^\intercal\bm{w}_g.
    \label{eq:idmp_vel_multi_full}
\end{align}
}

\paragraph{Parameter distribution to trajectory distribution}
In a manner analogous to the description provided for \promp from Equation \eqref{eq:promp} to Equation \eqref{eq:promp_p_of_y}, \pdmp also exhibits a comparable architecture framework. This similarity is particularly evident in the structure of the learnable parameters, denoted as $\vw_g$, which follow a linear basis function format. Consequently, it is reasonable to delineate the trajectory distribution for \pdmp in fashion akin to that of \promp. 
Given that the parameter distribution $\bm{w}_g$ follows a Gaussian distribution $\bm{w}_g\sim\mathcal{N}(\bm{w}_g|\bm{\mu_{w_g}},\bm{\Sigma_{w_g}})$ and adhering to the probabilistic formulation analogous to \promp as indicated in \eqref{eq:promp_p_of_y}, the trajectory distribution for \pdmp can be expressed as:
{
\begin{equation}
    p([y_t]_{t=0:T};~\bm{\mu}_{\vy}, \bm{\Sigma}_{\vy}) = ~\mathcal{N}([\vy_t]_{t=0:T}|~\bm{\mu}_{\vy}, \bm{\Sigma}_{\vy}),\label{eq:idmp_mean_cov}
\end{equation}
where
\begin{align*}    
    &\bm{\mu}_{\vy} = \bm{\xi}_1 \bm{y}_b + \bm{\xi}_2 \dot{\bm{y}}_b + \bm{H}^\intercal_{0:T} \bm{\mu}_{w_g}, \quad\quad
    \bm{\Sigma}_{\vy} =\bm{H}^\intercal_{0:T}\bm{\Sigma}_{w_g} \bm{H}_{0:T}+ \sigma_n^2 \mathbf{I},\\
    &\bm{H}_{0:T} = \bm{\xi}_3 \bipbb + \bm{\xi}_4 \bivbb+ \bipb_{0:T}, \quad\quad~~ \bm{\xi}_k = [\xi_k(t)]_{t=0:T}.
\end{align*}
}%
In this context, the trajectory mean, denoted as $\bm{\mu}_{\vy}$ constitutes a vector of dimension $DT$, whereas the trajectory covariance, represented by $\bm{\Sigma}_{\vy}$ is a $DT \times DT$ dimensional matrix. 
These quantities serve to integrate the trajectory values across all degrees of freedom (\dof) and temporal steps, encapsulating them within a single distribution.
This multi-\dof \pdmp representation can be seen as an enhancement of the \promp framework, augmented by the inclusion of initial condition terms. This ensures that the trajectories sampled under this distribution start from the specified initial state.
\textbf{Additionally, the time range $t=0:T$ is flexible and can be replaced by any set of discrete time points. For instance, in the \tcp method, a pair of time points $t_k$ and $t_k'$ can define a trajectory segment, allowing for down-sampling of the trajectory distribution to specific trajectroy segment.}

\newpage
\section{Experiment Details}
\label{app:exp}
\subsection{Details of Methods Implementation}
\label{appsub:baseline_details}

\paragraph{PPO} Proximal Policy Optimization (PPO) \citep{schulman2017proximal} is a prominent on-policy step-based RL algorithm that refines the policy gradient objective, ensuring policy updates remain close to the behavior policy. PPO branches into two main variants: PPO-Penalty, which incorporates a KL-divergence term into the objective for regularization, and PPO-Clip, which employs a clipped surrogate objective. In this study, we focus our comparisons on PPO-Clip due to its prevalent use in the field. Our implementation of PPO is based on the implementation of \cite{stable-baselines3}.

\paragraph{SAC} Soft Actor-Critic (SAC) \citep{haarnoja2018soft, haarnoja2018sacaa} employs a stochastic step-based policy in an off-policy setting and utilizes double Q-networks to mitigate the overestimation of Q-values for stable updates. By integrating entropy regularization into the learning objective, SAC balances between expected returns and policy entropy, preventing the policy from premature convergence. Our implementation of SAC is based on the implementation of \cite{stable-baselines3}. 

\paragraph{TRPL} Trust Region Projection Layers (TRPL) \citep{otto2021differentiable}, akin to PPO, addresses the problem of stabilizing the on-policy policy gradient by constraining the learning policy staying close to the behavior policy. TRPL formulates the constrained optimization problem as a projection problem, providing a mathematically rigorous and scalable technique that precisely enforces trust regions on each state, leading to stable and efficient on-policy updates. We evaluated its performance based on the implementation of the original work.

\paragraph{gSDE}
Generalized State Dependent Exploration (gSDE) \citep{raffin2022smooth, ruckstiess2008state, ruckstiess2010exploring} is an exploration method designed to address issues with traditional step-based exploration techniques and aims to provide smoother and more efficient exploration in the context of robotic reinforcement learning, reducing jerky motion patterns and potential damage to robot motors while maintaining competitive performance in learning tasks.

To achieve this, gSDE replaces the traditional approach of independently sampling from a Gaussian noise at each time step with a more structured exploration strategy, that samples in a state-dependent manner. The generated samples not only depend on parameter of the Gaussian distribution $\mu$ \& $\Sigma$, but also on the activations of the policy network's last hidden layer ($s$). We generate disturbances $\epsilon_t$ using the equation
\begin{equation*}
    \epsilon_t = \theta_\epsilon s \text{, where } \theta_\epsilon \thicksim \mathcal{N}^d\left(0, \Sigma\right).
\end{equation*}
The exploration matrix $\theta_\epsilon$ is composed of vectors of length $\text{Dim}(a)$ that were drawn from the Gaussian distribution we want gSDE to follow. The vector $s$ describes how this set of pre-computed exploration vectors are mixed. 
The exploration matrix is resampled at regular intervals, as guided by the 'sde sampling frequency' (ssf), occurring every n-th step if n is our ssf.

gSDE is versatile, applicable as a substitute for the Gaussian Noise source in numerous on- and off-policy algorithms. We evaluated its performance in an on-policy setting using PPO by utilizing the reference implementation for gSDE from \citet{raffin2022smooth}.
In order for training with gSDE to remain stable and reach high performance the usage of a linear schedule over the clip range had to be used for some environments.

\paragraph{PINK}

We utilize SAC to evaluate the effectiveness of pink noise for efficient exploration. 
\citet{eberhard2022pink} propose to replace the independent action noise \(\eps_t\) of 
\begin{equation*}
    a_t = \mu_t + \sigma_t \cdot \eps_t  
\end{equation*}
with correlated noise from particular random processes, whose power spectral density follow a power law.
In particular, the use of pink noise, with the exponent \(\beta = 1\) in \({S(f) = \left|\mathcal{F}[\eps](f)\right|^2 \propto f^{-\beta}}\), should be considered \citep{eberhard2022pink}.

We follow the reference implementation and sample chunks of Gaussian pink noise using the inverse Fast Fourier Transform method proposed by \citet{Timmer1995OnGP}.
These noise variables are used for SAC's exploration but the the actor and critic updates sample the independent action distribution without pink noise.
Each action dimension uses an independent noise process which causes temporal correlation within each dimension but not across dimensions.
Furthermore, we fix the chunk size and maximum period to $10000$ which avoids frequent jumps of chunk borders and increases relative power of low frequencies.

\paragraph{BBRL-Cov/Std} Black-Box Reinforcement Learning (BBRL) \citep{otto2023deep, otto2023mp3} is a recent developed episodic reinforcement learning method. By utilizing ProMPs \citep{paraschos2013probabilistic} as the trajectory generator, BBRL learns a policy that explores at the trajectory level. The method can effectively handle sparse and non-Markovian rewards by perceiving an entire trajectory as a unified data point, neglecting the temporal structure within sampled trajectories. However, on the other hand, BBRL suffers from relatively low sample efficiency due to its black-box nature. Moreover, the original BBRL employs a degenerate Gaussian policy with diagonal covariance. In this study, we extend BBRL to learn Gaussian policy with full covariance to build a more competitive baseline. For clarity, we refer to the original method as BBRL-Std and the full covariance version as BBRL-Cov. 
We integrate BBRL with \pdmp\citep{li2023prodmp}, aiming to isolate the effects attributable to different MP approaches. 

\subsection{Metaworld Performance Profile Analysis}
\label{appsub:metaperfprof}
The distribution of success rates, reported in the performance profile in \fig\ref{subfig:meta_perform}, may seem to contradict the nearly perfect IQM of \tcp but in reality provide insight into the consistency of \tcp.
Nearly two thirds of runs exceed 99\% success rate and are therefore able to perfectly solve the task with this seed.
The individual performances reported in Appendix \ref{appsub:meta50} show that only very few tasks, e.g., Disassemble and Hammer, have a significant fraction of unsuccessful seeds.
This consistency per task is also visible in the profile, as only two percent of runs fall between zero and sixty percent success rate which is visible by the near zero slope in this range.
All methods are entirely unable to solve a small set of tasks and therefore show a gap in the profile.
This does not contradict the very high IQM success rate of \tcp, as the IQM trims the upper and lower 25\% of results.
The commonly reported median effectively trims 50\% and would result in even higher values.
Due to the smaller and later decline in the profile compared to the other methods, the intersection between 75\% of runs and the profile is located at a success rate of 90\%.
Therefore, only a small fraction of runs, roughly ten percent, fall within the 25\% trim but only slightly decrease the value of the IQM due their high success rate.
Other methods have a larger fraction of imperfect runs with lower success rate within the quartiles.

\newpage
\subsection{Performance on Individual Metaworld Tasks}
\label{appsub:meta50}
We report the Interquartile Mean (IQM) of success rates for each Metaworld task. The plots clearly illustrate the varying levels of difficulty across different tasks.
\begin{figure}[b!]
    \hspace*{\fill}%
    \resizebox{0.8\textwidth}{!}{    
        \begin{tikzpicture} 
\def\linewidthtcp{1mm}
\def\linewidthothers{0.5mm}

    \begin{axis}[%
    hide axis,
    xmin=10,
    xmax=50,
    ymin=0,
    ymax=0.4,
    legend style={
        draw=white!15!black,
        legend cell align=left,
        legend columns=-1, 
        legend style={
            draw=none,
            column sep=1ex,
            line width=1pt
        }
    },
    ]
    \addlegendimage{C0, line width=\linewidthtcp}
    \addlegendentry{\acrshort{tcp} (ours)};    
    \addlegendimage{C9, line width=\linewidthothers}
    \addlegendentry{\acrshort{bbrl}};
    \addlegendimage{C2, line width=\linewidthothers}
    \addlegendentry{\acrshort{ppo}};
    \addlegendimage{C3, line width=\linewidthothers}
    \addlegendentry{\acrshort{trpl}};
    \addlegendimage{C4, line width=\linewidthothers}
    \addlegendentry{\acrshort{sac}};    
    \addlegendimage{C5, line width=\linewidthothers}
    \addlegendentry{\acrshort{gsde}};
    \addlegendimage{C6, line width=\linewidthothers}
    \addlegendentry{\acrshort{pink}};

    \end{axis}
\end{tikzpicture}
    }%
    \hspace*{\fill}%
    \newline
    
  \centering
  \begin{subfigure}{0.32\textwidth}
    \includegraphics[width=\linewidth]{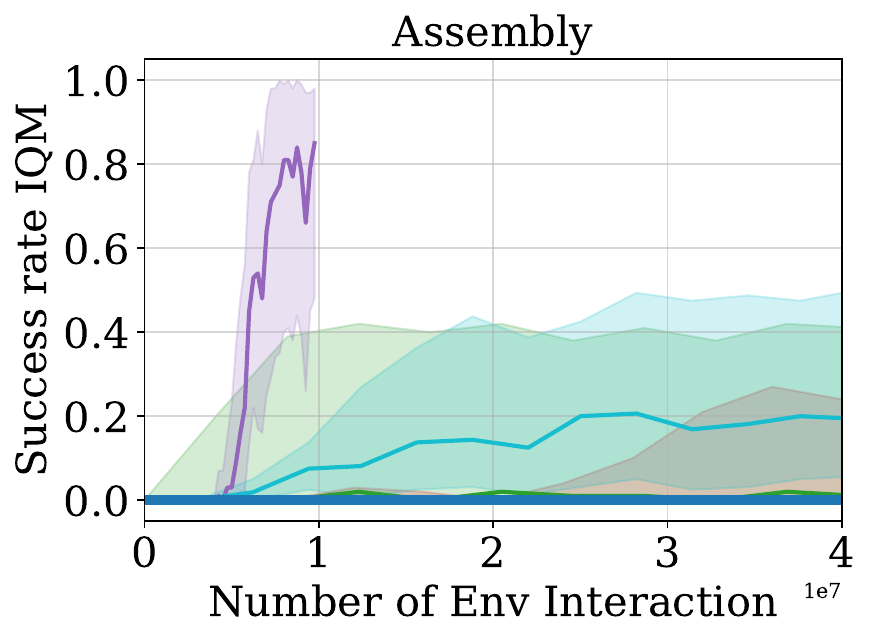}
  \end{subfigure}\hfill
  \begin{subfigure}{0.32\textwidth}
    \includegraphics[width=\linewidth]{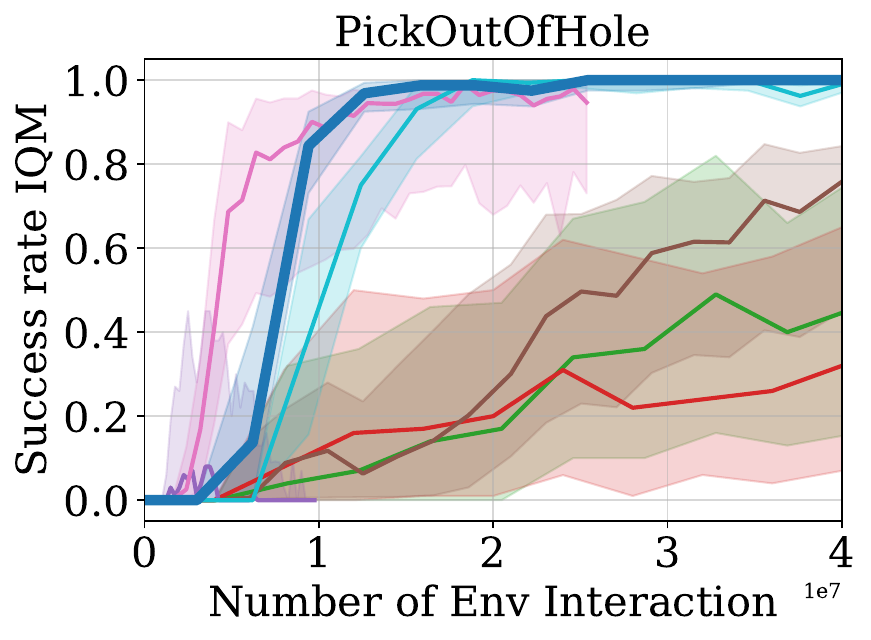}
  \end{subfigure}\hfill
  \begin{subfigure}{0.32\textwidth}
    \includegraphics[width=\linewidth]{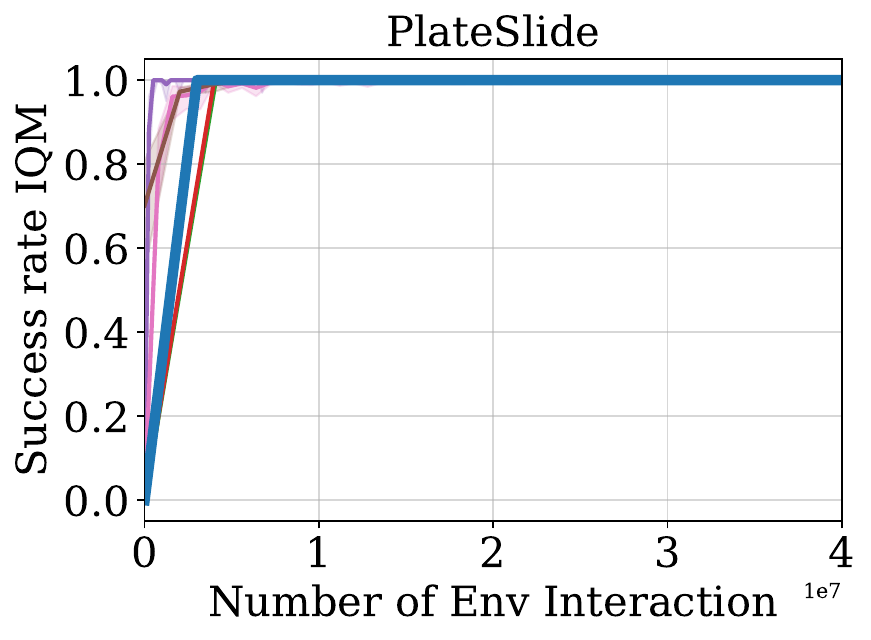}
  \end{subfigure}
  \vspace{0.1cm} 
  \begin{subfigure}{0.32\textwidth}
    \includegraphics[width=\linewidth]{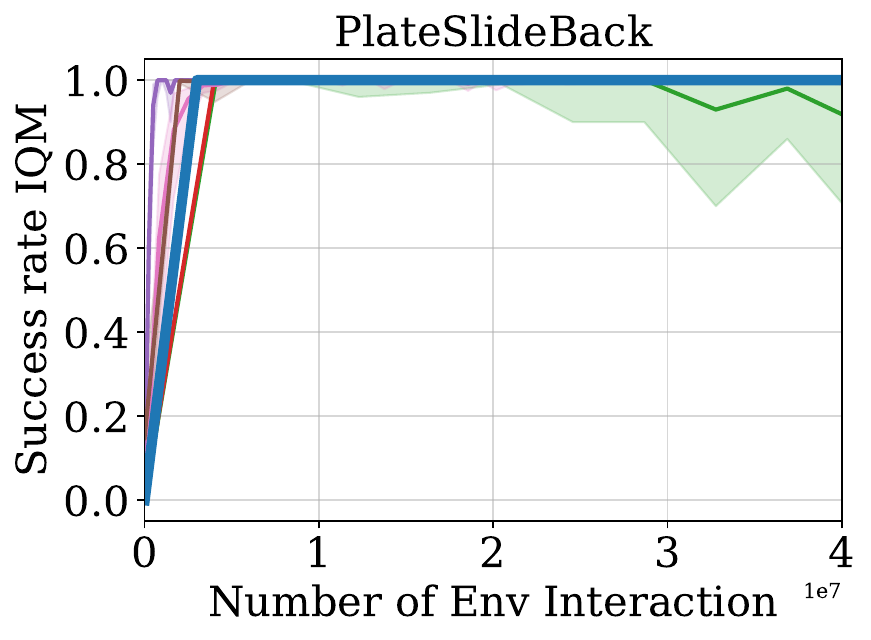}
  \end{subfigure}\hfill
  \begin{subfigure}{0.32\textwidth}
    \includegraphics[width=\linewidth]{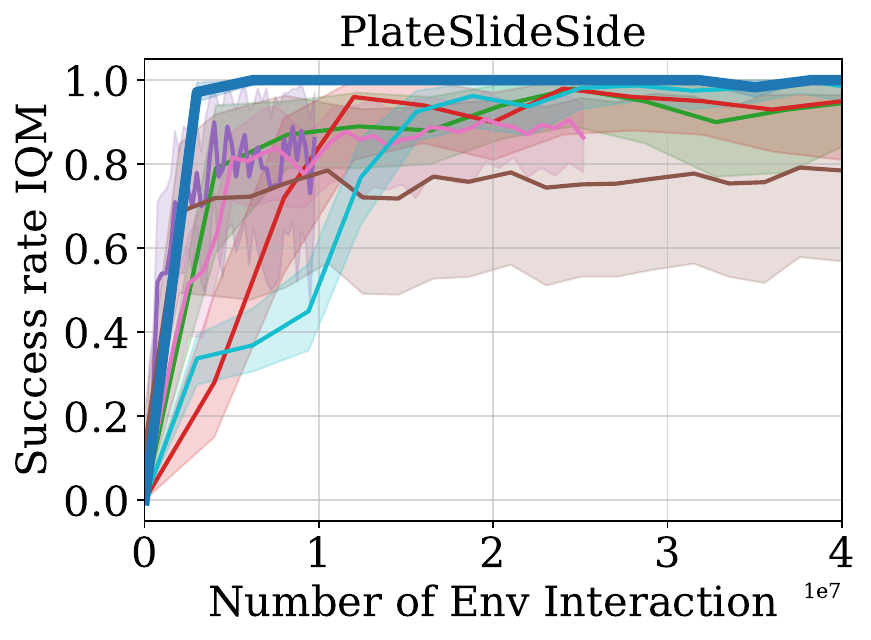}
  \end{subfigure}\hfill
  \begin{subfigure}{0.32\textwidth}
    \includegraphics[width=\linewidth]{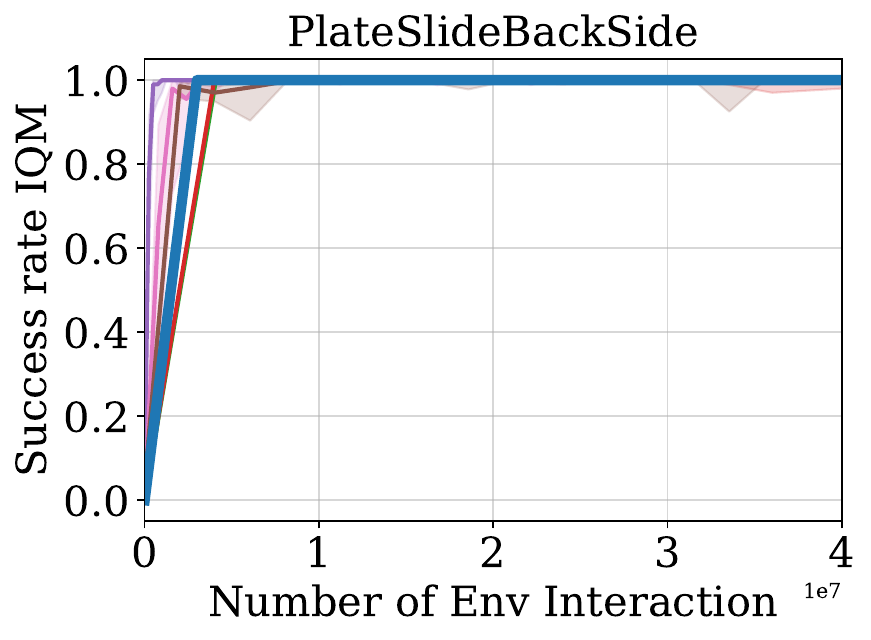}
  \end{subfigure}
  \vspace{0.1cm} 
  \begin{subfigure}{0.32\textwidth}
    \includegraphics[width=\linewidth]{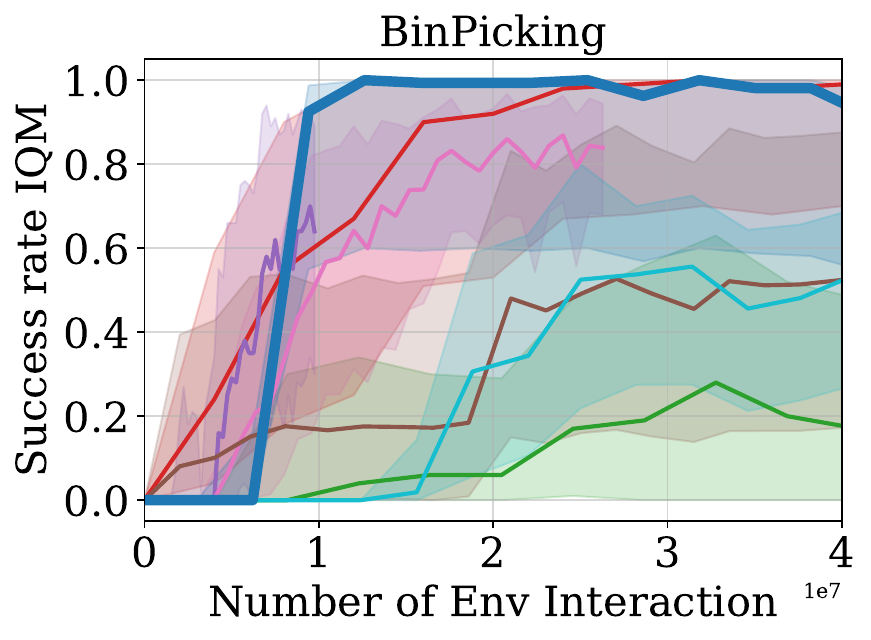}
  \end{subfigure}\hfill
  \begin{subfigure}{0.32\textwidth}
    \includegraphics[width=\linewidth]{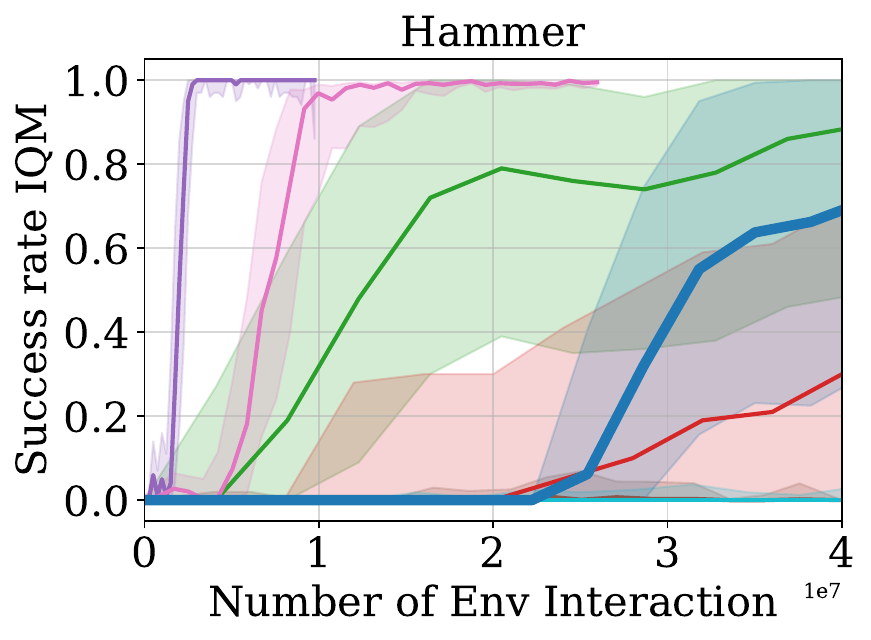}
  \end{subfigure}\hfill
  \begin{subfigure}{0.32\textwidth}
    \includegraphics[width=\linewidth]{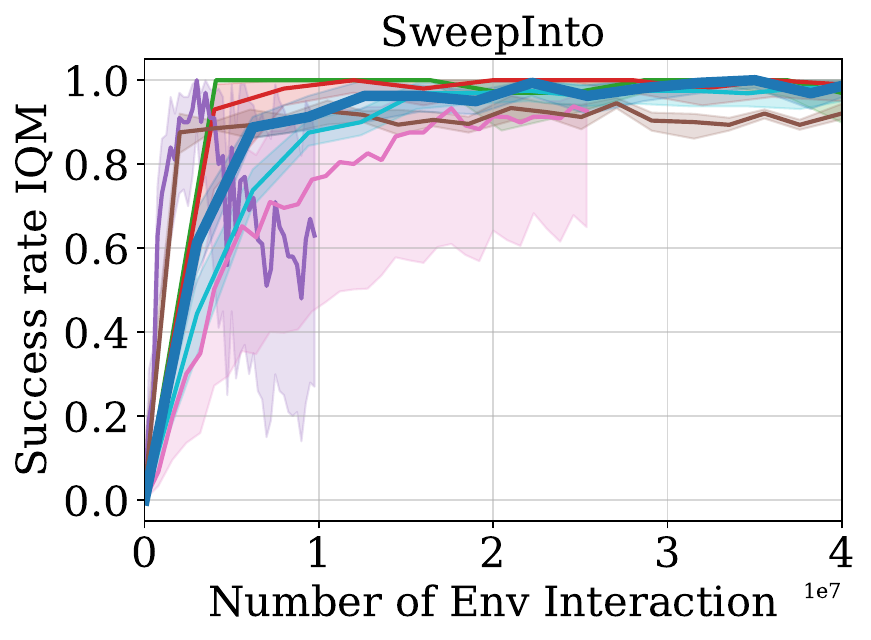}
  \end{subfigure}
  \vspace{0.1cm} 
  \begin{subfigure}{0.32\textwidth}
    \includegraphics[width=\linewidth]{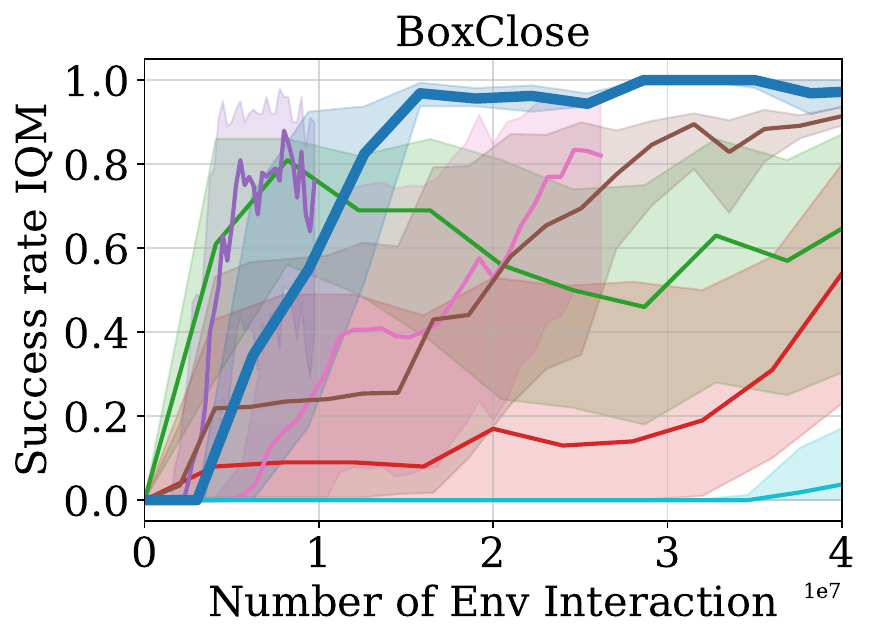}
  \end{subfigure}\hfill
  \begin{subfigure}{0.32\textwidth}
    \includegraphics[width=\linewidth]{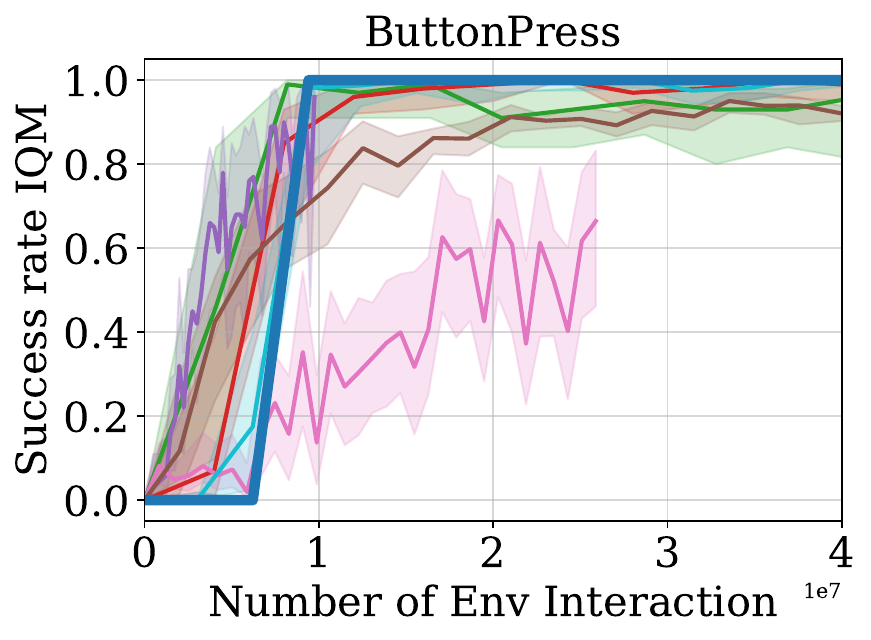}
  \end{subfigure}\hfill
  \begin{subfigure}{0.32\textwidth}
    \includegraphics[width=\linewidth]{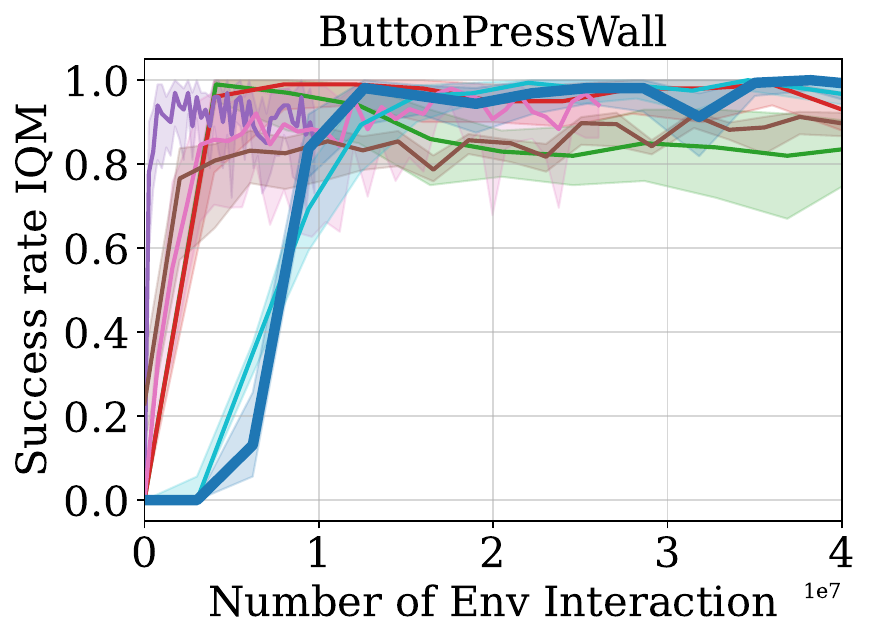}
  \end{subfigure}
  \vspace{0.1cm} 
  \begin{subfigure}{0.32\textwidth}
    \includegraphics[width=\linewidth]{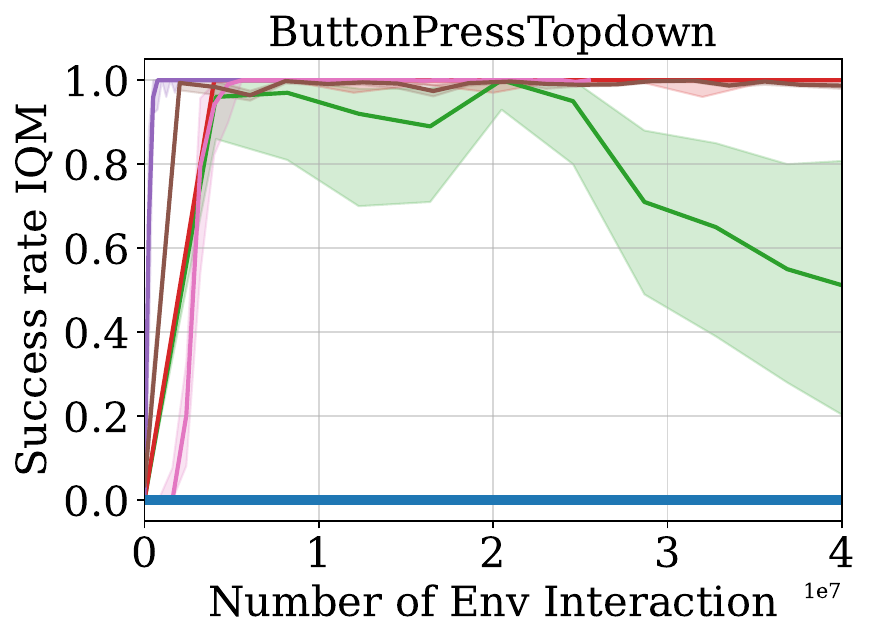}
  \end{subfigure}\hfill
  \begin{subfigure}{0.32\textwidth}
    \includegraphics[width=\linewidth]{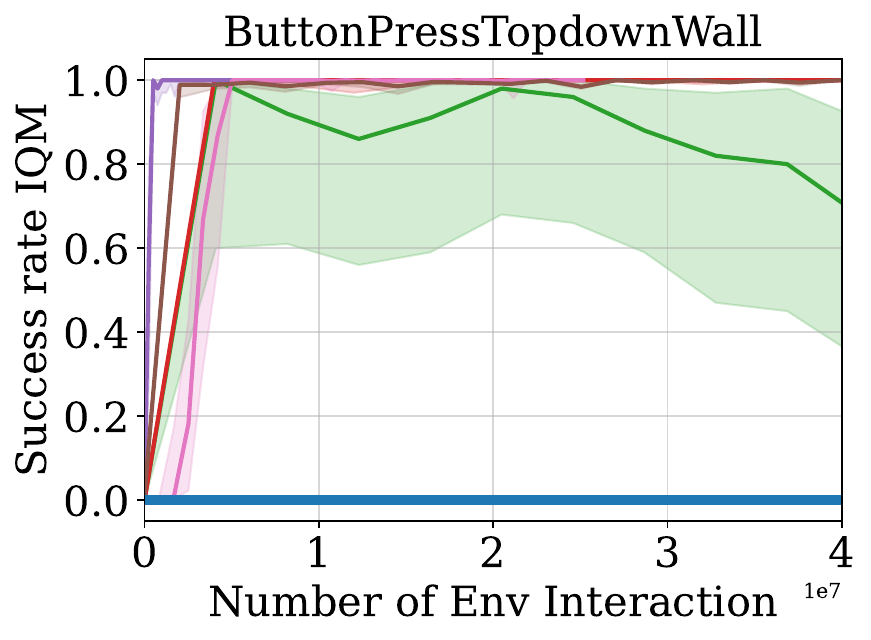}
  \end{subfigure}\hfill
  \begin{subfigure}{0.32\textwidth}
    \includegraphics[width=\linewidth]{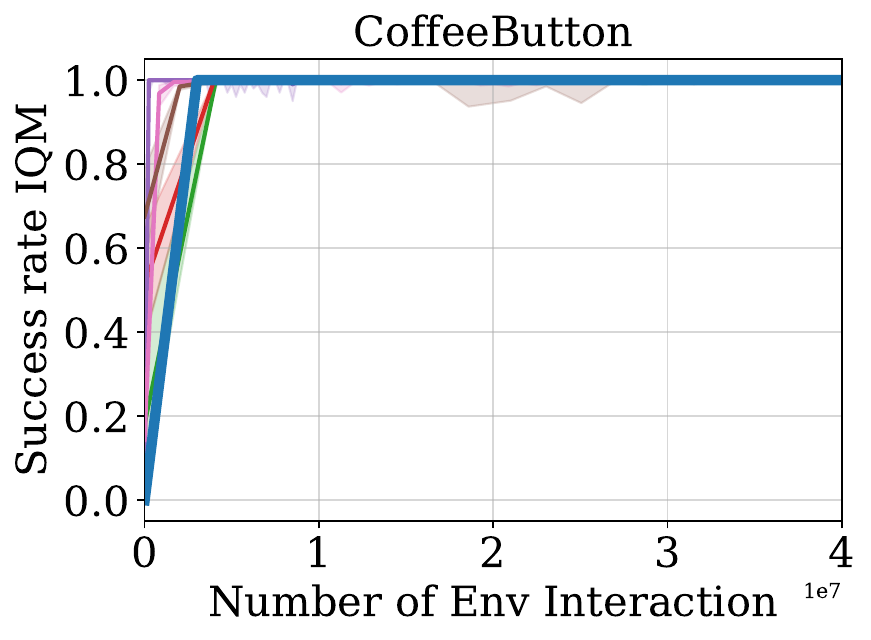}
  \end{subfigure}
  \vspace{0.1cm} 
  \caption{Success Rate IQM of each individual Metaworld tasks.}
  \label{fig:your-label}
\end{figure}
\newpage
\begin{figure}[t!]
    \hspace*{\fill}%
    \resizebox{0.8\textwidth}{!}{    
        \begin{tikzpicture} 
\def\linewidthtcp{1mm}
\def\linewidthothers{0.5mm}

    \begin{axis}[%
    hide axis,
    xmin=10,
    xmax=50,
    ymin=0,
    ymax=0.4,
    legend style={
        draw=white!15!black,
        legend cell align=left,
        legend columns=-1, 
        legend style={
            draw=none,
            column sep=1ex,
            line width=1pt
        }
    },
    ]
    \addlegendimage{C0, line width=\linewidthtcp}
    \addlegendentry{\acrshort{tcp} (ours)};    
    \addlegendimage{C9, line width=\linewidthothers}
    \addlegendentry{\acrshort{bbrl}};
    \addlegendimage{C2, line width=\linewidthothers}
    \addlegendentry{\acrshort{ppo}};
    \addlegendimage{C3, line width=\linewidthothers}
    \addlegendentry{\acrshort{trpl}};
    \addlegendimage{C4, line width=\linewidthothers}
    \addlegendentry{\acrshort{sac}};    
    \addlegendimage{C5, line width=\linewidthothers}
    \addlegendentry{\acrshort{gsde}};
    \addlegendimage{C6, line width=\linewidthothers}
    \addlegendentry{\acrshort{pink}};

    \end{axis}
\end{tikzpicture}
    }%
    \hspace*{\fill}%
    \newline
    
  \centering
  \begin{subfigure}{0.32\textwidth}
    \includegraphics[width=\linewidth]{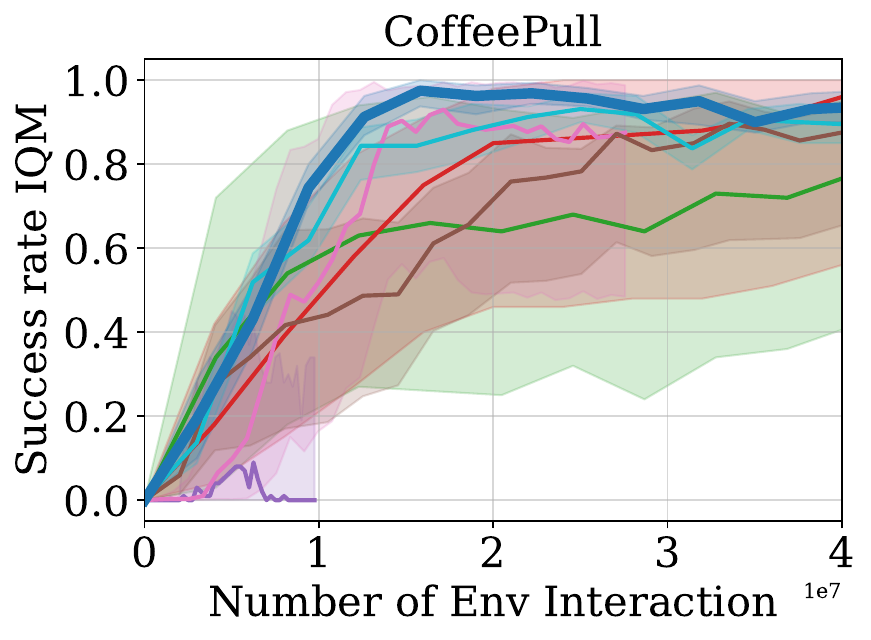}
  \end{subfigure}\hfill
  \begin{subfigure}{0.32\textwidth}
    \includegraphics[width=\linewidth]{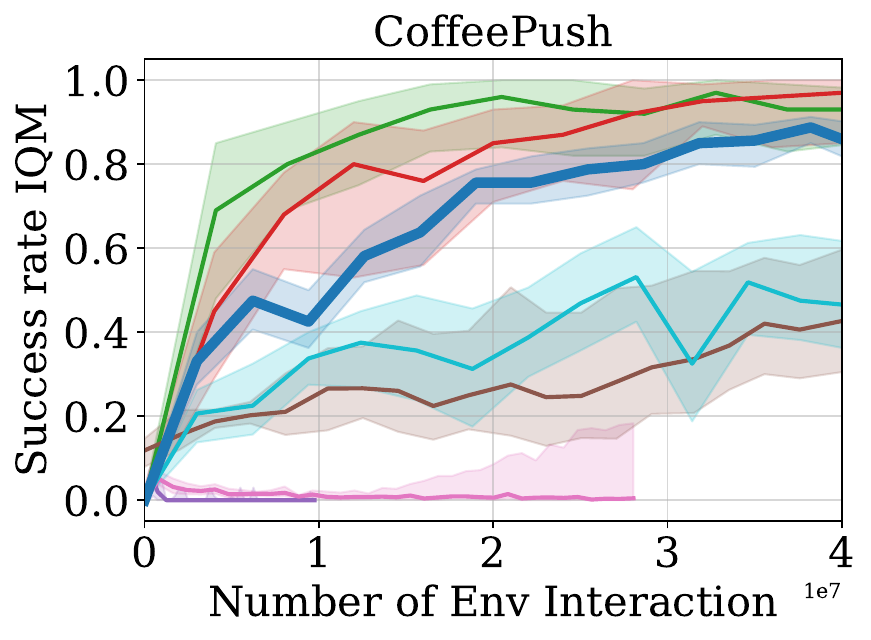}
  \end{subfigure}\hfill
  \begin{subfigure}{0.32\textwidth}
    \includegraphics[width=\linewidth]{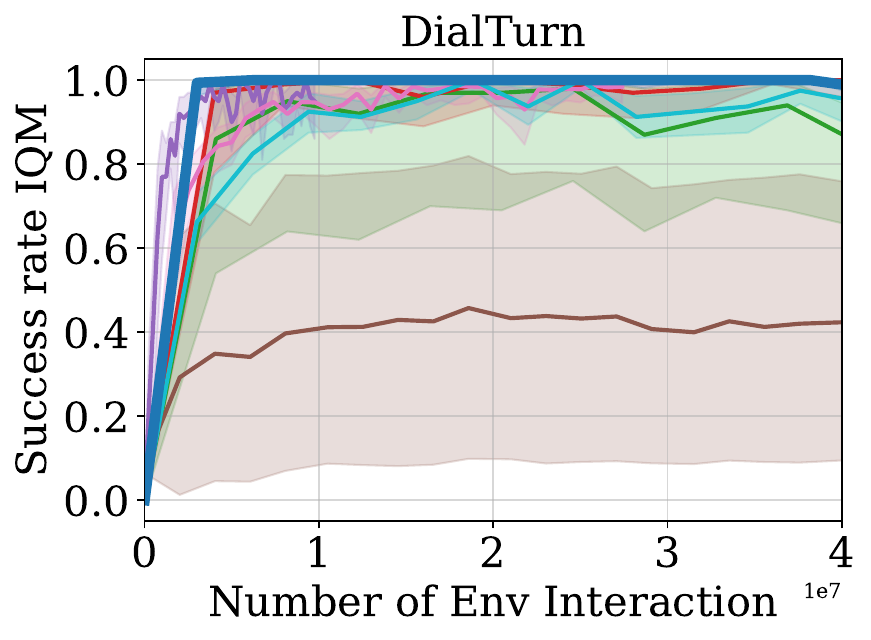}
  \end{subfigure}
  \vspace{0.1cm}   
  \begin{subfigure}{0.32\textwidth}
    \includegraphics[width=\linewidth]{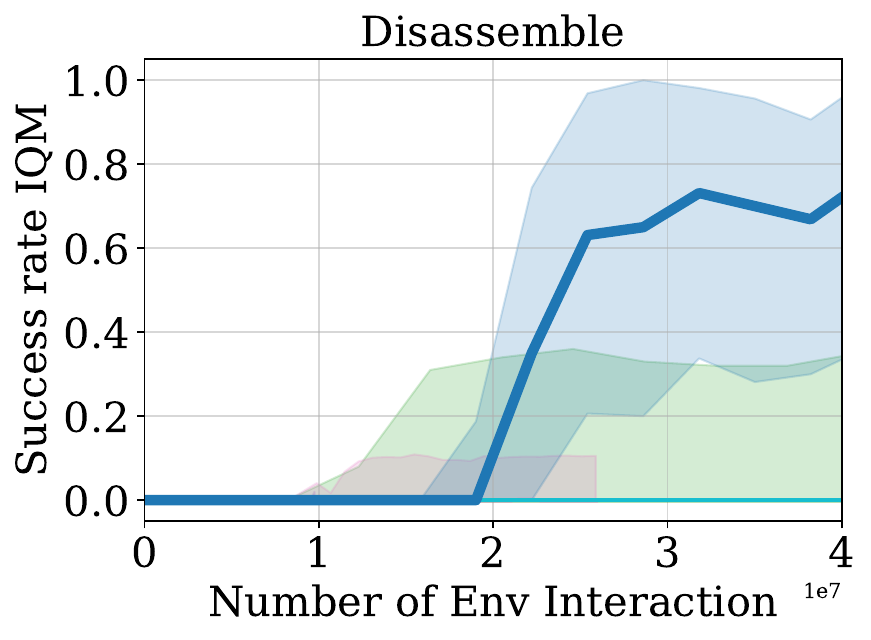}
  \end{subfigure}\hfill
  \begin{subfigure}{0.32\textwidth}
    \includegraphics[width=\linewidth]{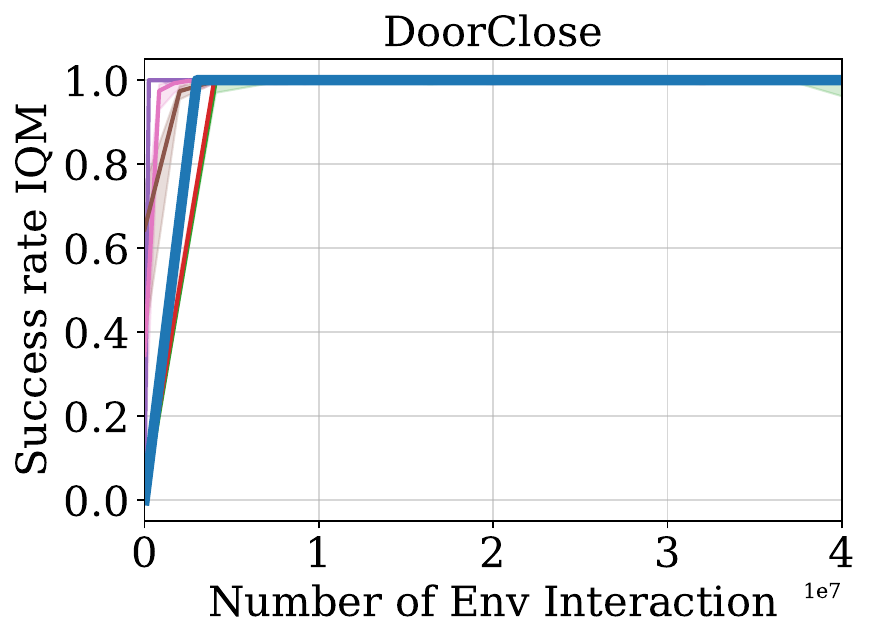}
  \end{subfigure}\hfill
  \begin{subfigure}{0.32\textwidth}
    \includegraphics[width=\linewidth]{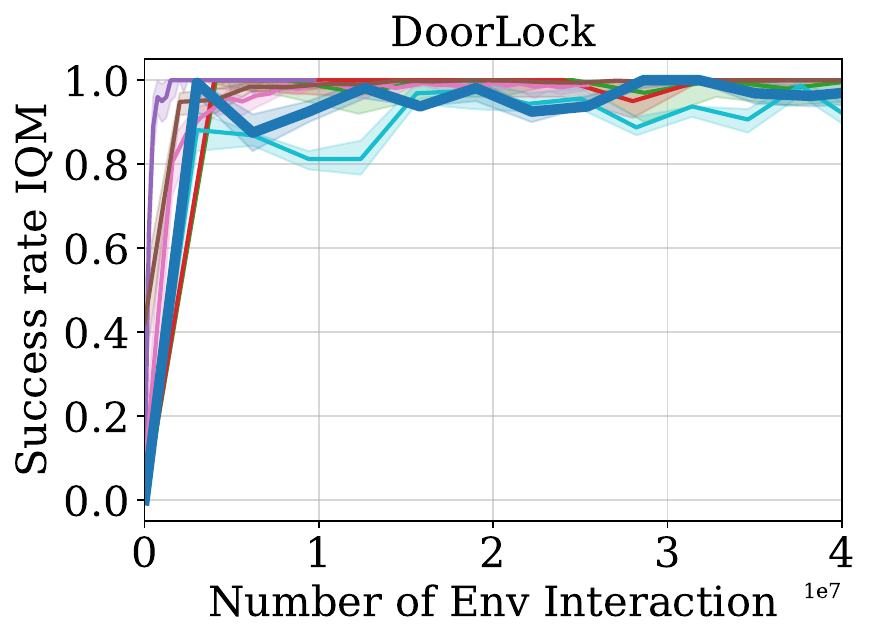}
  \end{subfigure}
  \vspace{0.1cm} 
  \begin{subfigure}{0.32\textwidth}
    \includegraphics[width=\linewidth]{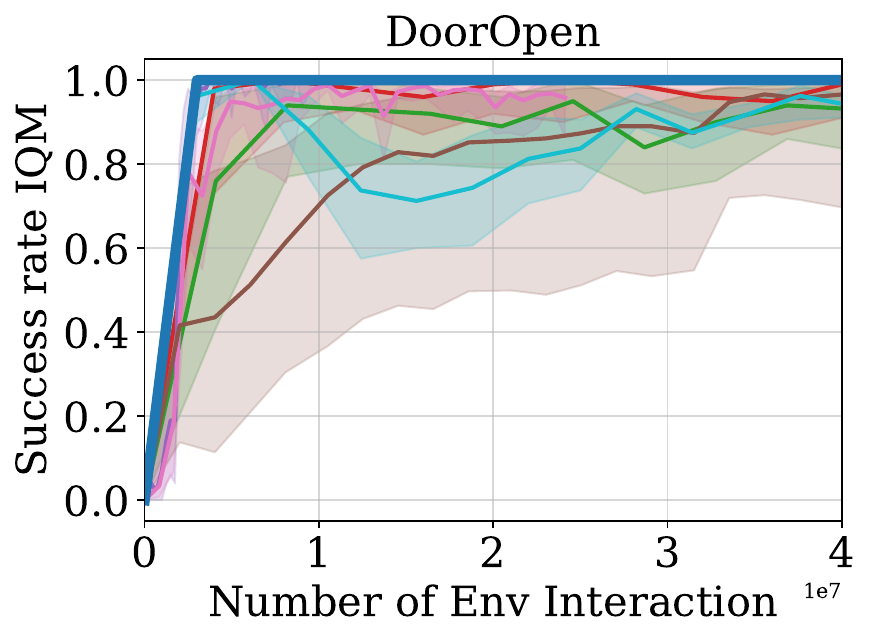}
  \end{subfigure}\hfill
  \begin{subfigure}{0.32\textwidth}
    \includegraphics[width=\linewidth]{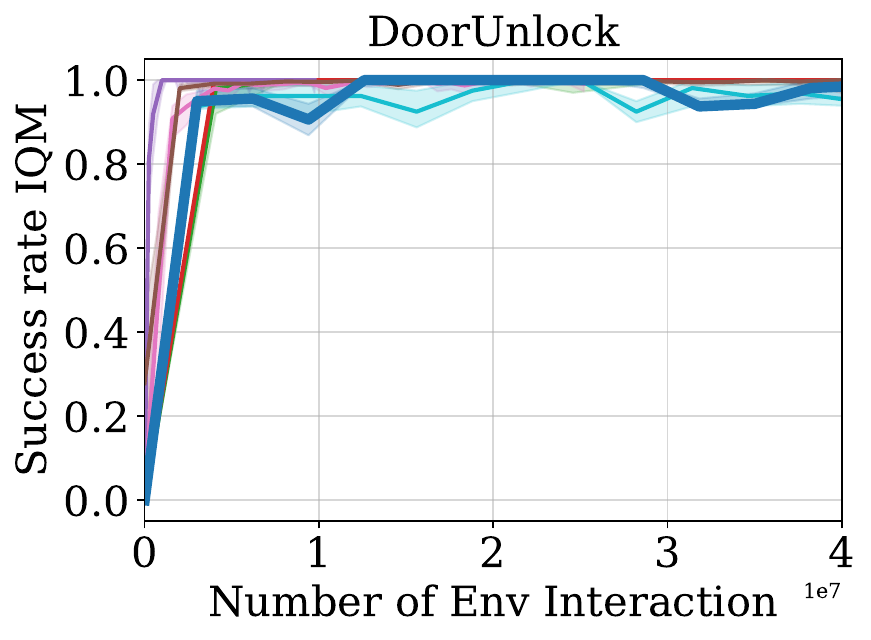}
  \end{subfigure}\hfill
  \begin{subfigure}{0.32\textwidth}
    \includegraphics[width=\linewidth]{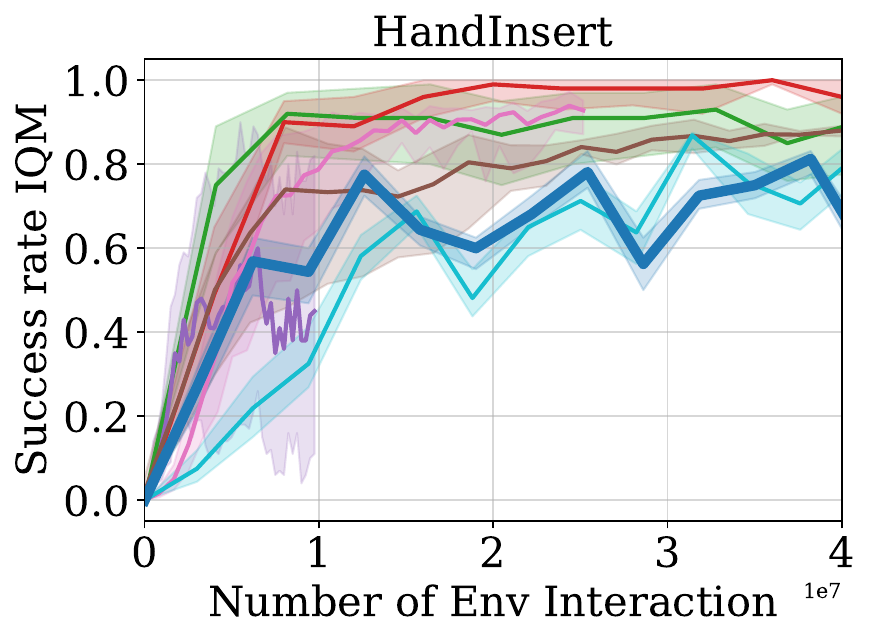}
  \end{subfigure}
  \vspace{0.1cm} 
  \begin{subfigure}{0.32\textwidth}
    \includegraphics[width=\linewidth]{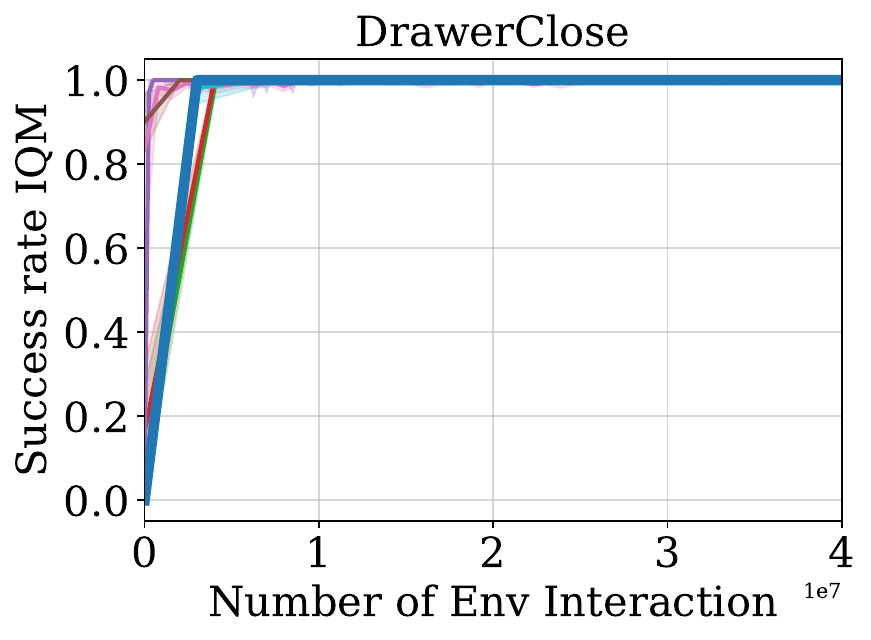}
  \end{subfigure}\hfill
  \begin{subfigure}{0.32\textwidth}
    \includegraphics[width=\linewidth]{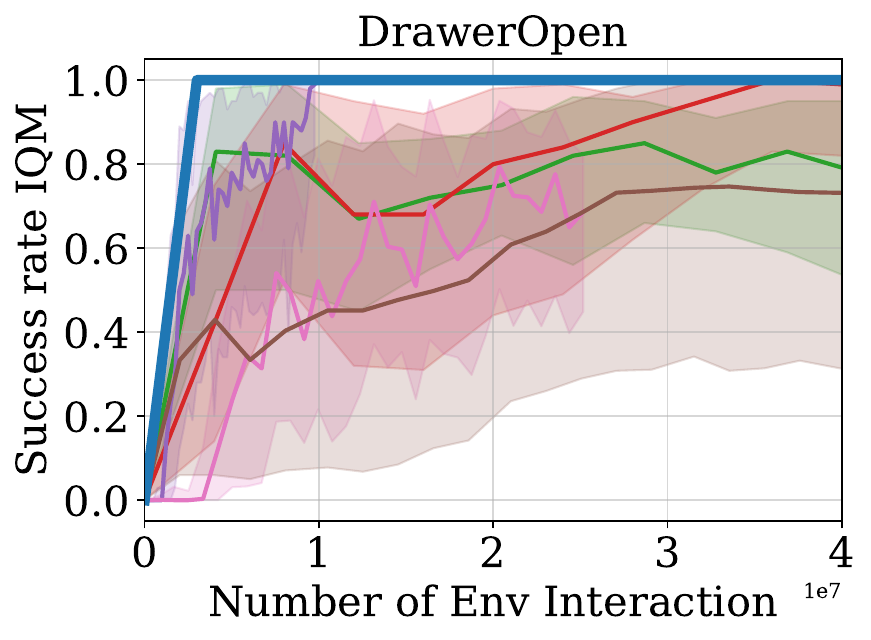}
  \end{subfigure}\hfill
  \begin{subfigure}{0.32\textwidth}
    \includegraphics[width=\linewidth]{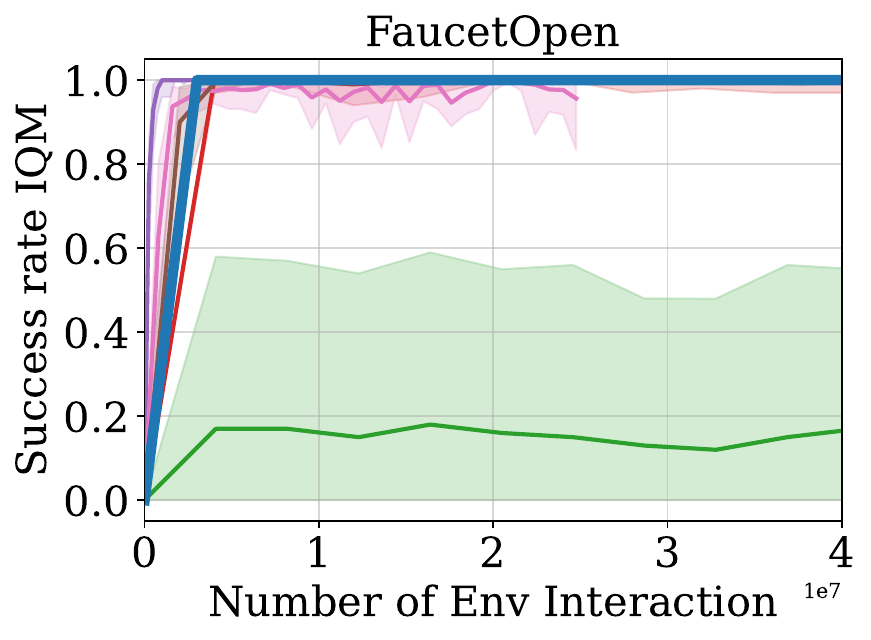}
  \end{subfigure}
  \vspace{0.1cm} 
  \begin{subfigure}{0.32\textwidth}
    \includegraphics[width=\linewidth]{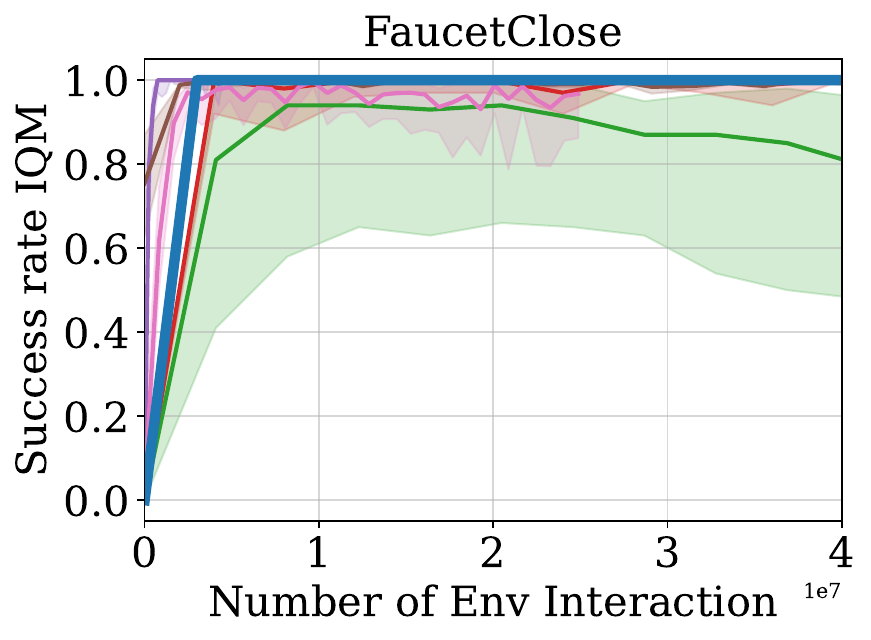}
  \end{subfigure}\hfill
  \begin{subfigure}{0.32\textwidth}
    \includegraphics[width=\linewidth]{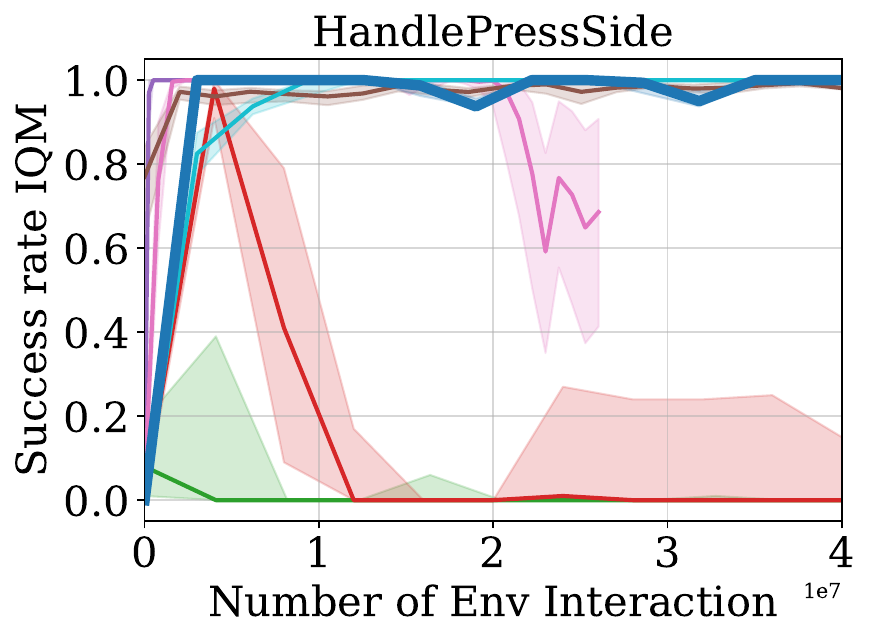}
  \end{subfigure}\hfill
  \begin{subfigure}{0.32\textwidth}
    \includegraphics[width=\linewidth]{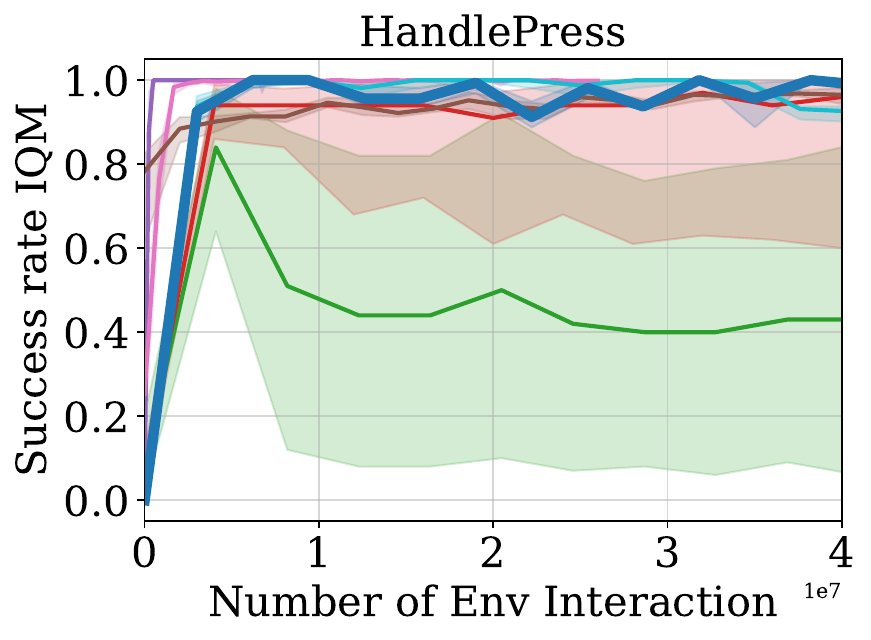}
  \end{subfigure}
  \vspace{0.1cm} 
  \begin{subfigure}{0.32\textwidth}
    \includegraphics[width=\linewidth]{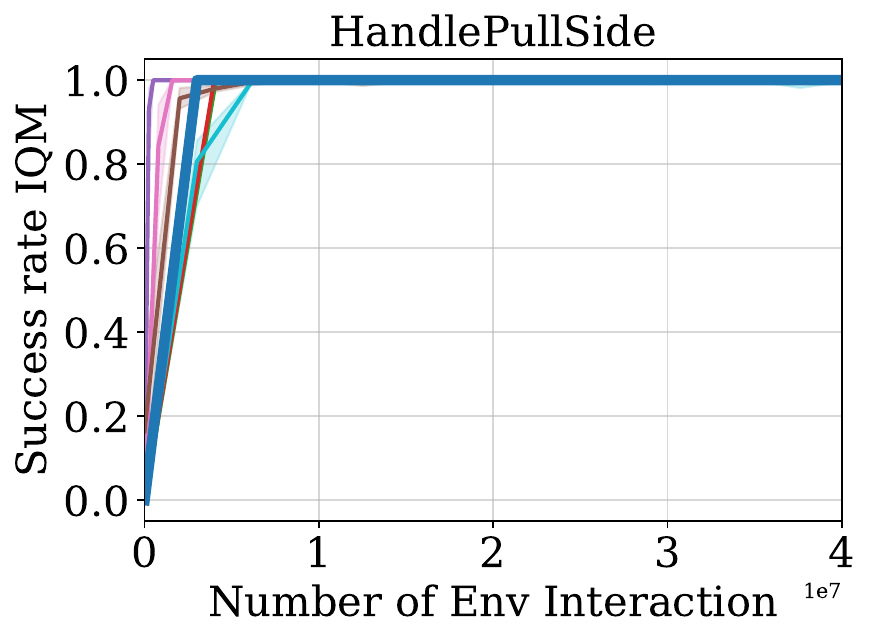}
  \end{subfigure}\hfill
  \begin{subfigure}{0.32\textwidth}
    \includegraphics[width=\linewidth]{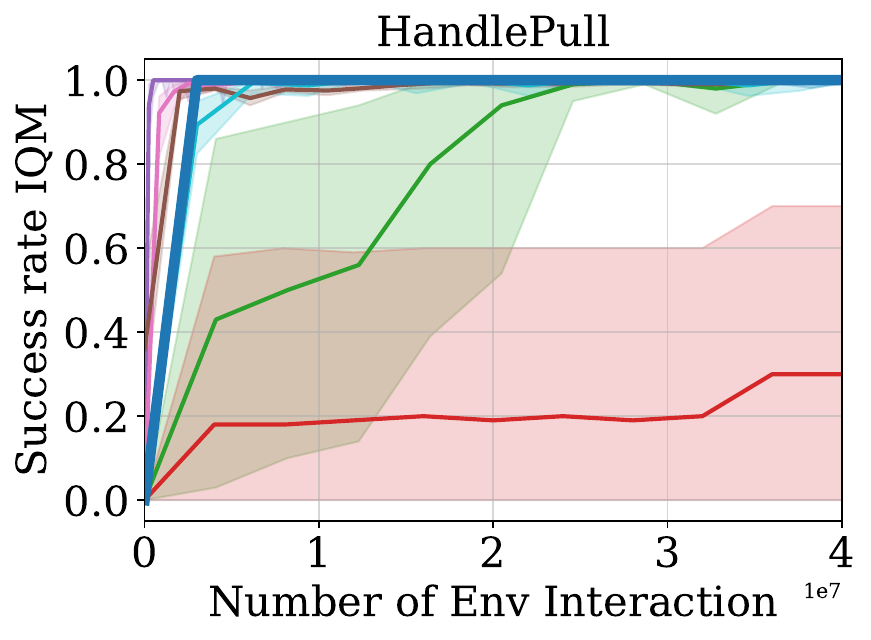}
  \end{subfigure}\hfill
  \begin{subfigure}{0.32\textwidth}
    \includegraphics[width=\linewidth]{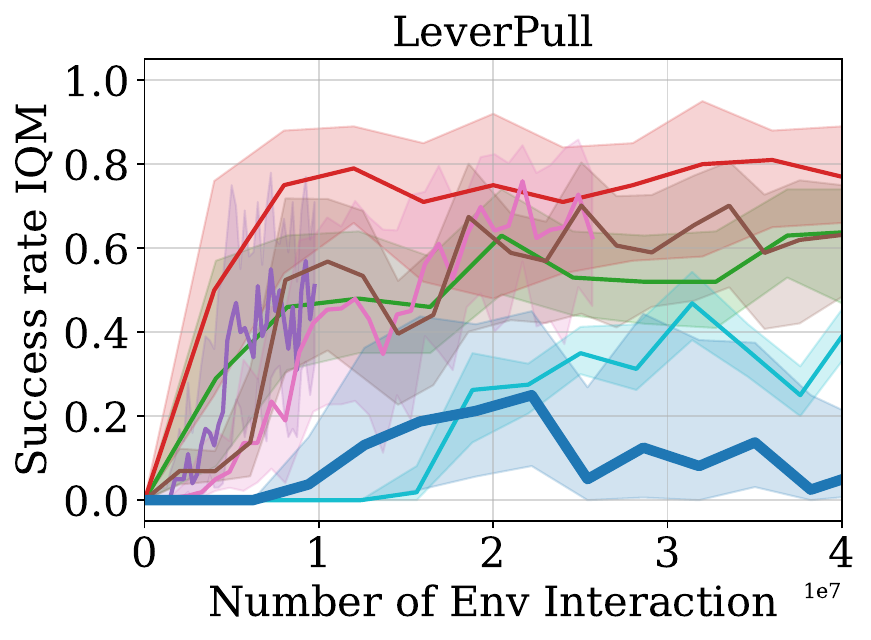}
  \end{subfigure}
  \vspace{0.1cm} 
  \caption{Success Rate IQM of each individual Metaworld tasks.}
  \label{fig:your-label}
\end{figure}
\newpage
\newpage
\begin{figure}[t!]
    \hspace*{\fill}%
    \resizebox{0.8\textwidth}{!}{    
        \begin{tikzpicture} 
\def\linewidthtcp{1mm}
\def\linewidthothers{0.5mm}

    \begin{axis}[%
    hide axis,
    xmin=10,
    xmax=50,
    ymin=0,
    ymax=0.4,
    legend style={
        draw=white!15!black,
        legend cell align=left,
        legend columns=-1, 
        legend style={
            draw=none,
            column sep=1ex,
            line width=1pt
        }
    },
    ]
    \addlegendimage{C0, line width=\linewidthtcp}
    \addlegendentry{\acrshort{tcp} (ours)};    
    \addlegendimage{C9, line width=\linewidthothers}
    \addlegendentry{\acrshort{bbrl}};
    \addlegendimage{C2, line width=\linewidthothers}
    \addlegendentry{\acrshort{ppo}};
    \addlegendimage{C3, line width=\linewidthothers}
    \addlegendentry{\acrshort{trpl}};
    \addlegendimage{C4, line width=\linewidthothers}
    \addlegendentry{\acrshort{sac}};    
    \addlegendimage{C5, line width=\linewidthothers}
    \addlegendentry{\acrshort{gsde}};
    \addlegendimage{C6, line width=\linewidthothers}
    \addlegendentry{\acrshort{pink}};

    \end{axis}
\end{tikzpicture}
    }%
    \hspace*{\fill}%
    \newline
    
  \centering
  \begin{subfigure}{0.32\textwidth}
    \includegraphics[width=\linewidth]{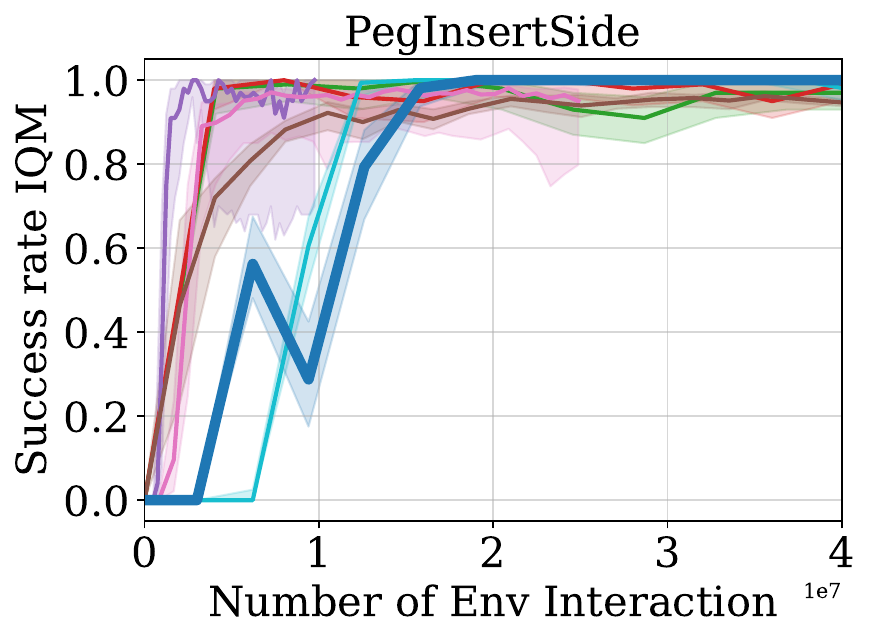}
  \end{subfigure}\hfill
  \begin{subfigure}{0.32\textwidth}
    \includegraphics[width=\linewidth]{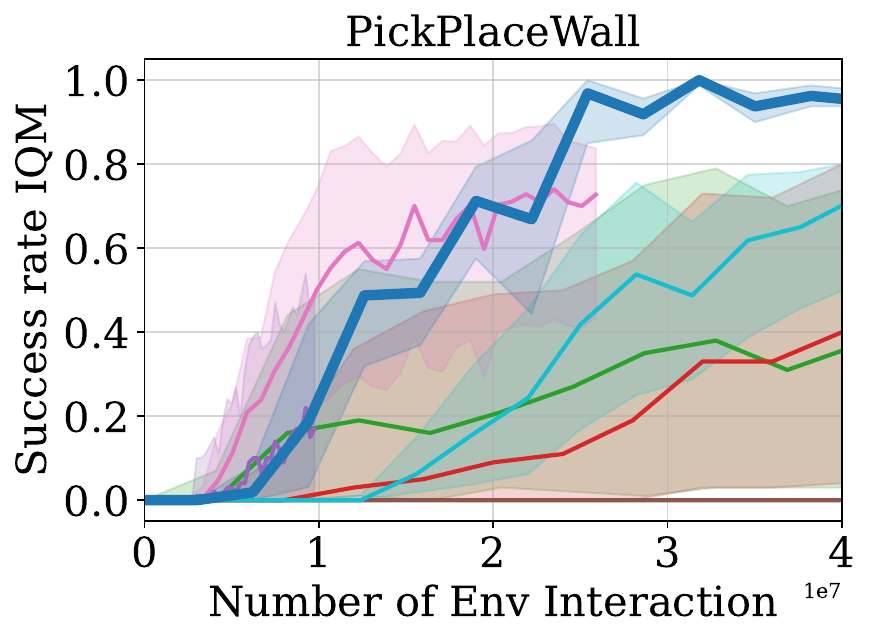}
  \end{subfigure}\hfill
  \begin{subfigure}{0.32\textwidth}
    \includegraphics[width=\linewidth]{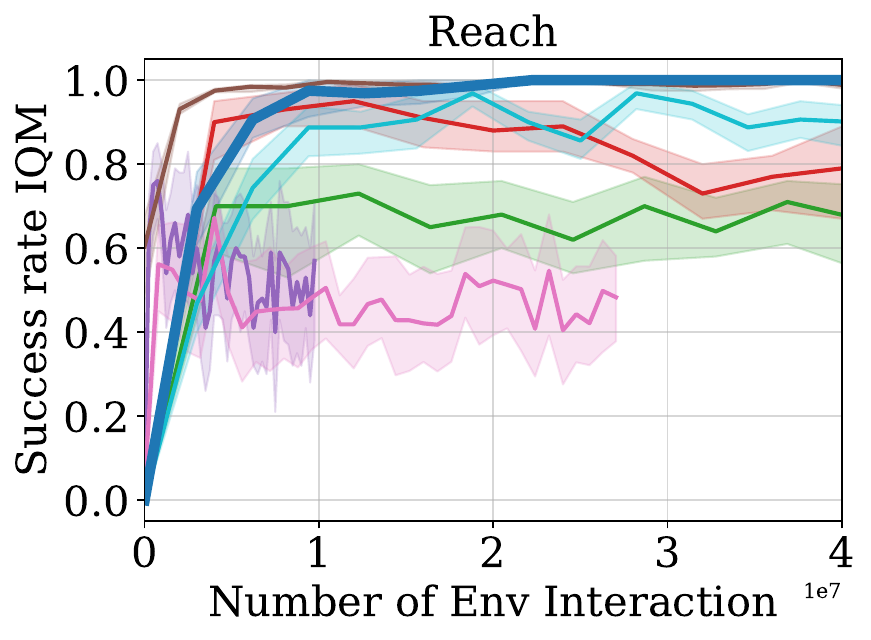}
  \end{subfigure}
  \vspace{0.1cm} 
  \begin{subfigure}{0.32\textwidth}
    \includegraphics[width=\linewidth]{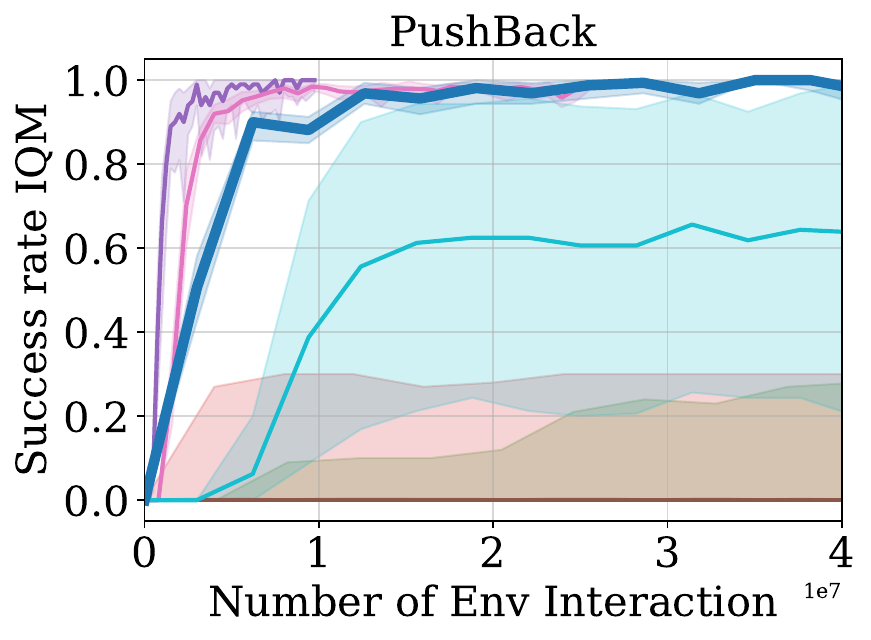}
  \end{subfigure}\hfill
  \begin{subfigure}{0.32\textwidth}
    \includegraphics[width=\linewidth]{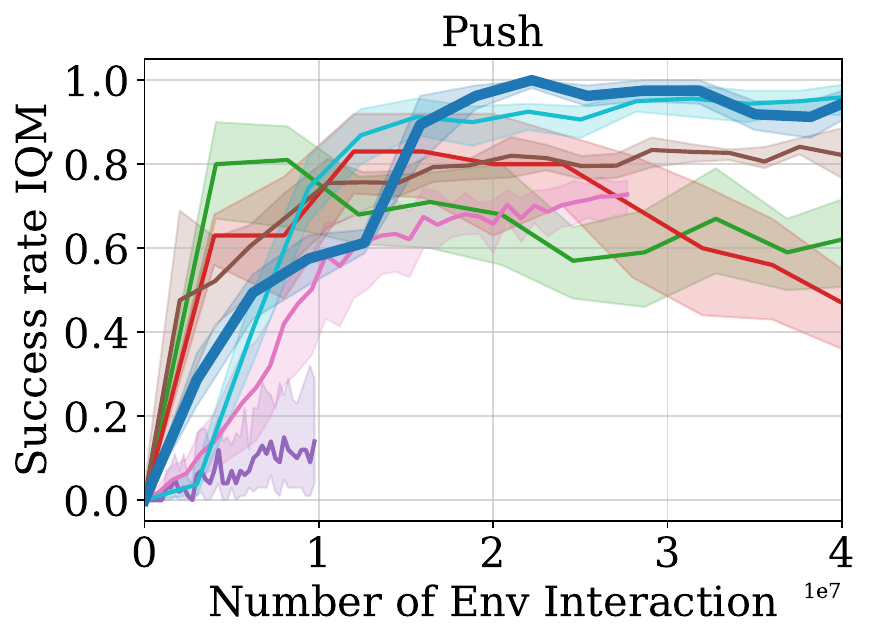}
  \end{subfigure}\hfill
  \begin{subfigure}{0.32\textwidth}
    \includegraphics[width=\linewidth]{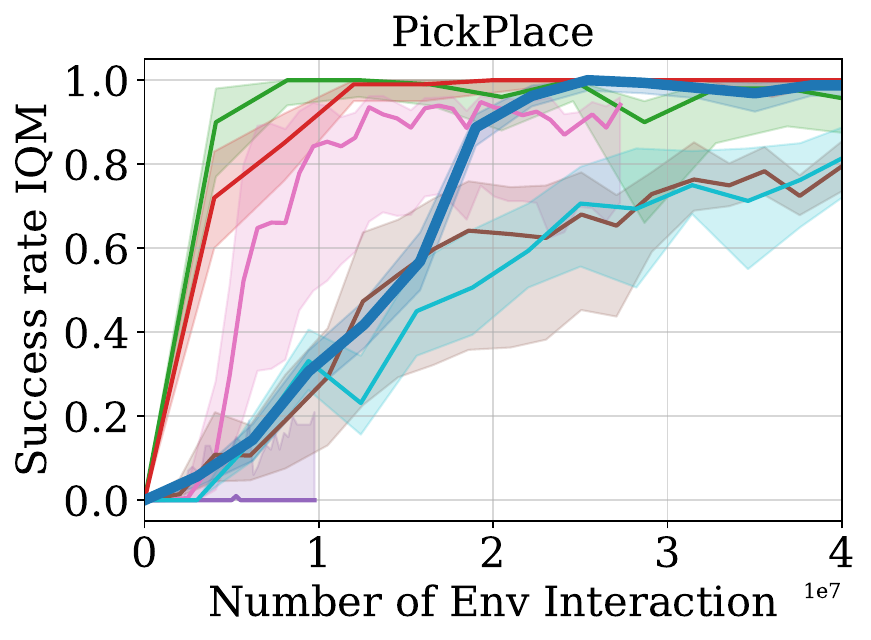}
  \end{subfigure}
  \vspace{0.1cm} 
  \begin{subfigure}{0.32\textwidth}
    \includegraphics[width=\linewidth]{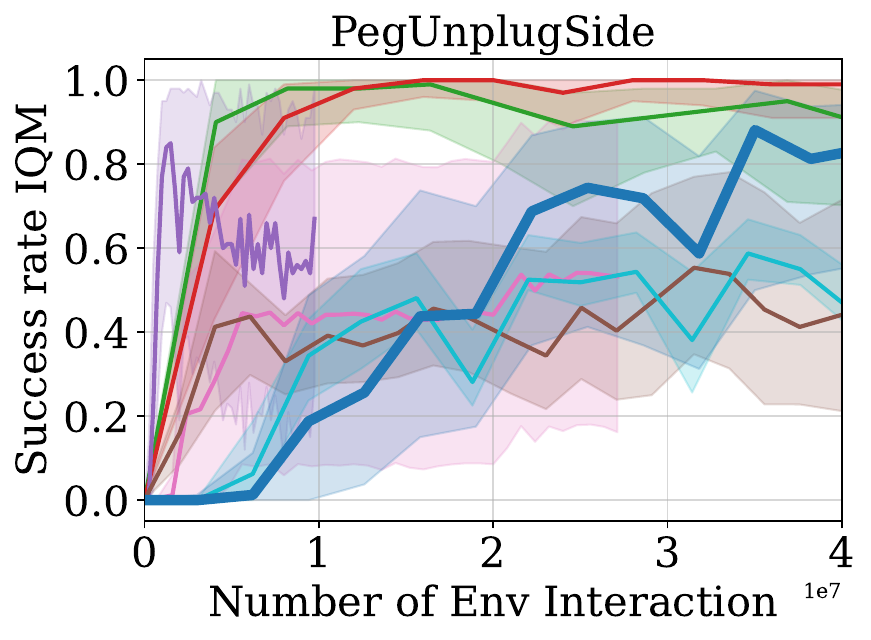}
  \end{subfigure}\hfill
  \begin{subfigure}{0.32\textwidth}
    \includegraphics[width=\linewidth]{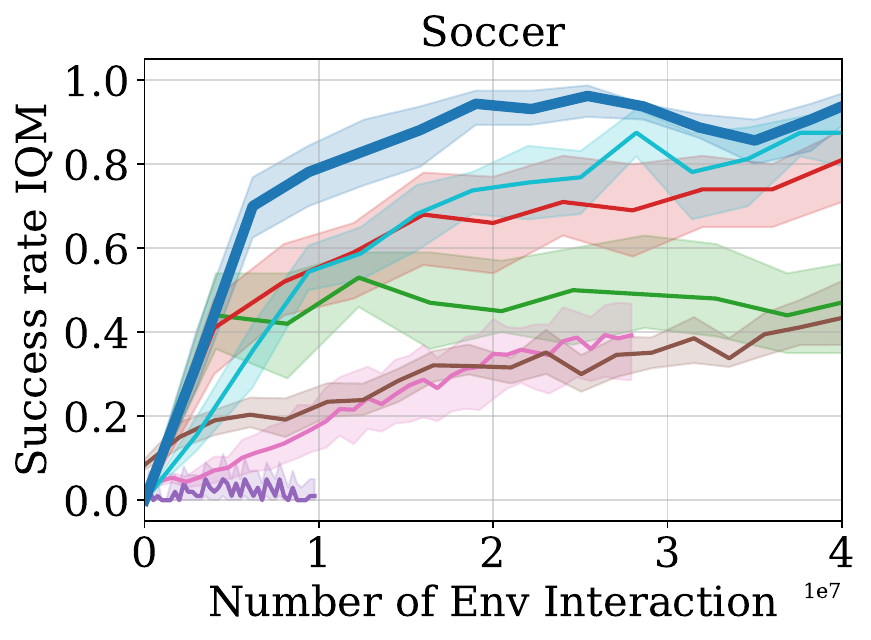}
  \end{subfigure}\hfill
  \begin{subfigure}{0.32\textwidth}
    \includegraphics[width=\linewidth]{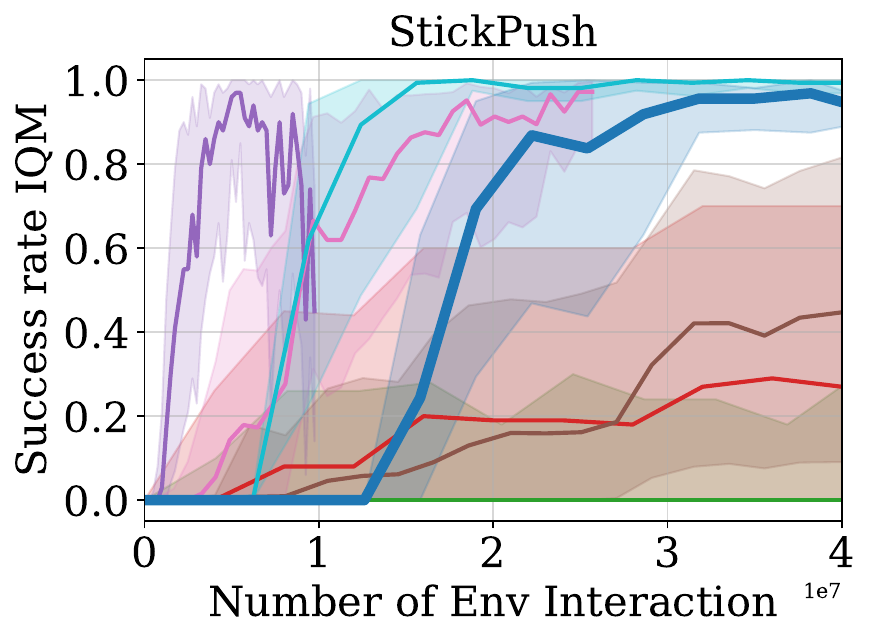}
  \end{subfigure}
  \vspace{0.1cm} 
  \begin{subfigure}{0.32\textwidth}
    \includegraphics[width=\linewidth]{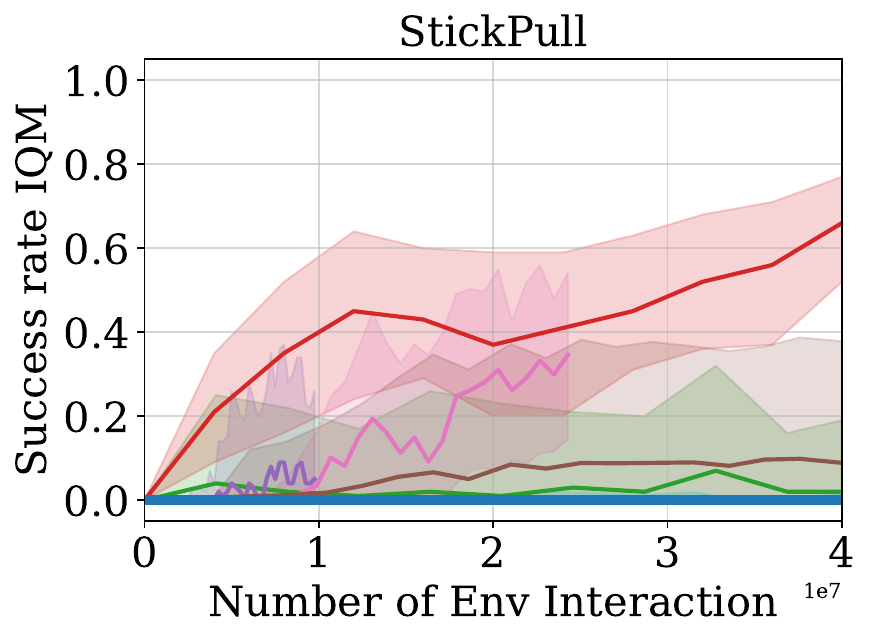}
  \end{subfigure}\hfill
  \begin{subfigure}{0.32\textwidth}
    \includegraphics[width=\linewidth]{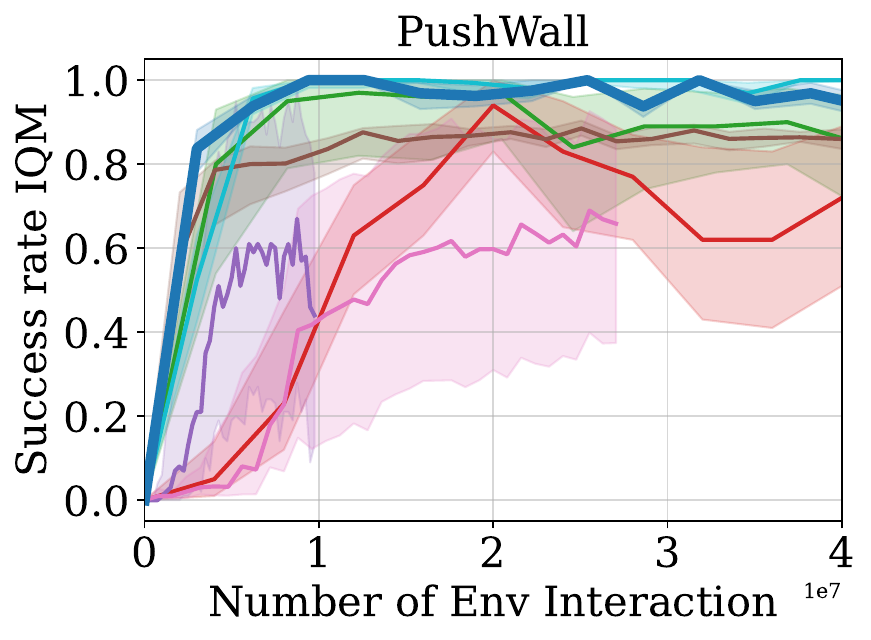}
  \end{subfigure}\hfill
  \begin{subfigure}{0.32\textwidth}
    \includegraphics[width=\linewidth]{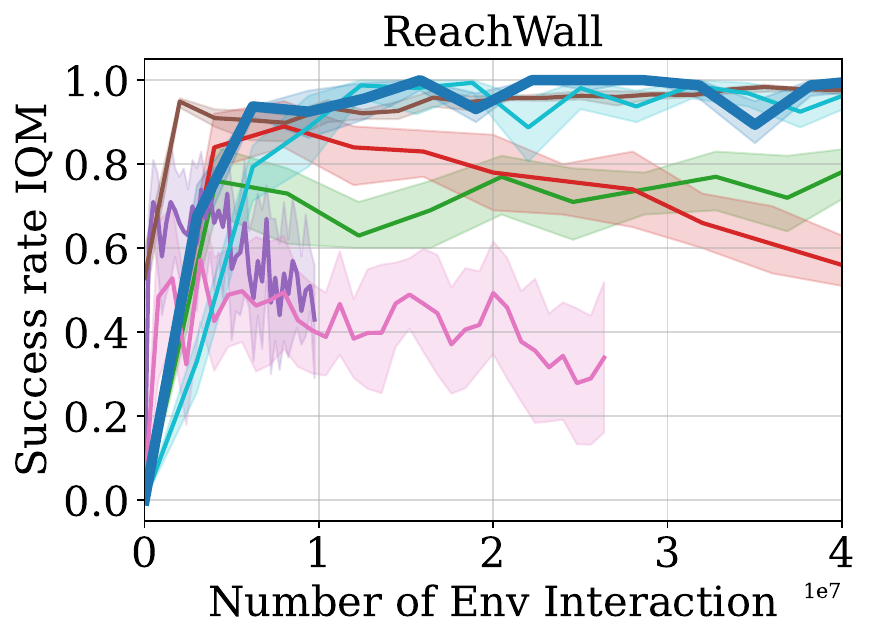}
  \end{subfigure}
  \vspace{0.1cm} 
  \begin{subfigure}{0.32\textwidth}
    \includegraphics[width=\linewidth]{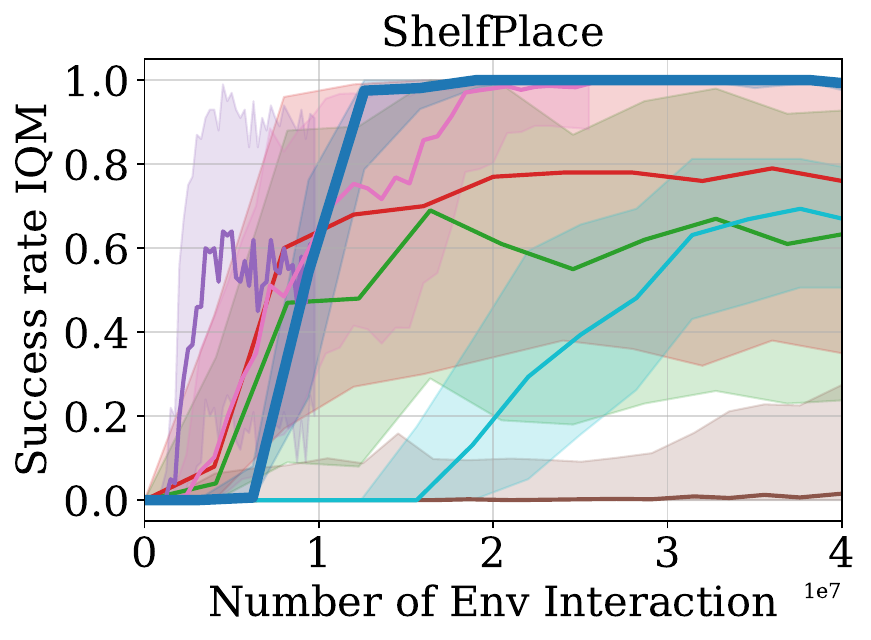}
  \end{subfigure}\hfill
  \begin{subfigure}{0.32\textwidth}
    \includegraphics[width=\linewidth]{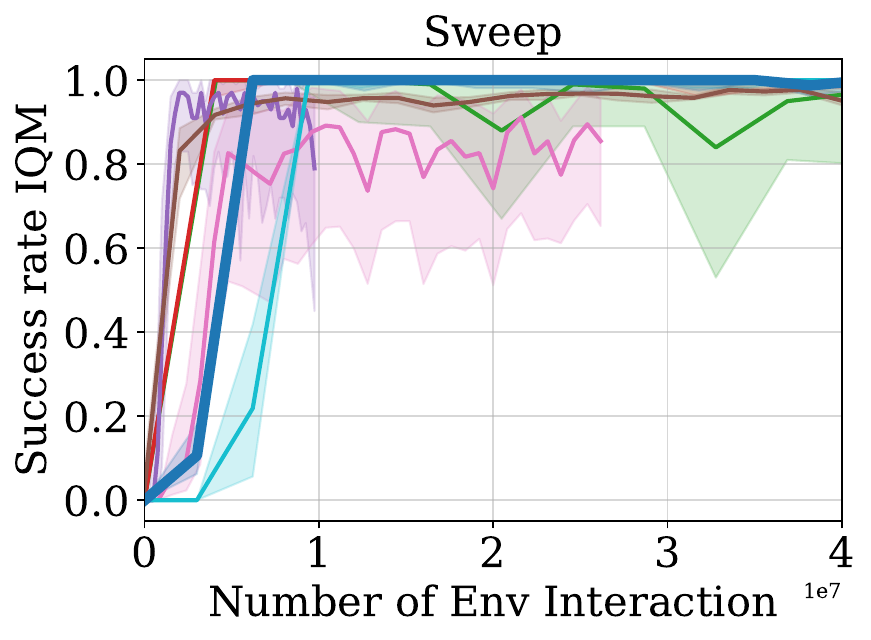}
  \end{subfigure}\hfill
  \begin{subfigure}{0.32\textwidth}
    \includegraphics[width=\linewidth]{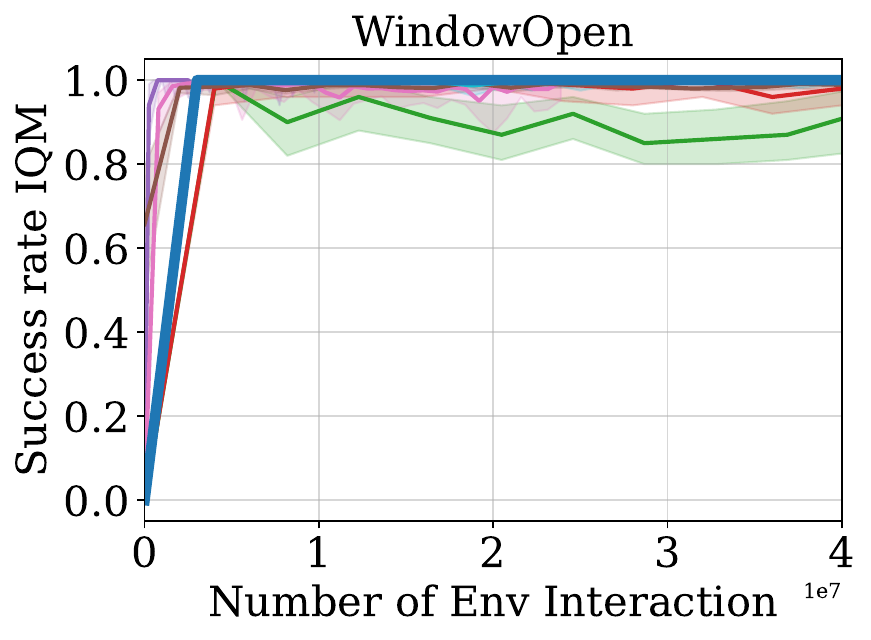}
  \end{subfigure}
  \vspace{0.1cm} 
  \begin{subfigure}{0.32\textwidth}
    \includegraphics[width=\linewidth]{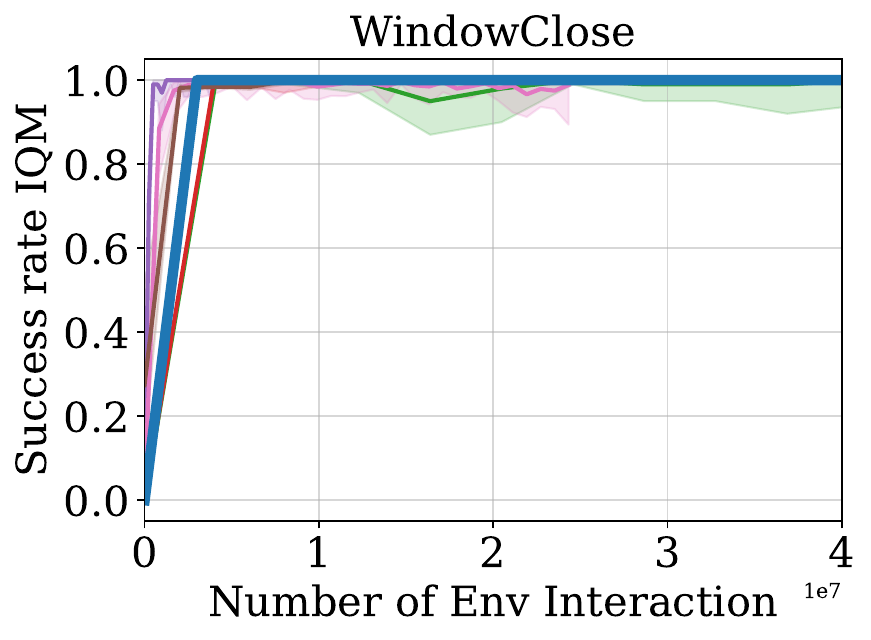}
  \end{subfigure}
  \begin{subfigure}{0.32\textwidth}
    \includegraphics[width=\linewidth]{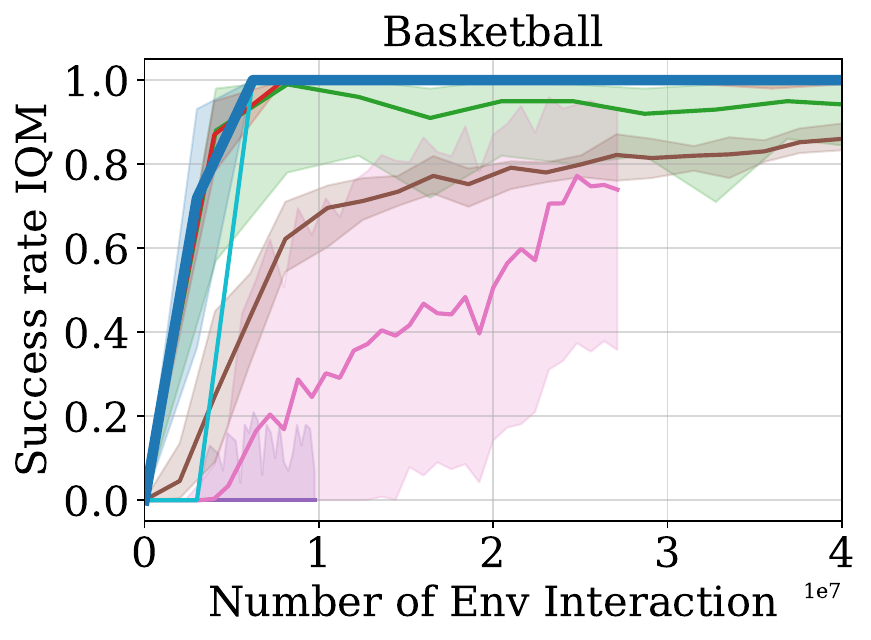}
  \end{subfigure}\hfill
  
  \vspace{0.1cm} 
  \caption{Success Rate IQM of each individual Metaworld tasks.}
  \label{fig:your-label}
\end{figure}
\afterpage{\newpage}

\subsection{Hopper Jump}
\label{appsub:hopper}
\begin{figure}[h!]
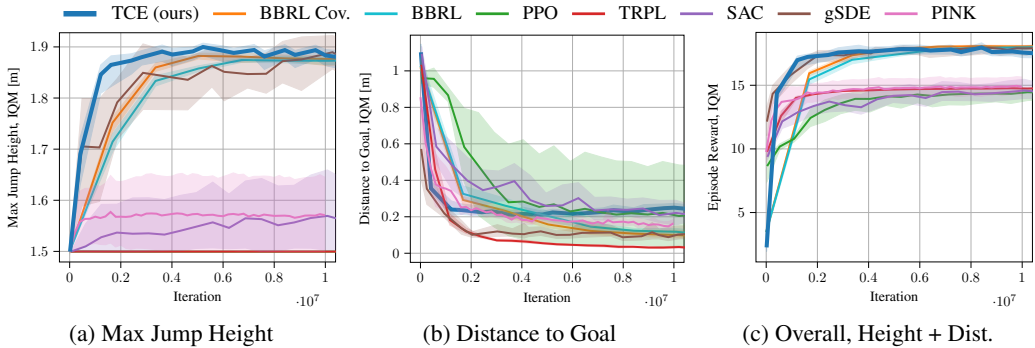
    
    \hspace*{\fill}%
    \resizebox{0.9\textwidth}{!}{    
        \begin{tikzpicture} 
\def\linewidthtcp{1mm}
\def\linewidthothers{0.5mm}

    \begin{axis}[%
    hide axis,
    xmin=10,
    xmax=50,
    ymin=0,
    ymax=0.4,
    legend style={
        draw=white!15!black,
        legend cell align=left,
        legend columns=-1, 
        legend style={
            draw=none,
            column sep=1ex,
            line width=1pt
        }
    },
    ]
    \addlegendimage{C0, line width=\linewidthtcp}
    \addlegendentry{\acrshort{tcp} (ours)};
    \addlegendimage{C1, line width=\linewidthothers}
    \addlegendentry{\acrshort{bbrl} Cov.};
    \addlegendimage{C9, line width=\linewidthothers}
    \addlegendentry{\acrshort{bbrl}};    
    \addlegendimage{C2, line width=\linewidthothers}
    \addlegendentry{\acrshort{ppo}};
    \addlegendimage{C3, line width=\linewidthothers}
    \addlegendentry{\acrshort{trpl}};
    \addlegendimage{C4, line width=\linewidthothers}
    \addlegendentry{\acrshort{sac}};    
    \addlegendimage{C5, line width=\linewidthothers}
    \addlegendentry{\acrshort{gsde}};
    \addlegendimage{C6, line width=\linewidthothers}
    \addlegendentry{\acrshort{pink}};

    \end{axis}
\end{tikzpicture}
    }%
    \hspace*{\fill}%
    \newline
    \centering
    \hspace*{\fill}%
    \begin{subfigure}{0.32\textwidth}        
        \resizebox{\textwidth}{!}{\input{figure/hopper_jump/plot/height}}%
        \caption{Max Jump Height}
        \label{subfig:hopper_height_appendix}
    \end{subfigure}
    \hfill%
    \begin{subfigure}{0.32\textwidth}        
        \resizebox{\textwidth}{!}{        
        \input{figure/hopper_jump/plot/to_goal}}%
        \caption{Distance to Goal}
        \label{subfig:hopper_goal}
    \end{subfigure}
    \hfill%
    \begin{subfigure}{0.32\textwidth}        
        \resizebox{\textwidth}{!}{        
        \input{figure/hopper_jump/plot/reward}}%
        \caption{Overall, Height + Dist.}
        \label{subfig:hopper_reward}
    \end{subfigure}
    \hspace*{\fill}%
    \caption{Hopper Jump}
    \label{fig:hopper_jump}
\end{figure}

\begin{wrapfigure}{r}{0.20\textwidth}
    \vspace{-0.5cm}
    \hspace*{\fill}%
    \begin{subfigure}{0.20\textwidth}
        \resizebox{\textwidth}{!}{\includegraphics{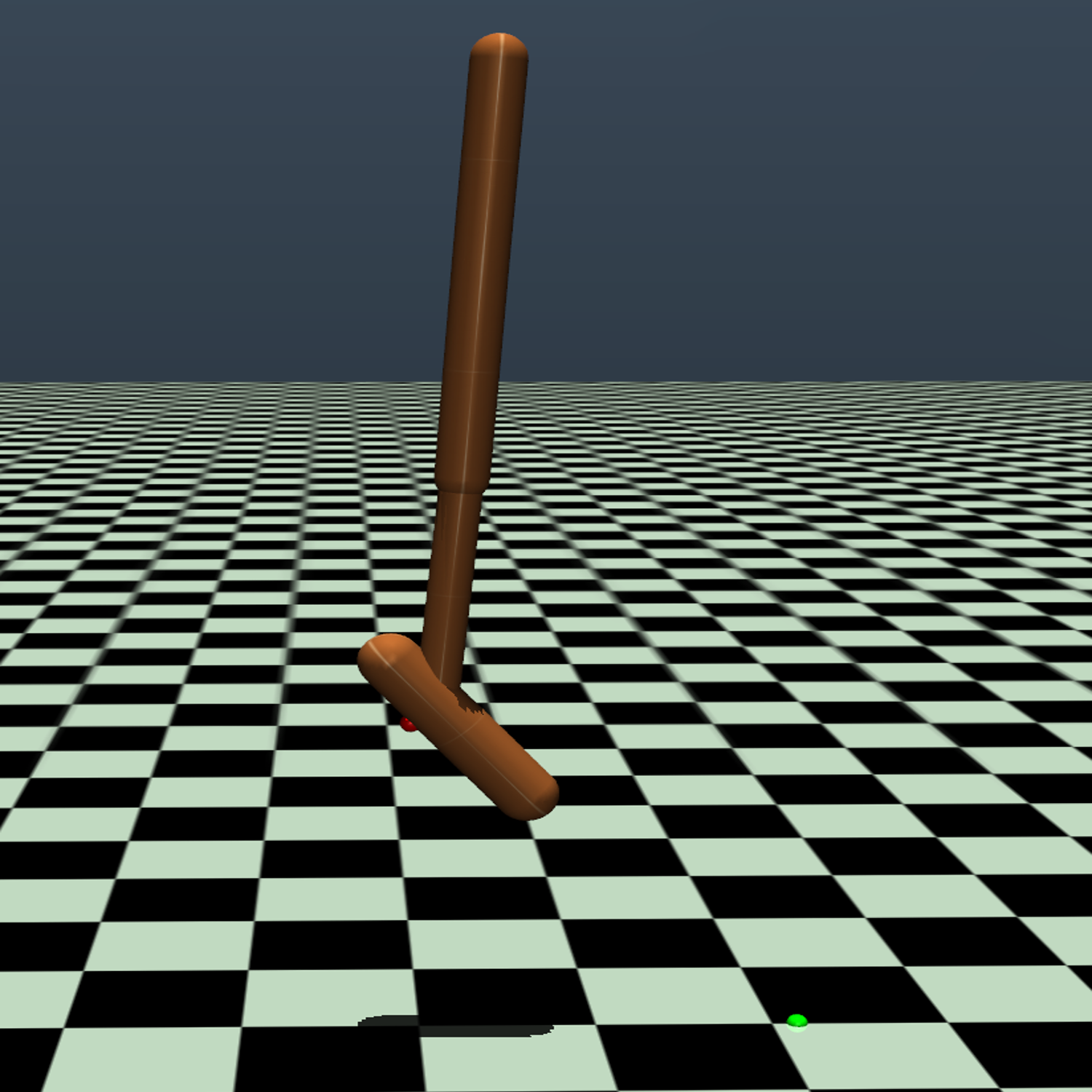}}%
    \end{subfigure}
    \hspace*{\fill}%
    \vspace{-0.5cm}
    \label{fig:hopper}
  \vspace{-0.3cm}
\end{wrapfigure}

As an addition to the main paper, we provide more details on the Hopper Jump task. We look at both the main goal of maximizing jump height and the secondary goal of landing on a desired position. These are shown along with the overall episode reward in \fig\ref{fig:hopper_jump}.
Our method shows quick learning and does well in achieving high jump height, consistent with what we reported earlier. While it's not as strong in landing accuracy, it still ranks high in overall performance.
Both versions of BBRL have similar results. However, they train more slowly compared to \tcp, highlighting the speed advantage of our method due to the use of intermediate states for policy updates.
Looking at other methods, step-based ones like PPO and TRPL focus too much on landing distance and miss out on jump height, leading to less effective policies. On the other hand, gSDE performs well but is sensitive to the initial setup, as shown by the wide confidence ranges in the results.
Lastly, SAC and PINK shows inconsistent results in jump height, indicating the limitations of using pink noise for exploration, especially when compared to gSDE.

\subsection{Box Pushing}
\label{appsub:bp}

\begin{wrapfigure}{r}{0.20\textwidth}
    \vspace{-0.5cm}
    \hspace*{\fill}%
    \begin{subfigure}{0.20\textwidth}
        \resizebox{\textwidth}{!}{\includegraphics{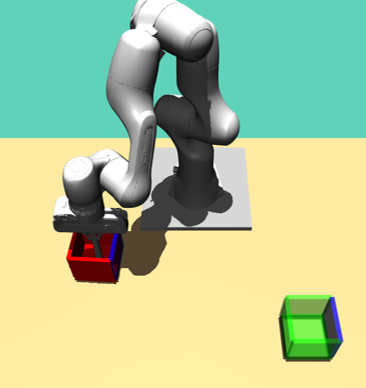}}%
    \end{subfigure}
    \hspace*{\fill}%
    \vspace{-0.5cm}
    \label{fig:box_pushing_env}
  \vspace{-0.3cm}
\end{wrapfigure}
The goal of the box-pushing task is to move a box to a specified goal location and orientation using the 7-DoFs Franka Emika Panda \citep{otto2023deep}. To make the environment more challenging, we extend the environment from a fixed initial box position and orientation to a randomized initial position and orientation. 
The range of both initial and target box pose varies from $x \in [0.3, 0.6], y \in [-0.45, 0.45], \theta_z \in [0, 2\pi]$.
Success is defined as a positional distance error of less than 5 cm and a z-axis orientation error of less than 0.5 rad. 
We refer to the original paper for the observation and action spaces definition and the reward function.

\subsection{Table Tennis}

\begin{figure}[h!]
    \centering
    \hspace*{\fill}%
    \resizebox{0.9\textwidth}{!}{
        \begin{tikzpicture} 
\def\linewidthtcp{1mm}
\def\linewidthothers{0.5mm}

    \begin{axis}[%
    hide axis,
    xmin=10,
    xmax=50,
    ymin=0,
    ymax=0.4,
    legend style={
        draw=white!15!black,
        legend cell align=left,
        legend columns=-1, 
        legend style={
            draw=none,
            column sep=1ex,
            line width=1pt
        }
    },
    ]
    \addlegendimage{C0, line width=\linewidthtcp}
    \addlegendentry{\acrshort{tcp} (ours)};
    \addlegendimage{C1, line width=\linewidthothers}
    \addlegendentry{\acrshort{bbrl} Cov.};
    \addlegendimage{C9, line width=\linewidthothers}
    \addlegendentry{\acrshort{bbrl}};
    \addlegendimage{C2, line width=\linewidthothers}
    \addlegendentry{\acrshort{ppo}};
    \addlegendimage{C3, line width=\linewidthothers}
    \addlegendentry{\acrshort{trpl}};
    \addlegendimage{C4, line width=\linewidthothers}
    \addlegendentry{\acrshort{sac}};    
    \addlegendimage{C5, line width=\linewidthothers}
    \addlegendentry{\acrshort{gsde}};
    \addlegendimage{C6, line width=\linewidthothers}
    \addlegendentry{\acrshort{pink}};

    \end{axis}
\end{tikzpicture}
    }%
    \hspace*{\fill}%
    \newline    
    \hspace*{\fill}%
    \begin{subfigure}{0.324\textwidth}
        \resizebox{\textwidth}{!}{\input{figure/table_tennis/plot/tt_iqm_success}}%
        \caption{Success Rate}
        \label{subfig:tt_success2}
    \end{subfigure}
    \hfill    
    \begin{subfigure}{0.324\textwidth}
        \resizebox{\textwidth}{!}{\input{figure/table_tennis/plot/tt_iqm_hit}}%
        \caption{Hit Rate}
        \label{subfig:tt_hit}
    \end{subfigure}
    \hfill    
    \begin{subfigure}{0.315\textwidth}
        \resizebox{\textwidth}{!}{\input{figure/table_tennis/plot/tt_iqm_reward}}%
        \caption{Episode Reward}
        \label{subfig:tt_ep_reward}
    \end{subfigure}
    \hspace*{\fill}%
    
    \caption{Robot Table Tennis Rand2Rand}
    \label{fig:tt}
\end{figure}

\begin{wrapfigure}{r}{0.20\textwidth}
    \vspace{-0.5cm}
    \hspace*{\fill}%
    \begin{subfigure}{0.20\textwidth}
        \resizebox{\textwidth}{!}{\includegraphics{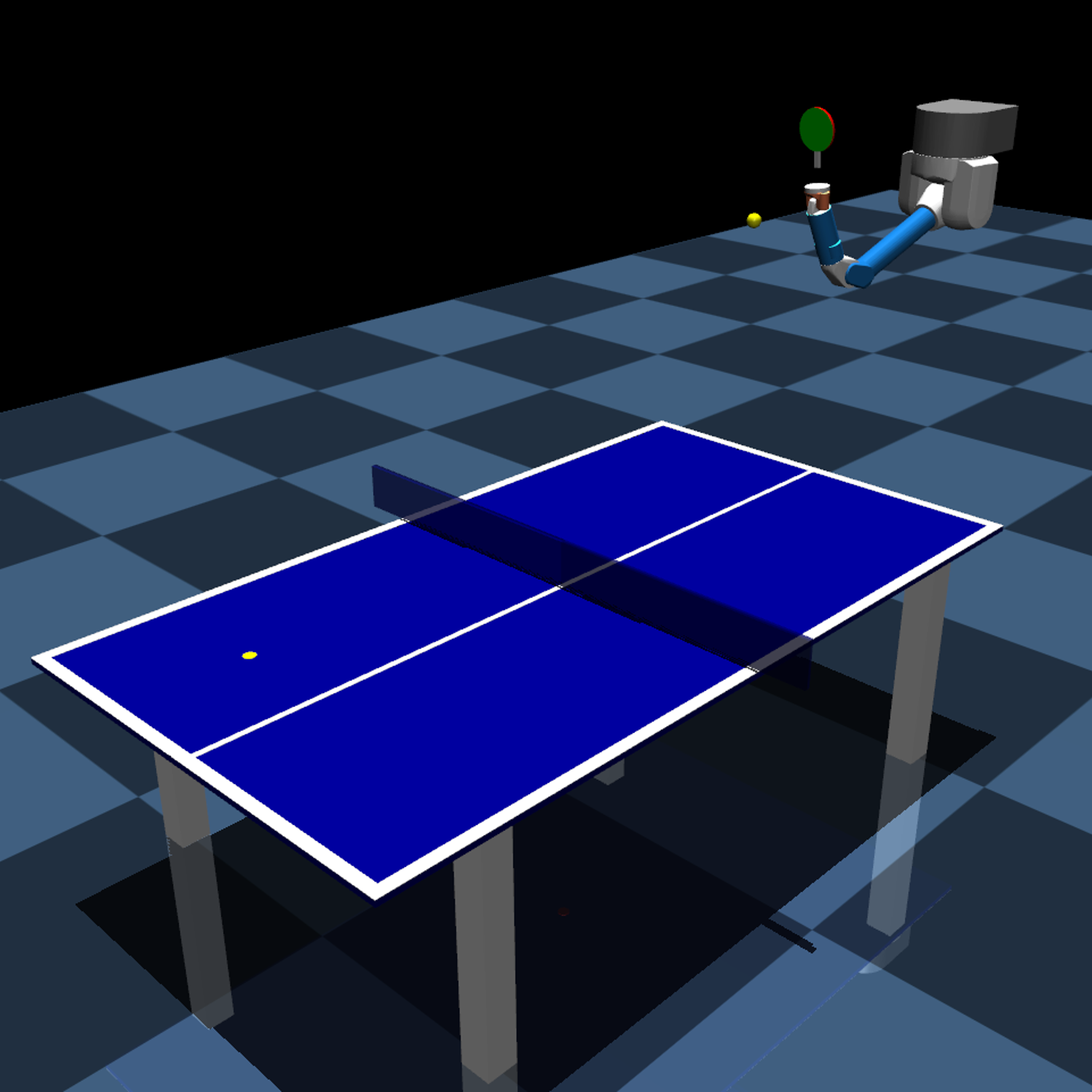}}%
    \end{subfigure}
    \hspace*{\fill}%
    \vspace{-0.5cm}
    \label{fig:table_tennis_env}
  \vspace{-0.3cm}
\end{wrapfigure}
 
The goal of table tennis environment to use the 7-DoFs robotic arm to hit the incoming ball and return it as close as possible to the specified goal location. We adapt the table tennis environment from \citet{celik2022specializing, otto2023deep} and extend it to a randomized initial robot joint configuration.
As context space we consider the initial ball position $x \in [-1, -0.2]$, $y \in [-0.65, 0.65]$ and the goal position $x \in [-1.2, -0.2]$, $y \in [-0.6, 0.6]$. 
The task is considered successful if the returned ball lands on the opponent's side of the table and within $\leq 0.2$m to the goal location. 
We refer to the original paper for the observation and action spaces definition and the reward function.

\newpage

\section{Additional Evaluation and Ablation Study}
\label{sec:add_exp}

\subsection{Trajectory Smoothness Metric}
We compared the trajectory smoothness of all methods in Table \ref{tab:smoothness}. To ensure a fair comparison, all methods were trained using the fixed start box pushing dense reward setting as originally reported in \cite{otto2023deep}, where each method achieved a minimum 50\% success rate. Trajectories for evaluation were generated using the mean prediction of the policy. The smoothness was assessed using three metrics: \emph{maximum jerk}, \emph{mean squared jerk} \citep{wininger2009spatial}, and \emph{dimensionless jerk} \citep{hogan2009sensitivity}. The first two metrics are standard in robot trajectory generation \citep{berscheid2021jerk, lange2015trajectory}, while the last is proposed as a more equitable measure of smoothness, eliminating the effects of motion magnitude and time scaling.
TCE and BBRL Cov outperformed all other methods in smoothness, followed by the original BBRL. This performance disparity likely stems from the original BBRL's inability to model inter-\dof movement correlations. In contrast, all step-based methods exhibited lower smoothness, attributable to their inherent per-step action selection approach.
\renewcommand{\arraystretch}{1.5}
\setlength{\aboverulesep}{0pt}
\setlength{\belowrulesep}{0pt}
\captionsetup{font=small}
\begin{figure}[h]
    \begin{minipage}[b]{\linewidth}
        \scriptsize
        \centering
        \vspace{1.2mm}
        \captionof{table}{Trajectory Smoothness, Mean (Std) of Three Metrics over 400 Trajectories. \label{tab:smoothness}}
        \begin{tabular}{ccccccccc}
            \toprule
            \textbf{Metric} & \textbf{TCE} &\textbf{BBRL Cov} & \textbf{BBRL} & \textbf{PPO} & \textbf{TRPL} & \textbf{gSDE} & \textbf{SAC} & \textbf{PINK}\\            
            \hline
            \textbf{Maximum Jerk}, ${\times10^3 \text{rad}/\text{s}^3}$  & \textbf{3.4 (1.9)} & 3.5 (1.5) & 4.3 (1.4) & 9.1 (3.3) & 12.9 (4.8) & 6.9 (2.2) & 9.3 (4.2) & 6.5 (1.7)\\            
            \textbf{Mean Sq. Jerk}, ${\times10^6 \text{rad}^2/\text{s}^6}$ & \textbf{0.2 (0.2)} & \textbf{0.2 (0.3)} & 0.6 (0.6) & 1.3 (0.9) & 5.5 (8.6) & 0.8 (0.6) & 3.9 (1.1) & 1.7 (0.7)\\            
            \textbf{Dimensionless Jerk}, ${\times10^6}$ & \textbf{61 (73)} & 64 (56) & 128 (49) & 141 (67) & 555 (472) & 122 (83) & 506 (262) & 311 (127)\\            
            \bottomrule
        \end{tabular}
    \end{minipage}
 \vspace{0.1cm}
\end{figure}

\newpage
\subsection{Action Correlations Predicted by Trained Policies}
\label{reb:action_correlation}
\begin{figure}[h]
    \centering
    \begin{subfigure}[b]{0.30\textwidth}
        \includegraphics[width=\textwidth]{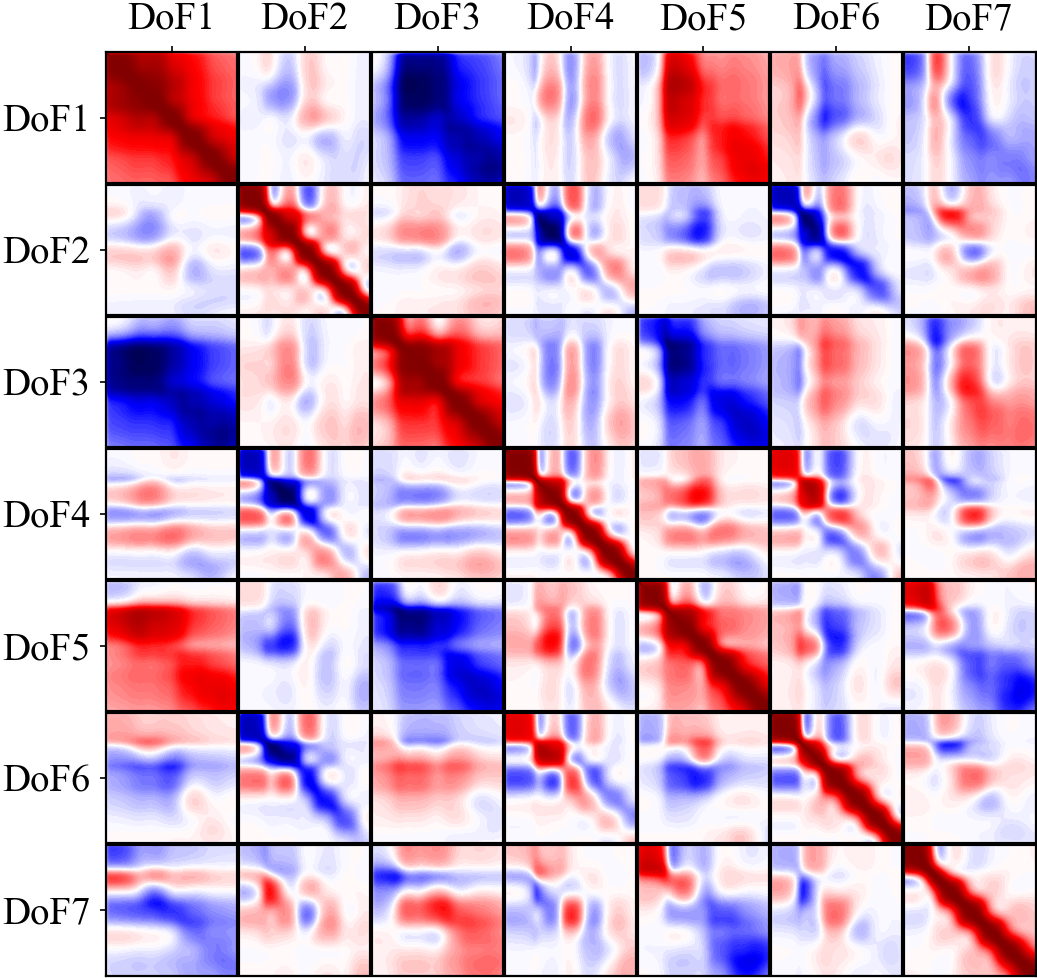}
        \subcaption{\tcp}
        \vspace{0.3cm}
    \end{subfigure}
    \hfill    
    \begin{subfigure}[b]{0.30\textwidth}
        \includegraphics[width=\textwidth]{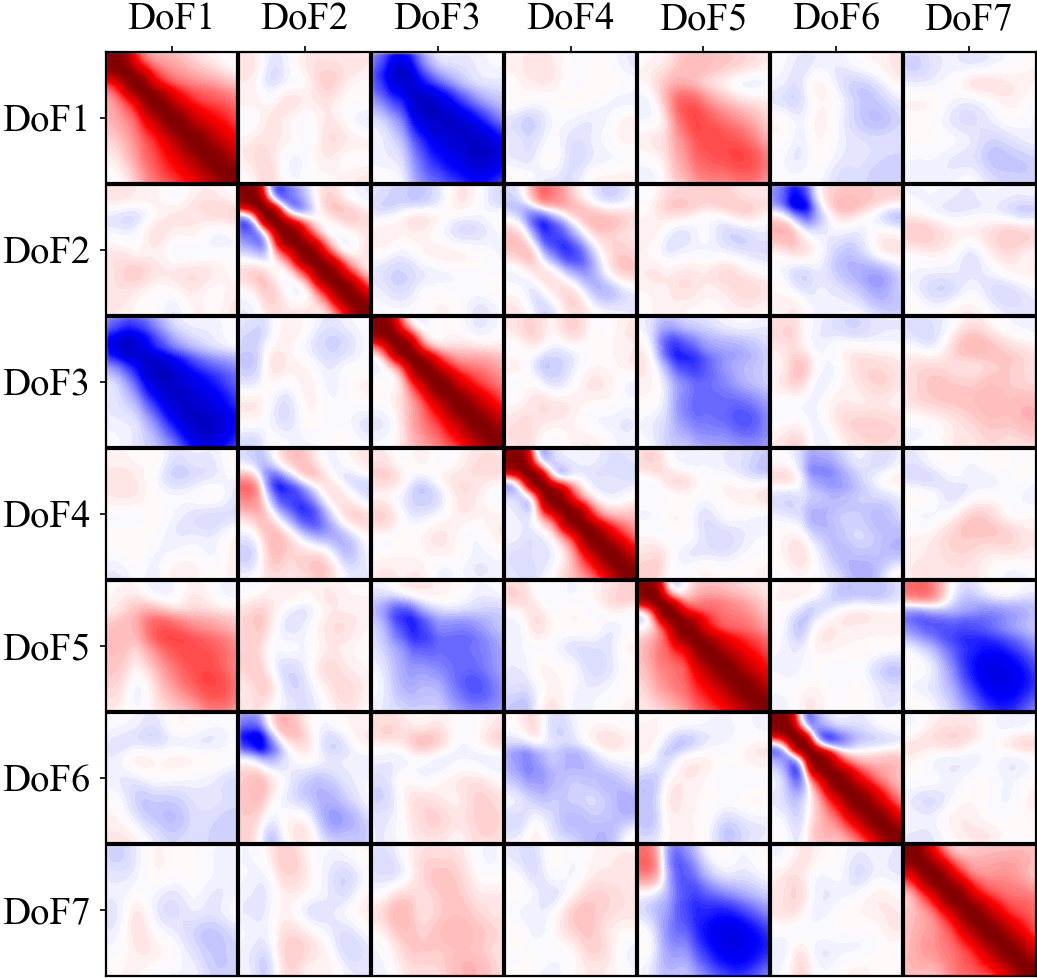}
        \subcaption{BBRL Cov}
        \vspace{0.3cm}
    \end{subfigure}
    \hfill    
    \begin{subfigure}[b]{0.30\textwidth}
        \includegraphics[width=\textwidth]{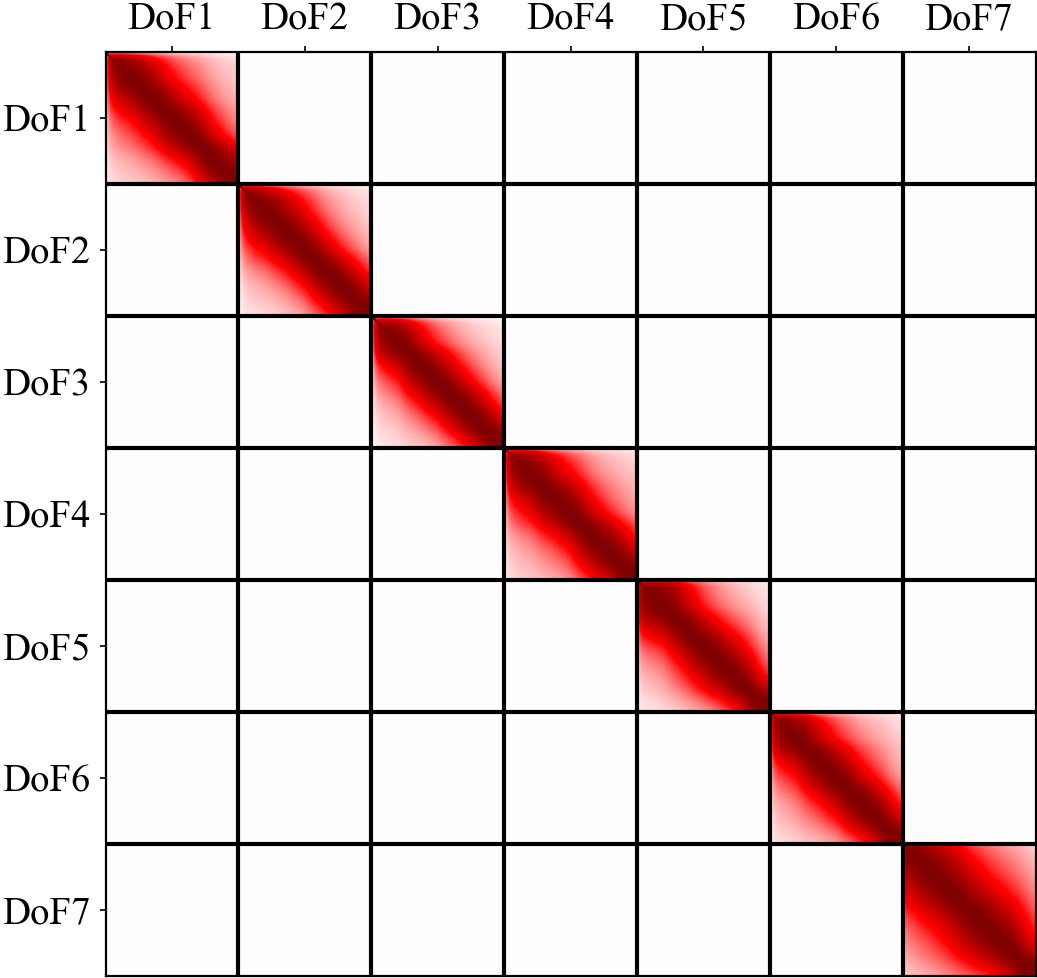}        
        \subcaption{BBRL (Original, Std)}
        \vspace{0.3cm}
    \end{subfigure}
    \begin{subfigure}[b]{0.30\textwidth}
        \includegraphics[width=\textwidth]{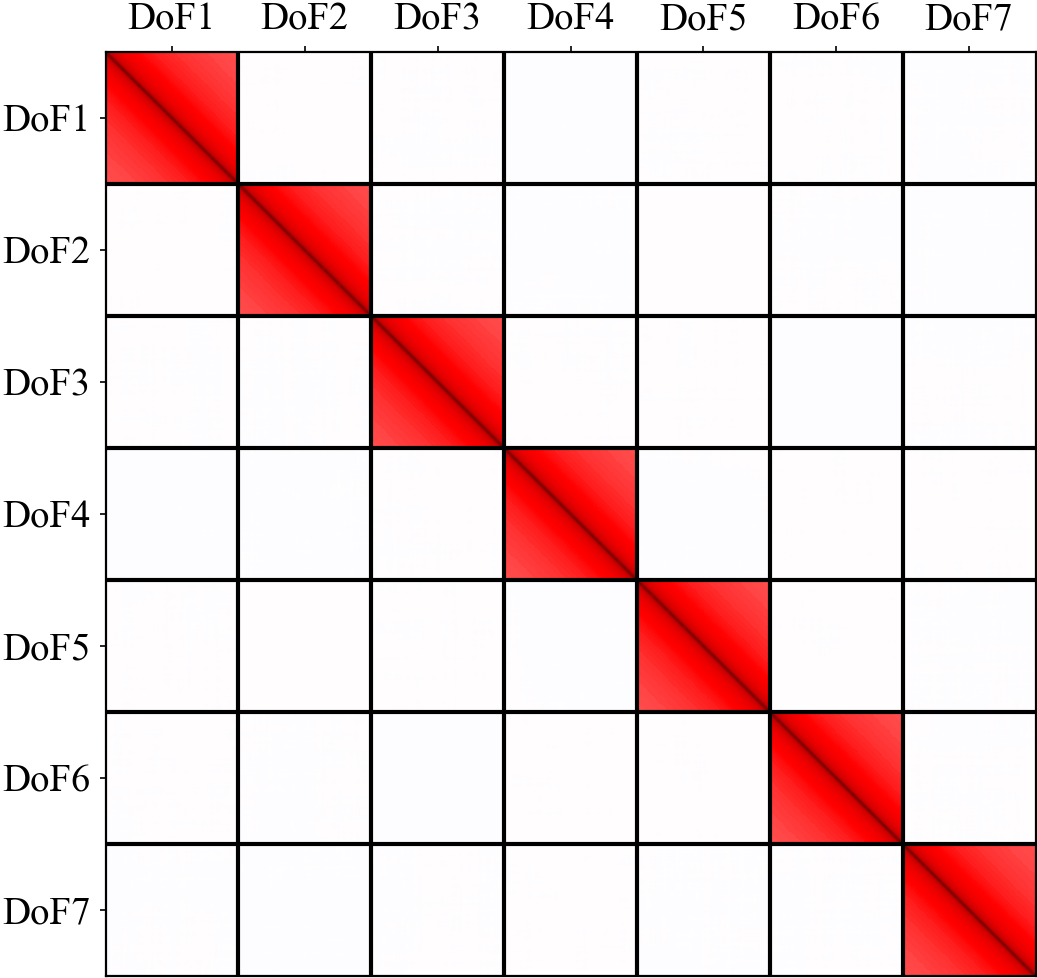}        
        \subcaption{PINK}
        \vspace{0.3cm}
    \end{subfigure}
    \hfill
    \begin{subfigure}[b]{0.30\textwidth}
        \includegraphics[width=\textwidth]{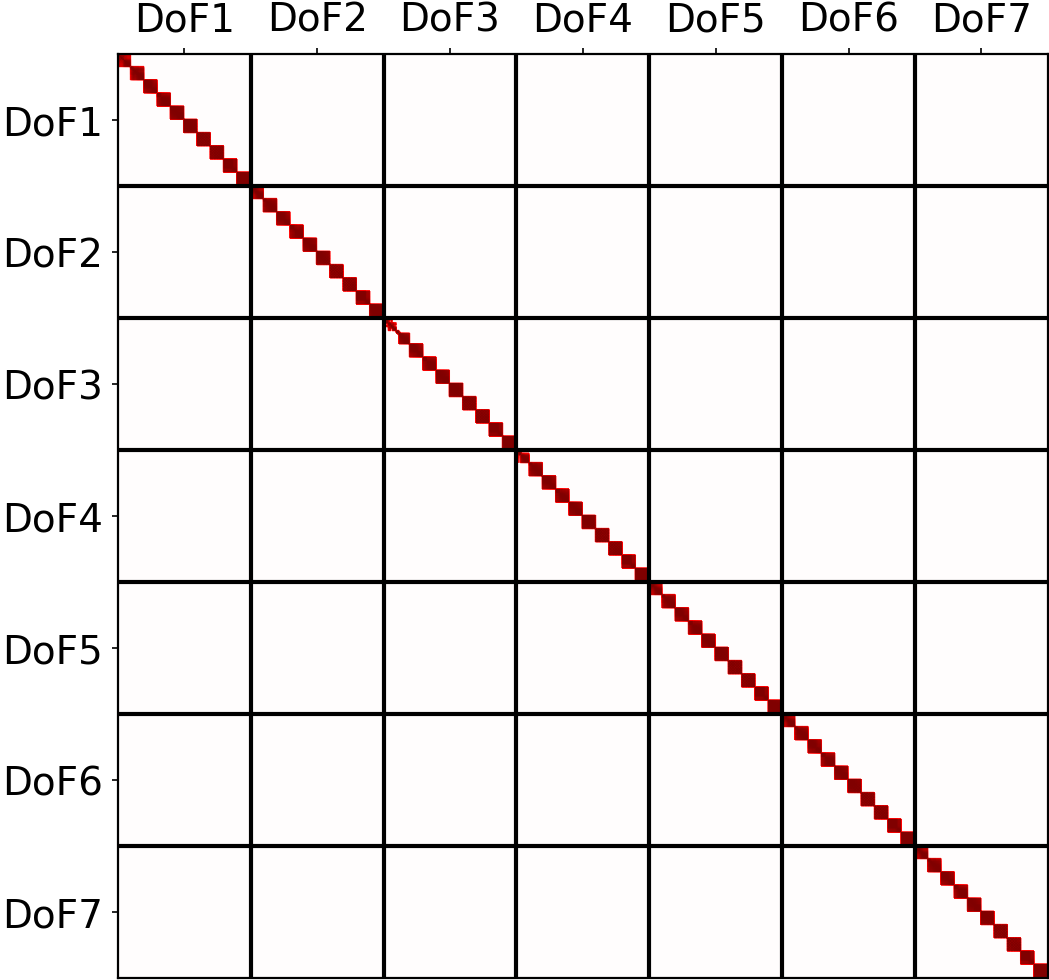}        
        \subcaption{gSDE}
        \vspace{0.3cm}
    \end{subfigure}
    \hfill
    \begin{subfigure}[b]{0.30\textwidth}
    \includegraphics[width=\textwidth]{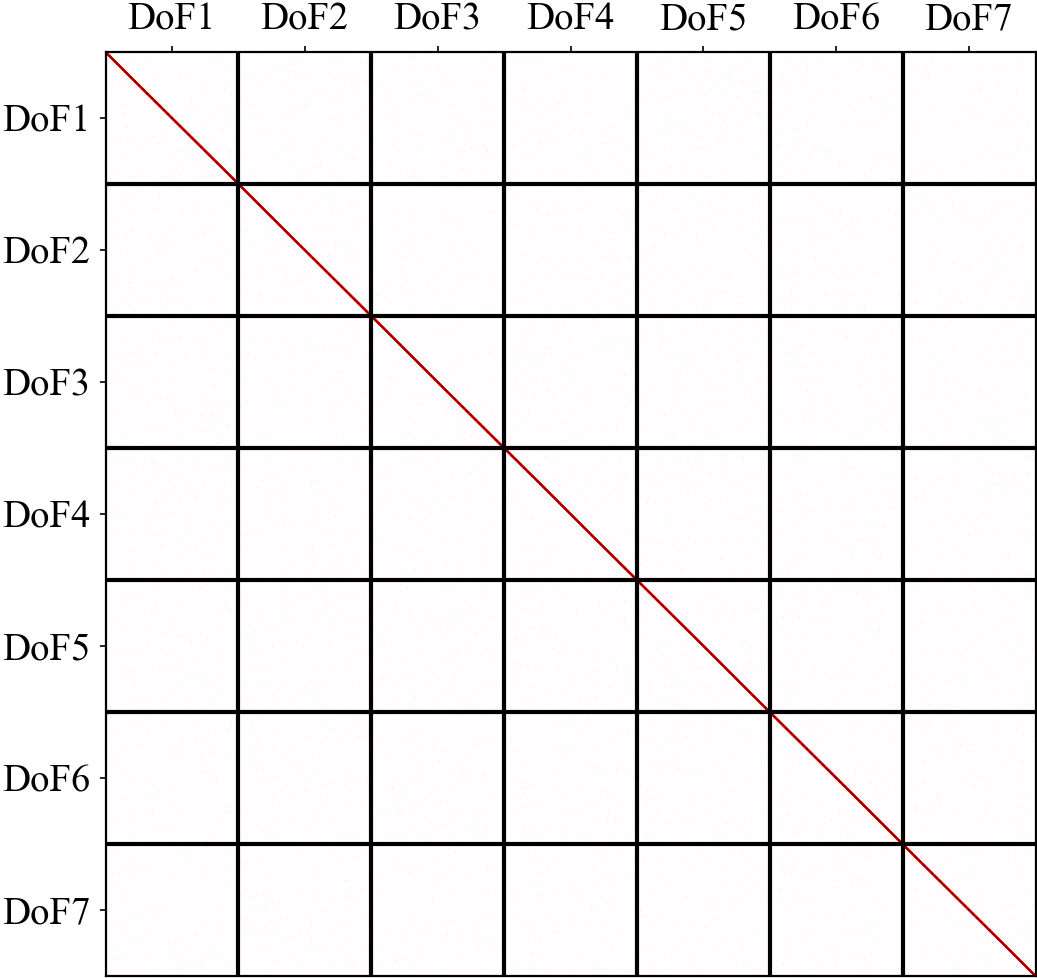}        
    \subcaption{PPO / TRPL / SAC}
    \vspace{0.3cm}
    \end{subfigure}   

    \caption{This figure presents predicted actions' correlation across 7 \dof and 100 time steps, visualized in a 700x700 correlation matrix. Each 100 $\times$ 100 square tile demonstrates the movement correlation between two \dof during these steps. Correlation values range from -1 (negative correlation, depicted in blue) to 1 (positive correlation, depicted in red), with white areas indicating no correlation. The action outputs for \tcp, BBRL, and BBRL Cov are the positions of the robot joints, whereas step-based methods predict actions in the torque space. TCE and BBRL Cov exhibit to a higher capacity of movement correlations. The original BBRL and PINK only model correlations within each \dof. gSDE models correlations over a few consecutive time steps. We show only one representative matrix for PPO, TRPL, and SAC, as their results are visually identical, typically resulting in matrices resembling the identity matrix.}
    \label{fig:action_correlation}
\end{figure}

We plot the action correlation coupling \dof and time steps in \fig \ref{fig:action_correlation}. 
All policies were trained under the box-pushing task with a dense reward setting. The action outputs for \tcp, BBRL, and BBRL Cov are the positions of the robot joints, while step-based methods, such as PPO, predict actions in the torque space.
TCE and BBRL Cov demonstrate the ability to predict actions correlated both temporally and across \dof, as indicated by the non-zero off-diagonal elements in their correlation matrices. In contrast, the original BBRL translates a factorized weight distribution into a block-diagonal action correlation matrix, capturing variance within individual \dof but not between them. Similarly, PINK is constrained to modeling intra-\dof correlations, which depend solely on the time difference. This limitation arises from the wide-sense stationarity of the noise, resulting in a constant value along each diagonal.
gSDE, however, models temporal correlation but only over a few consecutive time steps, observable along the diagonal elements.
Actions predicted by PPO, TRPL, and SAC lack both temporal and \dof correlation, resulting in correlation matrices resembling identity matrices.
Interestingly, for methods that only capture intra-\dof correlations, these correlations are uniformly positive. This trend may relate to the control cost in the reward function, promoting consistent movement within each \dof over time. On the other hand, TCE and BBRL Cov are unique in their ability to capture negative correlations, both between and within \dof, enhancing their flexibility in trajectory sampling.
\newline
\newline
\afterpage{\newpage}
\afterpage{\newpage}
\newpage

\afterpage{\newpage}
\subsection{Ablation: SAC + Motion Primitives-based Method}
Training movement primitive-based methods using standard RL techniques, such as PPO and SAC, generally poses challenges due to the complex, higher-dimensional trajectory parameter space. In the study by \citet{otto2023deep}, an ablation study employing a PPO-style trust region (likelihood clipping) for training BBRL demonstrated inferior performance compared to the use of a differentiable trust region projection layer.

\begin{figure}[h]
    \centering
    \hspace*{\fill}%
    \resizebox{0.4\textwidth}{!}{
        \begin{tikzpicture} 
\def\linewidthtcp{1mm}
\def\linewidthothers{0.5mm}

    \begin{axis}[%
    hide axis,
    xmin=10,
    xmax=50,
    ymin=0,
    ymax=0.4,
    legend style={
        draw=white!15!black,
        legend cell align=left,
        legend columns=-1, 
        legend style={
            draw=none,
            column sep=1ex,
            line width=1pt
        }
    },
    ]    
    \addlegendimage{C1, line width=\linewidthothers}
    \addlegendentry{\acrshort{bbrl} Cov.};
    \addlegendimage{C9, line width=\linewidthothers}
    \addlegendentry{\acrshort{bbrl}};        
    \addlegendimage{C4, line width=\linewidthothers}
    \addlegendentry{\acrshort{sac} $+$ MP};

    \end{axis}
\end{tikzpicture}
    }%
    \hspace*{\fill}%
    \newline        
    \hspace*{\fill}%
    \resizebox{0.3\textwidth}{!}{\input{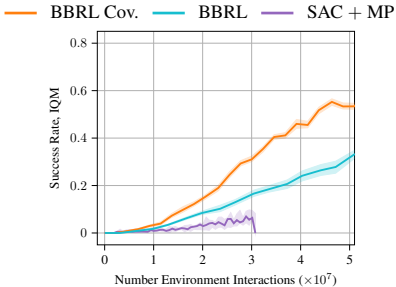}}%
    \hspace*{\fill}%
    \caption{By employing the standard SAC method to learn trajectory parameters $\vw$, we compared its performance with that of the original BBRL and BBRL Cov methods under the box pushing dense reward setting. The ablated method, using SAC, showed relatively poorer performance. }
    \label{fig:sac_mp}    
\end{figure}

In Figure \ref{fig:sac_mp}, we present an additional ablation study where SAC is used to learn the trajectory parameters of movement primitives. This study compares the performance of SAC with that of the original BBRL and BBRL Cov, leading to relatively poorer performance. 
The SAC model selected for reporting was the best performer among 40 different combinations of hyperparameters. The hyperparameters adjusted include the output action scaling factor (necessary because the SAC action space is bounded by $[-1, 1]$), policy/critic learning rate, batch size, and the size of the policy/critic network. 
The relatively shorter training curve of SAC can be attributed to its higher computational cost in policy update \citep{haarnoja2018sacaa}.

\afterpage{\newpage}
\subsection{Ablation: Using PPO Style Trust Regions for TCE Method}
\begin{figure}[h]
    \centering
    \hspace*{\fill}%
    \resizebox{0.4\textwidth}{!}{
        \begin{tikzpicture} 
\def\linewidthtcp{1mm}
\def\linewidthothers{0.5mm}

    \begin{axis}[%
    hide axis,
    xmin=10,
    xmax=50,
    ymin=0,
    ymax=0.4,
    legend style={
        draw=white!15!black,
        legend cell align=left,
        legend columns=-1, 
        legend style={
            draw=none,
            column sep=1ex,
            line width=1pt
        }
    },
    ]    
    \addlegendimage{C0, line width=\linewidthtcp}
    \addlegendentry{\acrshort{tcp} (ours)};    
    \addlegendimage{C3, line width=\linewidthothers}
    \addlegendentry{\acrshort{tcp} (PPO)};
    \addlegendimage{C2, line width=\linewidthothers}    
    \addlegendentry{\acrshort{ppo}};

    \end{axis}
\end{tikzpicture}
    }%
    \hspace*{\fill}%
    \newline        
    \hspace*{\fill}%
    \resizebox{0.3\textwidth}{!}{\input{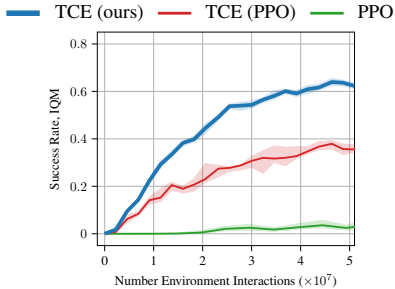}}%
    \hspace*{\fill}%
    \caption{Training TCE with a PPO-style trust region, employing likelihood clipping with $\epsilon=0.1$, yields suboptimal performance in the box pushing dense reward setting. Nevertheless, its superior performance compared to standard PPO underscores our method's effectiveness in episodic trajectory modeling.}
    \label{fig:tce_ppo}    
\end{figure}

We developed an ablated version of our method, incorporating the PPO-style trust region via likelihood clipping. We tuned the clipping ratio $\epsilon$ between 0.05 and 0.2. As illustrated in Figure \ref{fig:tce_ppo}, this version's performance falls between the original TCE and the standard PPO. The movement primitives' high-dimensional parameter space limits the effectiveness of likelihood clipping in precisely maintaining the trust region during policy updates. This limitation likely accounts for the performance gap between TCE and its PPO variant. Nonetheless, the ablated model still demonstrates a notable advantage over standard PPO, further substantiating our model's effectiveness in temporally correlated trajectory prediction.

\afterpage{\newpage}
\subsection{Ablation: Selection of the Amount of Segments K}
We conducted an ablation study to evaluate the effect of varying the number of segments (k) on model performance. The number of segments tested ranged from 2 to 100. Our experiments involved training in both dense and sparse box-pushing environments. The results revealed a greater sensitivity to the number of segments in the sparse reward environment compared to the dense environment. We attribute this to the challenges associated with the value function approximation under sparse reward settings. 
However, within an optimal range, such as 10-25 segments, this parameter is not overly sensitive compared to other hyper-parameters.
Consequently, we have adopted k=25 for all experiments in this paper.

\begin{figure}[h!]
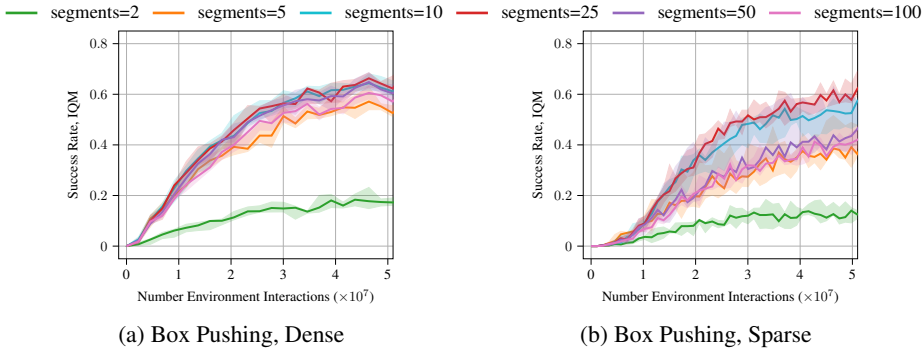

    \centering
    \hspace*{\fill}%
    \resizebox{0.9\textwidth}{!}{
        \begin{tikzpicture} 
\def\linewidthtcp{1mm}
\def\linewidthothers{0.5mm}

    \begin{axis}[%
    hide axis,
    xmin=10,
    xmax=50,
    ymin=0,
    ymax=0.4,
    legend style={
        draw=white!15!black,
        legend cell align=left,
        legend columns=-1, 
        legend style={
            draw=none,
            column sep=1ex,
            line width=1pt
        }
    },
    ]
    
    \addlegendimage{C2, line width=\linewidthothers}
    \addlegendentry{segments=2};
    \addlegendimage{C1, line width=\linewidthothers}
    \addlegendentry{segments=5};
    \addlegendimage{C9, line width=\linewidthothers}
    \addlegendentry{segments=10};    
    \addlegendimage{C3, line width=\linewidthothers}
    \addlegendentry{segments=25};
    \addlegendimage{C4, line width=\linewidthothers}
    \addlegendentry{segments=50};        
    \addlegendimage{C6, line width=\linewidthothers}
    \addlegendentry{segments=100};

    \end{axis}
\end{tikzpicture}
    }%
    \hspace*{\fill}%
    \newline    
    \hspace*{\fill}%
    \begin{subfigure}{0.32\textwidth}
        \resizebox{\textwidth}{!}{\input{figure/k_selection/plot/dense}}%
        \caption{Box Pushing, Dense}        
    \end{subfigure}
    \hfill    
    \begin{subfigure}{0.32\textwidth}
        \resizebox{\textwidth}{!}{\input{figure/k_selection/plot/sparse}}%
        \caption{Box Pushing, Sparse}        
    \end{subfigure}
    \hspace*{\fill}%
    \caption{Study of the number of segment per trajectory.}
    \label{fig:bp}
\end{figure}

\afterpage{\newpage}

\subsection{TCE Cov vs. STD} 
We conducted an ablation study to assess the impact of employing a full covariance policy in the \tcp framework. This involved comparing the standard \tcp with its variant, \tcp Std, which utilizes a factorized Gaussian policy  $\mathcal{N}(\vw|\bm{\mu}_{\vw},\bm{\sigma}^2_{\vw})$. The comparison was conducted in scenarios involving both dense and sparse reward settings in box pushing tasks. The findings revealed that the ablated version, \tcp Std, consistently underperformed compared to the full covariance version. This underperformance is attributed to the limited correlation capacity of the factorized Gaussian policy.

Furthermore, it is important to note that while the factorized Gaussian distribution results in a relatively lower computational load in the parameter space, it does not offer a marked advantage when translated into trajectory space. As illustrated in \figref{fig:action_correlation}(c) of Section \ref{reb:action_correlation}, a factorized parameter distribution ultimately converts into a blocked diagonal trajectory distribution. Although this format is visually simpler compared to a full trajectory covariance matrix, both share same time complexity in terms of likelihood computation. This computational process is significantly more resource-intensive than that for a purely diagonal matrix. Therefore, we utilize the techniques in \citet{li2023prodmp} to apply a likelihood estimation and reduce the computational cost.

\begin{figure}[h!]
    \centering
    \hspace*{\fill}%
    \resizebox{0.3\textwidth}{!}{
        \begin{tikzpicture} 
\def\linewidthtcp{1mm}
\def\linewidthothers{0.5mm}

    \begin{axis}[%
    hide axis,
    xmin=10,
    xmax=50,
    ymin=0,
    ymax=0.4,
    legend style={
        draw=white!15!black,
        legend cell align=left,
        legend columns=-1, 
        legend style={
            draw=none,
            column sep=1ex,
            line width=1pt
        }
    },
    ]
    
    \addlegendimage{C0, line width=\linewidthtcp}
    \addlegendentry{\acrshort{tcp} Cov (ours)};        
    \addlegendimage{C1, line width=\linewidthothers}
    \addlegendentry{\acrshort{tcp} Std (ablation)};

    \end{axis}
\end{tikzpicture}
    }%
    \hspace*{\fill}%
    \newline    
    \hspace*{\fill}%
    \begin{subfigure}{0.32\textwidth}
        \resizebox{\textwidth}{!}{\begin{tikzpicture}
\def\tcp_color{C0}
\def\bbrlcov_color{C1}
\def\bbrlcolor{C9}
\def\ppo_color{C2}
\def\trpl_color{C3}
\def\sac_color{C4}
\def\gsde_color{C5}
\def\pink_color{C6}
\def\linewidthtcp{1mm}
\def\linewidthothers{0.5mm}

\begin{axis}[
legend cell align={left},
legend style={fill opacity=0.8, draw opacity=1, text opacity=1, draw=lightgray204, at={(0.03,0.03)},  anchor=north west},
tick align=outside,
tick pos=left,
x grid style={darkgray176},
scaled x ticks=false,
xticklabels={,0,1,2,3,4,5},
xlabel={Number Environment Interactions ($\times 10^7$)},
xmajorgrids,
xmin=-1500000, xmax=51000000.05,
xtick style={color=black},
y grid style={darkgray176},
ylabel={Success Rate, IQM},
ymajorgrids,
ymin=-0.05, ymax=0.85,
ytick style={color=black}
]

\path [draw=\tcp_color, fill=\tcp_color, opacity=0.2]
(axis cs:15200,0)
--(axis cs:15200,0)
--(axis cs:2295200,0.0152549342105263)
--(axis cs:4575200,0.0892024150647615)
--(axis cs:6855200,0.133799660707775)
--(axis cs:9135200,0.209871680228726)
--(axis cs:11415200,0.276310310737708)
--(axis cs:13695200,0.325649439861548)
--(axis cs:15975200,0.370217706188354)
--(axis cs:18255200,0.388170632635958)
--(axis cs:20535200,0.422206593301904)
--(axis cs:23195200,0.478561903974283)
--(axis cs:25475200,0.525660208275975)
--(axis cs:27755200,0.520243235666162)
--(axis cs:30035200,0.529554776404656)
--(axis cs:32315200,0.552898960104421)
--(axis cs:34595200,0.56663194622836)
--(axis cs:36875200,0.591489464423172)
--(axis cs:39155200,0.578044160313277)
--(axis cs:41435200,0.591081124780351)
--(axis cs:43715200,0.605291674440277)
--(axis cs:46375200,0.625889744825062)
--(axis cs:48655200,0.622372769709884)
--(axis cs:50935200,0.612538137575972)
--(axis cs:50935200,0.636454736072786)
--(axis cs:48655200,0.646594249208239)
--(axis cs:46375200,0.654701387123502)
--(axis cs:43715200,0.630007616363199)
--(axis cs:41435200,0.62476808111555)
--(axis cs:39155200,0.604036007121634)
--(axis cs:36875200,0.609550431143027)
--(axis cs:34595200,0.603329090924197)
--(axis cs:32315200,0.573865324909552)
--(axis cs:30035200,0.560176153835983)
--(axis cs:27755200,0.553472112338833)
--(axis cs:25475200,0.546001738111999)
--(axis cs:23195200,0.498409132712363)
--(axis cs:20535200,0.463305018148498)
--(axis cs:18255200,0.411178069337232)
--(axis cs:15975200,0.391737385677915)
--(axis cs:13695200,0.348929763932351)
--(axis cs:11415200,0.306973885349457)
--(axis cs:9135200,0.234576984428261)
--(axis cs:6855200,0.15014038286711)
--(axis cs:4575200,0.100532338995683)
--(axis cs:2295200,0.0201685855263158)
--(axis cs:15200,0)
--cycle;

\addplot [line width=\linewidthtcp, \tcp_color, mark=*, mark size=0, mark options={solid}]
table {%
15200 0
2295200 0.0181537828947368
4575200 0.0945020173725329
6855200 0.142939306560316
9135200 0.224099065520261
11415200 0.292446072362854
13695200 0.334573394364883
15975200 0.382682600751653
18255200 0.398186277580882
20535200 0.444624829227277
23195200 0.491558919474906
25475200 0.537460544107651
27755200 0.53991287239233
30035200 0.543181224135956
32315200 0.564715282434002
34595200 0.581068747804992
36875200 0.601034224441818
39155200 0.591088924099346
41435200 0.609252119402932
43715200 0.616412889203135
46375200 0.639450677405937
48655200 0.636568469537933
50935200 0.623114025766515
};

\path [draw=\bbrlcov_color, fill=\bbrlcov_color, opacity=0.2]
(axis cs:15200,0)
--(axis cs:15200,0)
--(axis cs:2295200,0.00587942023026316)
--(axis cs:4575200,0.0385262419048108)
--(axis cs:6855200,0.0615858248660439)
--(axis cs:9135200,0.0949680643411059)
--(axis cs:11415200,0.117780338118135)
--(axis cs:13695200,0.150787787008535)
--(axis cs:15975200,0.173704685653811)
--(axis cs:18255200,0.17374926482637)
--(axis cs:20535200,0.191777961035926)
--(axis cs:23195200,0.205095873962579)
--(axis cs:25475200,0.227100636313589)
--(axis cs:27755200,0.239104378416513)
--(axis cs:30035200,0.215205555113261)
--(axis cs:32315200,0.223144277447664)
--(axis cs:34595200,0.246853545141867)
--(axis cs:36875200,0.228918026200465)
--(axis cs:39155200,0.253441928107425)
--(axis cs:41435200,0.266982804208244)
--(axis cs:43715200,0.27048696949218)
--(axis cs:46375200,0.278944579293138)
--(axis cs:48655200,0.257920958792823)
--(axis cs:50935200,0.266686007918883)
--(axis cs:53215200,0.285933070396419)
--(axis cs:55495200,0.310869307210636)
--(axis cs:55495200,0.326351882324211)
--(axis cs:53215200,0.312565643706664)
--(axis cs:50935200,0.287878522361815)
--(axis cs:48655200,0.282975350078537)
--(axis cs:46375200,0.296605587915981)
--(axis cs:43715200,0.295136657319669)
--(axis cs:41435200,0.283717312492045)
--(axis cs:39155200,0.27153065643926)
--(axis cs:36875200,0.248969034206463)
--(axis cs:34595200,0.262080747967997)
--(axis cs:32315200,0.245631849322906)
--(axis cs:30035200,0.231624574524586)
--(axis cs:27755200,0.252181323938252)
--(axis cs:25475200,0.245085085577683)
--(axis cs:23195200,0.228822454135876)
--(axis cs:20535200,0.211593224966486)
--(axis cs:18255200,0.185455132001525)
--(axis cs:15975200,0.184198999570197)
--(axis cs:13695200,0.164071347199182)
--(axis cs:11415200,0.132982090303834)
--(axis cs:9135200,0.109357884918389)
--(axis cs:6855200,0.0723770804154246)
--(axis cs:4575200,0.0474373425935444)
--(axis cs:2295200,0.0116981907894737)
--(axis cs:15200,0)
--cycle;

\addplot [line width=\linewidthothers, \bbrlcov_color, mark=*, mark size=0, mark options={solid}]
table {%
15200 0
2295200 0.00900493421052631
4575200 0.0428685238486842
6855200 0.0659190378691021
9135200 0.102285351329728
11415200 0.126173630505065
13695200 0.157017016007969
15975200 0.180160300258498
18255200 0.179393798633622
20535200 0.19993164403526
23195200 0.21835085121931
25475200 0.23346662979647
27755200 0.244583797703572
30035200 0.222589174360937
32315200 0.234822681873251
34595200 0.253836956327267
36875200 0.238590364623237
39155200 0.263072369428633
41435200 0.274465291632627
43715200 0.281652079993328
46375200 0.288492902999207
48655200 0.270519125976232
50935200 0.277528002375717
53215200 0.297944302932
55495200 0.319234267394405
};

\end{axis}
\end{tikzpicture}}%
        \caption{Box Pushing, Dense}        
    \end{subfigure}
    \hfill    
    \begin{subfigure}{0.32\textwidth}
        \resizebox{\textwidth}{!}{\input{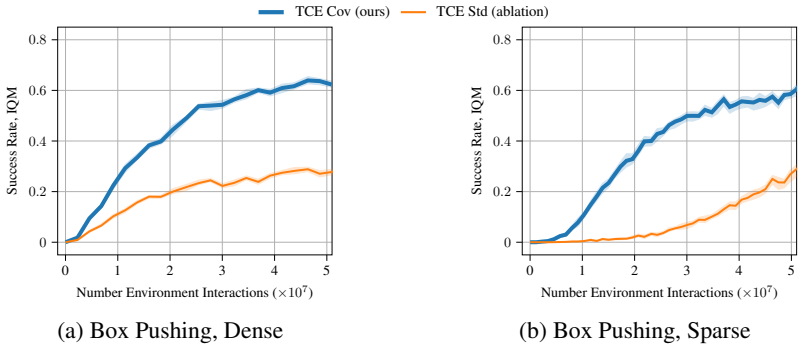}}%
        \caption{Box Pushing, Sparse}        
    \end{subfigure}
    \hspace*{\fill}%
    \caption{Study of the TCE Cov vs. TCE Std.}
    \label{fig:bp}
\end{figure}

\newpage
\section{Hyper Parameters}
\label{sec:hps}
We executed a large-scale grid search to fine-tune key hyperparameters for each baseline method. For other hyperparameters, we relied on the values specified in their respective original papers. Below is a list summarizing the parameters we swept through during this process.

\paragraph{BBRL:} Policy net size, critic net size, policy learning rate, critic learning rate, samples per iteration, trust region dissimilarity bounds, number of parameters per movement \dof.

\paragraph{TCE:} Same types of hyper-parameters listed in BBRL, plus the number of segments per trajectory.

\paragraph{PPO:} Policy network size, critic network size, policy learning rate, critic learning rate, batch size, samples per iteration.

\paragraph{TRPL:} Policy network size, critic network size, policy learning rate, critic learning rate, batch size, samples per iteration, trust region dissimilarity bounds. 

\paragraph{gSDE:} Same types of hyper-parameters listed in PPO, together with the state dependent exploration sampling frequency \citep{raffin2022smooth}.

\paragraph{SAC:} Policy network size, critic network size, policy learning rate, critic learning rate, alpha learning rate, batch size, Update-To-Data (UTD) ratio.

\paragraph{PINK:} Same types of hyper-parameters listed in SAC.

The detailed hyper parameters used are listed in the following tables. Unless stated otherwise, the notation lin\_x refers to a linear schedule. It interpolates linearly from x to 0 during training.
\newpage

\begin{table}[ht]
\centering
\centering
\caption{Hyperparameters for the Meta-World experiments.}
\label{tab:metaworld-HP}
\begin{adjustbox}{max width=\textwidth}
\scriptsize
\begin{tabular}{lccccccc}
\toprule
                                 & PPO        & gSDE        & TRPL       & SAC         & PINK              & \tcp   & BBRL      \\ 
\hline
\multicolumn{8}{l}{}                                                                                                                  \\
number samples                   & 16000        & 16000        & 16000        & 1000         & 4          & 16           & 16           \\
GAE $\lambda$                    & 0.95         & 0.95         & 0.95         & n.a.         & n.a.         &  0.95           & n.a.         \\
discount factor                  & 0.99         & 0.99         & 0.99         & 0.99         & 0.99         & 1            & n.a.        \\
\multicolumn{8}{l}{}                                                                                                                                 \\
\hline
\multicolumn{8}{l}{}                                                                                                                                 \\
$\epsilon_\mu$                   & n.a.         & n.a.         & 0.005        & n.a.         & n.a.         & 0.005      & 0.005        \\
$\epsilon_\Sigma$                & n.a.         & n.a.         & 0.0005       & n.a.         & n.a.         & 0.0005       & 0.0005         \\
\multicolumn{8}{l}{}                                                                                                                                 \\
\hline
\multicolumn{8}{l}{}                                                                                                                                 \\
optimizer                        & adam         & adam         & adam         & adam         & adam         & adam         & adam           \\
epochs                           & 10           & 10           & 20           & 1000         & 1            & 50           & 100            \\
learning rate                    & 3e-4         & 1e-3         & 5e-5         & 3e-4         & 3e-4         & 3e-4         & 3e-4          \\
use critic                       & True         & True         & True         & True         & True         & True         & True       \\
epochs critic                    & 10           & 10           & 10           & 1000         & 1            & 50           & 100           \\
learning rate critic (and alpha) & 3e-4         & 1e-3         & 3e-4         & 3e-4         & 3e-4         & 3e-4         & 3e-4          \\
number minibatches               & 32           & n.a.         & 64           & n.a.         & n.a.         & 1           & 1              \\
batch size                       & n.a.         & 500          & n.a.         & 256          & 512          & n.a.         & n.a.        \\
buffer size                      & n.a.         & n.a.         & n.a.         & 1e6          & 2e6          & n.a.         & n.a.           \\
learning starts                  & 0            & 0            & 0            & 10000        & 1e5          & 0            & 0                \\
polyak\_weight                   & n.a.         & n.a.         & n.a.         & 5e-3         & 5e-3         & n.a.         & n.a.        \\
trust region loss weight         & n.a.         & n.a.         & 10           & n.a.         & n.a.         & 1           & 25           \\
SDE sampling frequency           & n.a.         & 4            & n.a.         & n.a.         & n.a.     & n.a.    & n.a.               \\
entropy coefficient              & 0           & 0            & 0           & auto           & auto       & 0      & 0                \\
\multicolumn{8}{l}{}                                                                                                                                 \\
\hline
\multicolumn{8}{l}{}                                                                                                                                 \\
normalized observations          & True         & True         & True         & False        & False        & True         & False         \\
normalized rewards               & True         & True         & False        & False        & False        & False        & False        \\
observation clip                 & 10.0         & n.a.         & n.a.         & n.a.         & n.a.         & n.a.         & n.a.           \\
reward clip                      & 10.0         & 10.0         & n.a.         & n.a.         & n.a.         & n.a.         & n.a.          \\
critic clip                      & 0.2          & lin\_0.3\footnotemark[1]  & n.a.         & n.a.         & n.a.         & n.a.             & n.a.       \\
importance ratio clip            & 0.2          & lin\_0.3\footnotemark[1]  & n.a.         & n.a.         & n.a.         & n.a.             & n.a.       \\
\multicolumn{8}{l}{}                                                                                                                                 \\
\hline
\multicolumn{8}{l}{}                                                                                                                                 \\
hidden layers                    & {[}128, 128] & {[}128, 128] & {[}128, 128] & {[}256, 256] & {[}256, 256] & {[}128, 128] & {[}128, 128] \\
hidden layers critic             & {[}128, 128] & {[}128, 128] & {[}128, 128] & {[}256, 256] & {[}256, 256] & {[}128, 128] & {[}128, 128]       \\
hidden activation                & tanh         & tanh         & tanh         & relu         & relu         & relu         & relu        \\
orthogonal initialization        & Yes          & No           & Yes          & fanin        & fanin         & Yes        & Yes               \\
initial std                      & 1.0          & 0.5          & 1.0          & 1.0          & 1.0          & 1.0          & 1.0              \\
\multicolumn{8}{l}{}                                                                                                                                 \\
\hline
\multicolumn{8}{l}{}                                                                                                                                 \\
\gls{mp} type                    & n.a.         & n.a.         & n.a.         & n.a.         & n.a.         & \pdmp   & \pdmp\\
number basis functions           & n.a.         & n.a.         & n.a.         & n.a.         & n.a.         & 8            & 5               \\
weight scale                     & n.a.         & n.a.         & n.a.         & n.a.         & n.a.         & 0.1           & 0.1               \\
\multicolumn{8}{l}{}\\   
\bottomrule
\end{tabular}
\end{adjustbox}
\end{table}

\footnotetext[1]{Linear Schedule from 0.3 to 0.01 during the first 25\% of the training. Then continued with 0.01.}

\afterpage{\newpage}
\begin{table}[ht]
\centering
\caption{Hyperparameters for the Box Pushing Dense}
\label{tab:boxpushing-HP}
\begin{adjustbox}{max width=\textwidth}
\begin{tabular}{lcccccccc}
\toprule
                                 & PPO          & gSDE         & TRPL         & SAC          & PINK     & \tcp      & BBRL        & BBRL Cov.   \\ 
\hline
\multicolumn{9}{l}{}                                                                                                  \\
number samples                   & 48000        & 80000        & 48000        & 8            & 8        & 152       & 152          & 152          \\
GAE $\lambda$                    & 0.95         & 0.95         & 0.95         & n.a.         & n.a.     & 0.95      & n.a.         & n.a.         \\
discount factor                  & 1.0          & 1.0          & 1.0          & 0.99         & 0.99     & 1.0       & n.a.         & n.a.         \\
\multicolumn{9}{l}{}                                                                                                                      \\ 
\hline
\multicolumn{9}{l}{}                                                                                                                      \\
$\epsilon_\mu$                   & n.a.         & n.a.         & 0.005        & n.a.         & n.a.     &0.05         & 0.1          & 0.05        \\
$\epsilon_\Sigma$                & n.a.         & n.a.         & 0.00005      & n.a.         & n.a.     &0.0005         & 0.00025         & 0.0005       \\
\multicolumn{9}{l}{}                                                                                                                      \\ 
\hline
\multicolumn{9}{l}{}                                                                                                                      \\
optimizer                        & adam         & adam         & adam         & adam         & adam     & adam         & adam         & adam         \\
epochs                           & 10           & 10           & 20           & 1            & 1        & 50           & 20          & 20          \\
learning rate                    & 5e-5         & 1e-4         & 5e-5         & 3e-4         & 3e-4     & 3e-4         & 3e-4         & 3e-4         \\
use critic                       & True         & True         & True         & True         & True     & True         & True         & True         \\
epochs critic                    & 10           & 10           & 10           & 1            & 1        & 50           & 10          & 10          \\
learning rate critic (and alpha) & 1e-4         & 1e-4         & 1e-4         & 3e-4         & 3e-4     & 1e-3    & 3e-4         & 3e-4         \\
number minibatches               & 40           & n.a.         & 40           & n.a.         & n.a.     & 1       & 1            & 1            \\
batch size                       & n.a.         & 2000         & n.a.         & 512          & 512      & n.a.    & n.a.         & n.a.         \\
buffer size                      & n.a.         & n.a.         & n.a.         & 2e6          & 2e6      & n.a.    & n.a.         & n.a.         \\
learning starts                  & 0            & 0            & 0            & 1e5          & 1e5      & 0       & 0            & 0            \\
polyak\_weight                   & n.a.         & n.a.         & n.a.         & 5e-3         & 5e-3     & n.a.    & n.a.         & n.a.         \\
trust region loss weight         & n.a.         & n.a.         & 10           & n.a.         & n.a.     & 1      & 10         & 10           \\
SDE sampling frequency           & n.a.         & 4            & n.a.         & n.a.         & n.a.     & n.a.    & n.a.         & n.a.         \\
entropy coefficient              & 0            & 0.01         & 0            & auto         & auto     & 0       & 0            & 0            \\
\multicolumn{9}{l}{}                                                                                                                      \\ 
\hline
\multicolumn{9}{l}{}                                                                                                                      \\
normalized observations          & True         & True         & True         & False        & False    & True    & False        & False        \\
normalized rewards               & True         & True         & False        & False        & False    & False    & False        & False        \\
observation clip                 & 10.0         & n.a.         & n.a.         & n.a.         & n.a.     & n.a.    & n.a.         & n.a.         \\
reward clip                      & 10.0         & 10.0         & n.a.         & n.a.         & n.a.     & n.a.    & n.a.         & n.a.         \\
critic clip                      & 0.2          & 0.2          & n.a.         & n.a.         & n.a.     & n.a.    & n.a.         & n.a.         \\
importance ratio clip            & 0.2          & 0.2          & n.a.         & n.a.         & n.a.     & n.a.    & n.a.         & n.a.         \\
\multicolumn{9}{l}{}                                                                                                                      \\ 
\hline
\multicolumn{9}{l}{}                                                                                                                      \\
hidden layers                    & {[}512, 512] & {[}256, 256] & {[}512, 512] & {[}256, 256] & {[}256, 256] &{[}128, 128] & {[}128, 128] & {[}128, 128] \\
hidden layers critic             & {[}512, 512] & {[}256, 256] & {[}512, 512] & {[}256, 256] & {[}256, 256] &{[}256, 256]   & {[}256, 256]   & {[}256, 256]   \\
hidden activation                & tanh         & tanh         & tanh         & tanh         & tanh         & relu        & relu         & relu         \\
orthogonal initialization        & Yes          & No           & Yes          & fanin        & fanin         & Yes        & Yes          & Yes         \\
initial std                      & 1.0          & 0.05         & 1.0          & 1.0          & 1.0          & 1.0         & 1.0          & 1.0          \\
\multicolumn{9}{l}{}                                                                                                                      \\ 
\hline
\multicolumn{9}{l}{}                                                                                                                                        \\
\gls{mp} type                    & n.a.         & n.a.         & n.a.         & n.a.         & n.a.        & \pdmp   & \pdmp    & \pdmp \\
number basis functions           & n.a.         & n.a.         & n.a.         & n.a.         & n.a.         & 8           & 8            & 8            \\
weight scale                     & n.a.         & n.a.         & n.a.         & n.a.         & n.a.         & 0.3        & 0.3         & 0.3
\\
\multicolumn{9}{l}{}\\   
\bottomrule
\end{tabular}
\end{adjustbox}
\end{table}

\afterpage{\newpage}
\begin{table}[ht]
\centering
\caption{Hyperparameters for the Box Pushing Sparse}
\label{tab:boxpushing-HP}
\begin{adjustbox}{max width=\textwidth}
\begin{tabular}{lcccccccc}
\toprule
                                 & PPO          & gSDE         & TRPL         & SAC          & PINK     & \tcp      & BBRL       & BBRL Cov.   \\ 
\hline
\multicolumn{9}{l}{}                                                                                                  \\
number samples                   & 48000        & 80000        & 48000        & 8            & 8        & 76       & 76          & 76          \\
GAE $\lambda$                    & 0.95         & 0.95         & 0.95         & n.a.         & n.a.     & 0.95      & n.a.         & n.a.         \\
discount factor                  & 1.0          & 1.0          & 1.0          & 0.99         & 0.99     & 1.0      & n.a.         & n.a.         \\
\multicolumn{9}{l}{}                                                                                                                      \\ 
\hline
\multicolumn{9}{l}{}                                                                                                                      \\
$\epsilon_\mu$                   & n.a.         & n.a.         & 0.005        & n.a.         & n.a.     &0.05         & 0.05         & 0.05        \\
$\epsilon_\Sigma$                & n.a.         & n.a.         & 0.00005      & n.a.         & n.a.     &0.0002         & 0.0005         & 0.0005       \\
\multicolumn{9}{l}{}                                                                                                                      \\ 
\hline
\multicolumn{9}{l}{}                                                                                                                      \\
optimizer                        & adam         & adam         & adam         & adam         & adam     & adam         & adam         & adam         \\
epochs                           & 10           & 10           & 20           & 1            & 1        & 50           & 100          & 100          \\
learning rate                    & 5e-4         & 1e-4         & 5e-5         & 3e-4         & 3e-4     & 3e-4         & 3e-4         & 3e-4         \\
use critic                       & True         & True         & True         & True         & True     & True         & True         & True         \\
epochs critic                    & 10           & 10           & 10           & 1            & 1        & 50    & 100          & 100          \\
learning rate critic (and alpha) & 1e-4         & 1e-4         & 1e-4         & 3e-4         & 3e-4     & 3e-4    & 3e-4       & 3e-4         \\
number minibatches               & 40           & n.a.         & 40           & n.a.         & n.a.     & 1    & 1            & 1            \\
batch size                       & n.a.         & 2000         & n.a.         & 512          & 512      & n.a.    & n.a.         & n.a.         \\
buffer size                      & n.a.         & n.a.         & n.a.         & 2e6          & 2e6      & n.a.    & n.a.         & n.a.         \\
learning starts                  & 0            & 0            & 0            & 1e5          & 1e5      & 0    & 0            & 0            \\
polyak\_weight                   & n.a.         & n.a.         & n.a.         & 5e-3         & 5e-3     & n.a.    & n.a.         & n.a.         \\
trust region loss weight         & n.a.         & n.a.         & 10           & n.a.         & n.a.     & 1    & 25         & 25           \\
SDE sampling frequency           & n.a.         & 4            & n.a.         & n.a.         & n.a.     & n.a.    & n.a.         & n.a.         \\
entropy coefficient              & 0            & 0.01         & 0            & auto         & auto       & 0      & 0           & 0           \\
\multicolumn{9}{l}{}                                                                                                                      \\ 
\hline
\multicolumn{9}{l}{}                                                                                                                      \\
normalized observations          & True         & True         & True         & False        & False    & True    & False        & False        \\
normalized rewards               & True         & True         & False        & False        & False    & False    & False        & False        \\
observation clip                 & 10.0         & n.a.         & n.a.         & n.a.         & n.a.     & n.a.    & n.a.         & n.a.         \\
reward clip                      & 10.0         & 10.0         & n.a.         & n.a.         & n.a.     & n.a.    & n.a.         & n.a.         \\
critic clip                      & 0.2          & 0.2          & n.a.         & n.a.         & n.a.     & n.a.    & 0.2          & n.a.         \\
importance ratio clip            & 0.2          & 0.2          & n.a.         & n.a.         & n.a.     & n.a.    & 0.2          & n.a.         \\
\multicolumn{9}{l}{}                                                                                                                      \\ 
\hline
\multicolumn{9}{l}{}                                                                                                                      \\
hidden layers                    & {[}512, 512] & {[}256, 256] & {[}512, 512] & {[}256, 256] & {[}256, 256] &{[}128, 128] & {[}128, 128] & {[}128, 128] \\
hidden layers critic             & {[}512, 512] & {[}256, 256] & {[}512, 512] & {[}256, 256] & {[}256, 256] &{[}256, 256]   & {[}256, 256]   & {[}256, 256]   \\
hidden activation                & tanh         & tanh         & tanh         & tanh         & tanh         & relu        & relu         & relu         \\
orthogonal initialization        & Yes          & No           & Yes          & fanin        & fanin         & Yes        & Yes          & Yes         \\
initial std                      & 1.0          & 0.05         & 1.0          & 1.0          & 1.0          & 1.0         & 1.0          & 1.0          \\
\multicolumn{9}{l}{}                                                                                                                      \\ 
\hline
\multicolumn{9}{l}{}                                                                                                                                        \\
\gls{mp} type                    & n.a.         & n.a.         & n.a.         & n.a.         & n.a.        & \pdmp   & \pdmp    & \pdmp \\
number basis functions           & n.a.         & n.a.         & n.a.         & n.a.         & n.a.         & 4           & 5            & 5            \\
weight scale                     & n.a.         & n.a.         & n.a.         & n.a.         & n.a.         & 0.3        & 0.3         & 0.3       
\\
\multicolumn{9}{l}{}\\   
\bottomrule
\end{tabular}
\end{adjustbox}
\end{table}

\afterpage{\newpage}
\begin{table}[ht]
\centering
\caption{Hyperparameters for the Hopper Jump}
\label{tab:boxpushing-HP}
\begin{adjustbox}{max width=\textwidth}
\begin{tabular}{lcccccccc}
\toprule
                                 & PPO          & gSDE         & TRPL         & SAC          & PINK     & \tcp   & BBRL   & BBRL Cov.   \\ 
\hline
\multicolumn{9}{l}{}                                                                                                  \\
number samples                   & 8000        & 8192         & 8000        & 1000         & 1        & 64       & 160          & 40          \\
GAE $\lambda$                    & 0.95         & 0.99         & 0.95         & n.a.       & n.a.     & 0.95      & n.a.         & n.a.         \\
discount factor                  & 1.0          & 0.999        & 1.0          & 0.99         & 0.99     & 1.0      & n.a.         & n.a.         \\
\multicolumn{9}{l}{}                                                                                                                      \\ 
\hline
\multicolumn{9}{l}{}                                                                                                                      \\
$\epsilon_\mu$                   & n.a.         & n.a.         & 0.05        & n.a.         & n.a.     &0.1         & n.a.         & 0.05        \\
$\epsilon_\Sigma$                & n.a.         & n.a.         & 0.0005      & n.a.         & n.a.     &0.005         & n.a.         & 0.0005       \\
\multicolumn{9}{l}{}                                                                                                                      \\ 
\hline
\multicolumn{9}{l}{}                                                                                                                      \\
optimizer                        & adam         & adam         & adam         & adam         & adam     & adam         & adam         & adam         \\
epochs                           & 10           & 10           & 20           & 1000         & 1        & 50           & 100          & 20          \\
learning rate                    & 3e-4         & 9.5e-5       & 3e-4         & 1e-4         & 2e-4     & 5e-4         & 1e-4         & 3e-4         \\
use critic                       & True         & True         & True         & True         & True     & True         & True         & True         \\
epochs critic                    & 10           & 10           & 10           & 1000         & 1        & 50    & 100          & 10          \\
learning rate critic (and alpha) & 3e-4         & 9.5e-5       & 3e-4         & 1e-4         & 2e-4     & 5e-4    & 1e-4         & 3e-4         \\
number minibatches               & 40           & n.a.         & 40           & n.a.         & n.a.     & 1    & 1            & 1            \\
batch size                       & n.a.         & 128          & n.a.         & 256          & 256      & n.a.    & n.a.         & n.a.         \\
buffer size                      & n.a.         & n.a.         & n.a.         & 1e6          & 1e6      & n.a.    & n.a.         & n.a.         \\
learning starts                  & 0            & 0            & 0            & 10000        & 1e5      & 0    & 0            & 0            \\
polyak\_weight                   & n.a.         & n.a.         & n.a.         & 5e-3         & 5e-3     & n.a.    & n.a.         & n.a.         \\
trust region loss weight         & n.a.         & n.a.         & 10           & n.a.         & n.a.     & 1    & n.a.         & 10           \\
SDE sampling frequency           & n.a.         & 8            & n.a.         & n.a.         & n.a.     & n.a.    & n.a.         & n.a.         \\
entropy coefficient              & 0           & 0.0025         & 0           & auto          & auto     & 0       & 0            & 0            \\
\multicolumn{9}{l}{}                                                                                                                      \\ 
\hline
\multicolumn{9}{l}{}                                                                                                                      \\
normalized observations          & True         & False        & True         & False        & False    & True    & False        & False        \\
normalized rewards               & True         & False        & False        & False        & False    & False    & False        & False        \\
observation clip                 & 10.0         & n.a.         & n.a.         & n.a.         & n.a.     & n.a.    & n.a.         & n.a.         \\
reward clip                      & 10.0         & 10.0         & n.a.         & n.a.         & n.a.     & n.a.    & n.a.         & n.a.         \\
critic clip                      & 0.2          & lin\_0.4     & n.a.         & n.a.         & n.a.     & n.a.    & 0.2          & n.a.         \\
importance ratio clip            & 0.2          & lin\_0.4     & n.a.         & n.a.         & n.a.     & n.a.    & 0.2          & n.a.         \\
\multicolumn{9}{l}{}                                                                                                                      \\ 
\hline
\multicolumn{9}{l}{}                                                                                                                      \\
hidden layers                    & {[}32, 32] & {[}256, 256] & {[}256, 256] & {[}256, 256] & {[}32, 32] &{[}128, 128] & {[}128, 128] & {[}128, 128] \\
hidden layers critic             & {[}32, 32] & {[}256, 256] & {[}256, 256] & {[}256, 256] & {[}32, 32]         &{[}128, 128]   & {[}32, 32]   & {[}32, 32]   \\
hidden activation                & tanh         & tanh         & tanh         & relu         & relu         & relu        & tanh         & relu         \\
orthogonal initialization        & Yes          & No           & Yes          & fanin        & fanin         & Yes        & Yes          & Yes         \\
initial std                      & 1.0          & 0.1          & 1.0          & 1.0          & 1.0          & 1.0         & 1.0          & 1.0          \\
\multicolumn{9}{l}{}                                                                                                                      \\ 
\hline
\multicolumn{9}{l}{}                                                                                                                                        \\
\gls{mp} type                    & n.a.         & n.a.        & n.a.         & n.a.         & n.a.        & \pdmp   & \pdmp    & \pdmp \\
number basis functions           & n.a.         & n.a.        & n.a.         & n.a.         & n.a.         & 5           & 5            & 5            \\
weight scale                     & n.a.         & n.a.        & n.a.         & n.a.         & n.a.         & 1        & 1         & 1       
\\
\multicolumn{9}{l}{}\\   
\bottomrule
\end{tabular}
\end{adjustbox}
\end{table}

\afterpage{\newpage}
\begin{table}[ht]
\centering
\caption{Hyperparameters for the Table Tennis}
\label{tab:boxpushing-HP}
\begin{adjustbox}{max width=\textwidth}
\begin{tabular}{lcccccccc}
\toprule
                                 & PPO          & gSDE         & TRPL         & SAC          & PINK     & \tcp   & BBRL   & BBRL Cov.   \\ 
\hline
\multicolumn{9}{l}{}                                                                                                  \\
number samples                   & 48000        & 24000        & 48000        & 8            & 8        & 76       & 76          & 76          \\
GAE $\lambda$                    & 0.95         & 0.95         & 0.95         & n.a.         & n.a.     & 0.95      & n.a.         & n.a.         \\
discount factor                  & 1.0          & 1.0          & 1.0          & 0.99         & 0.99     & 1.0      & n.a.         & n.a.         \\
\multicolumn{9}{l}{}                                                                                                                      \\ 
\hline
\multicolumn{9}{l}{}                                                                                                                      \\
$\epsilon_\mu$                   & n.a.         & n.a.         & 0.005        & n.a.         & n.a.     &0.005         & 0.005         & 0.005        \\
$\epsilon_\Sigma$                & n.a.         & n.a.         & 0.0005      & n.a.         & n.a.     &0.00005         & 0.001         & 0.001       \\
\multicolumn{9}{l}{}                                                                                                                      \\ 
\hline
\multicolumn{9}{l}{}                                                                                                                      \\
optimizer                        & adam         & adam         & adam         & adam         & adam     & adam         & adam         & adam         \\
epochs                           & 10           & 10           & 20           & 1            & 1        & 50           & 100          & 100          \\
learning rate                    & 5e-5         & 1e-4         & 5e-5         & 3e-4         & 3e-4     & 3e-4         & 1e-4         & 3e-4         \\
use critic                       & True         & True         & True         & True         & True     & True         & True         & True         \\
epochs critic                    & 10           & 10           & 10           & 1            & 1        & 50    & 100          & 100          \\
learning rate critic (and alpha) & 1e-4         & 1e-4         & 1e-4         & 3e-4         & 3e-4     & 3e-4    & 1e-4         & 3e-4         \\
number minibatches               & 40           & n.a.         & 40           & n.a.         & n.a.     & 1    & 1            & 1            \\
batch size                       & n.a.         & 4000         & n.a.         & 512          & 512      & n.a.    & n.a.         & n.a.         \\
buffer size                      & n.a.         & n.a.         & n.a.         & 4e6          & 4e6      & n.a.    & n.a.         & n.a.         \\
learning starts                  & 0            & 0            & 0            & 1e5          & 1e5      & 0    & 0            & 0            \\
polyak\_weight                   & n.a.         & n.a.         & n.a.         & 5e-3         & 5e-3     & n.a.    & n.a.         & n.a.         \\
trust region loss weight         & n.a.         & n.a.         & 10           & n.a.         & n.a.     & 1    & 25         & 25           \\
SDE sampling frequency           & n.a.         & 8            & n.a.         & n.a.         & n.a.     & n.a.    & n.a.         & n.a.         \\
entropy coefficient              & 0            & 0            & 0            & auto          & auto     & 0       & 0            & 0            \\
\multicolumn{9}{l}{}                                                                                                                      \\ 
\hline
\multicolumn{9}{l}{}                                                                                                                      \\
normalized observations          & True         & True         & True         & False        & False    & True    & False        & False        \\
normalized rewards               & True         & True         & False        & False        & False    & False    & False        & False        \\
observation clip                 & 10.0         & n.a.         & n.a.         & n.a.         & n.a.     & n.a.    & n.a.         & n.a.         \\
reward clip                      & 10.0         & 10.0         & n.a.         & n.a.         & n.a.     & n.a.    & n.a.         & n.a.         \\
critic clip                      & 0.2          & 0.2          & n.a.         & n.a.         & n.a.     & n.a.    & 0.2          & n.a.         \\
importance ratio clip            & 0.2          & 0.2          & n.a.         & n.a.         & n.a.     & n.a.    & n.a.          & n.a.         \\
\multicolumn{9}{l}{}                                                                                                                      \\ 
\hline
\multicolumn{9}{l}{}                                                                                                                      \\
hidden layers                    & {[}512, 512] & {[}256, 256] & {[}256, 256] & {[}256, 256] & {[}256, 256] &{[}256] & {[}256] & {[}256] \\
hidden layers critic             & {[}512, 512] & {[}256, 256] & {[}512, 512] & {[}256, 256] & {[}256, 256] &{[}256, 256]   & {[}256]   & {[}256]   \\
hidden activation                & tanh         & tanh         & tanh         & tanh         & tanh         & tanh        & tanh         & tanh         \\
orthogonal initialization        & Yes          & Yes          & Yes          & fanin        & fanin         & Yes        & Yes          & Yes         \\
initial std                      & 1.0          & 0.1          & 1.0          & 1.0          & 1.0          & 1.0         & 1.0          & 1.0          \\
\multicolumn{9}{l}{}                                                                                                                      \\ 
\hline
\multicolumn{9}{l}{}                                                                                                                                        \\
\gls{mp} type                    & n.a.         & n.a.         & n.a.         & n.a.         & n.a.        & \pdmp   & \pdmp    & \pdmp \\
number basis functions           & n.a.         & n.a.         & n.a.         & n.a.         & n.a.         & 3           & 3            & 3            \\
weight scale                     & n.a.         & n.a.         & n.a.         & n.a.         & n.a.         & 0.7        & 0.7         & 0.7       
\\
\multicolumn{9}{l}{}\\   
\bottomrule
\end{tabular}
\end{adjustbox}
\end{table}
\afterpage{\newpage}

\afterpage{\newpage}

\end{document}